\documentclass[10pt,twocolumn,letterpaper]{article}
\usepackage{cvpr}

\usepackage[T1]{fontenc}
\usepackage[utf8]{inputenc}
\usepackage[english]{babel}
\usepackage{newtxtext}
\usepackage[vvarbb]{newtxmath}
\usepackage{courier}
\usepackage{textcomp}
\usepackage[final]{microtype}
\usepackage{hyphenat}
\usepackage[all]{nowidow}
\usepackage{enumitem}

\usepackage{amsmath}
\usepackage{amssymb}
\usepackage{amsfonts}
\usepackage{bm}
\usepackage{xfrac}
\usepackage{siunitx}

\usepackage{booktabs}
\usepackage{tabulary}

\usepackage{epsfig}
\usepackage{graphicx}
\usepackage[labelsep=period,labelfont=bf,font={small}]{caption}
\usepackage{subcaption}
\usepackage{dblfloatfix} %
\usepackage{pifont}
\usepackage{lipsum}
\usepackage[export]{adjustbox}
\usepackage{float}

\usepackage{algorithm}
\usepackage{algpseudocode}
\usepackage{algorithmicx}
\usepackage{listings}

\usepackage{xcolor}
\usepackage{soul}

\usepackage[pagebackref=true,breaklinks=true,letterpaper=true,colorlinks,bookmarks=false]{hyperref}
\usepackage{cleveref}
\usepackage{multicol}

\hypersetup{
    pdfstartview={FitH},
    pdftitle={Cloth in the Wind: A Case Study of Physical Measurement through Simulation},
    pdfauthor={Tom Runia, Kirill Gavrilyuk, Cees Snoek, Arnold Smeulders},
    pdfsubject={Computer Vision and Pattern Recognition},
    pdfcreator={Tom Runia},
    pdfproducer={Tom Runia},
    pdfkeywords={Deep Learning} {Wind Estimation} {Physical Similarity} {CVPR} {CVPR 2020} {Seattle} {Computer Vision and Pattern Recognition},
}

\graphicspath{{./images/}}

\newcommand{\comment}[1]{\ignorespaces}

\newcommand{\overbar}[1]{\mkern 1.5mu\overline{\mkern-1.5mu#1\mkern-1.5mu}\mkern 1.5mu}

\def\*#1{\bm{#1}}  %
\DeclareMathAlphabet\mathcalbf{OMS}{cmsy}{b}{n}
\newcommand{\btheta}{\bm{\theta}}

\DeclareMathOperator*{\argmin}{arg\,min}

\AtBeginDocument{
 \abovedisplayskip=10pt plus 3pt minus 6pt
 \abovedisplayshortskip=0pt plus 3pt
 \belowdisplayskip=10pt plus 3pt minus 6pt
 \belowdisplayshortskip=7pt plus 3pt minus 4pt
}

\definecolor{highlightcolor}{RGB}{255,237,191}
\definecolor{textboxcolor}{RGB}{155,155,155}
\definecolor{paperboxcolor}{RGB}{204,148,73}
\definecolor{highlightred}{RGB}{224,103,103}
\definecolor{greencustom}{rgb}{0.0, 0.5, 0.0}
\definecolor{lightgray}{rgb}{0.83, 0.83, 0.83}

\newcommand{\ra}[1]{\renewcommand{\arraystretch}{#1}}

\newcommand{\ms}{\si{\meter\per\second}}

\newcommand{\kgmm}{\si{\kilo\gram\per\square\meter}}

\colorlet{punct}{red!60!black}
\definecolor{background}{HTML}{f4f4f4}
\definecolor{delim}{RGB}{20,105,176}
\colorlet{numb}{magenta!60!black}

\definecolor{bluekeywords}{rgb}{0.13,0.13,1}
\definecolor{greencomments}{rgb}{0,0.5,0}
\definecolor{redstrings}{rgb}{0.9,0,0}

\lstdefinelanguage{json}{
    basicstyle=\small\ttfamily,
    showspaces=false,
    showtabs=false,
    breaklines=true,
    showstringspaces=false,
    breakatwhitespace=true,
    numbers=left,
    numberstyle=\scriptsize,
    stepnumber=1,
    numbersep=8pt,
    framexleftmargin=15pt,
    frame=single,
    showstringspaces=false,
    backgroundcolor=\color{background},
    literate=
     *{0}{{{\color{numb}0}}}{1}
      {1}{{{\color{numb}1}}}{1}
      {2}{{{\color{numb}2}}}{1}
      {3}{{{\color{numb}3}}}{1}
      {4}{{{\color{numb}4}}}{1}
      {5}{{{\color{numb}5}}}{1}
      {6}{{{\color{numb}6}}}{1}
      {7}{{{\color{numb}7}}}{1}
      {8}{{{\color{numb}8}}}{1}
      {9}{{{\color{numb}9}}}{1}
      {:}{{{\color{punct}{:}}}}{1}
      {,}{{{\color{punct}{,}}}}{1}
      {\{}{{{\color{delim}{\{}}}}{1}
      {\}}{{{\color{delim}{\}}}}}{1}
      {[}{{{\color{delim}{[}}}}{1}
      {]}{{{\color{delim}{]}}}}{1},
}

\cvprfinalcopy %

\def\cvprPaperID{2463} %

\ifcvprfinal\fi

\title{Cloth in the Wind:\\[0.25em] {\large \normalfont \em A Case Study of Physical Measurement through Simulation}}

\author{Tom F.~H. Runia \hspace{1.4em} Kirill Gavrilyuk \hspace{1.4em} Cees G.~M. Snoek \hspace{1.4em} Arnold W.~M. Smeulders \\[1mm]
    QUVA Deep Vision Lab, University of Amsterdam\\[1mm]
    {\tt\small \{runia,kgavrilyuk,cgmsnoek,a.w.m.smeulders\}@uva.nl}
}

\begin{document}

\maketitle
\thispagestyle{empty}

\begin{abstract}
  \vspace{-3mm}
  For many of the physical phenomena around us, we have developed sophisticated models explaining their behavior. Nevertheless, measuring physical properties from visual observations is challenging due to the high number of causally underlying physical parameters -- including material properties and external forces. In this paper, we propose to measure latent physical properties for cloth in the wind without ever having seen a real example before. Our solution is an iterative refinement procedure with simulation at its core. The algorithm gradually updates the physical model parameters by running a simulation of the observed phenomenon and comparing the current simulation to a real-world observation. The correspondence is measured using an embedding function that maps physically similar examples to nearby points. We consider a case study of cloth in the wind, with curling flags as our leading example -- a seemingly simple phenomena but physically highly involved. Based on the physics of cloth and its visual manifestation, we propose an instantiation of the embedding function. For this mapping, modeled as a deep network, we introduce a spectral layer that decomposes a video volume into its temporal spectral power and corresponding frequencies. Our experiments demonstrate that the proposed method compares favorably to prior work on the task of measuring cloth material properties and external wind force from a real-world video.
\end{abstract}

\vspace{-5mm}

\section{Introduction}
\label{sec:introduction}

There is substantial evidence \cite{hegarty2004mechanical,craik1967nature} that humans run mental models to predict physical phenomena. We predict the trajectory of objects in mid-air, estimate a liquid's viscosity and gauge the velocity at which an object slides down a ramp. In analogy, simulation models usually optimize their parameters by performing trial runs and selecting the best. Over the years, physical models of the world have become so visually appealing through simulations and rendering \cite{wang2011data,narain2012adaptive,bridson2005simulation,schreck2019fundamental} that it is worthwhile to consider them for physical scene understanding. This alleviates the need for meticulous annotation of the pose, illumination, texture and scene dynamics as the model delivers them for free.

In this paper, we consider flags and cloth in the wind as a case study. Measurements and visual models of flags and cloth are important for virtual clothing try-on \cite{yang2018physics}, energy harvesting and biological systems \cite{shelley2011flapping,huang2010three}. The cloth's intrinsic material properties, together with the external wind force, determine its dynamics. Untangling the dynamics of fabric is challenging due to the involved nature of the air-cloth interaction: a flag exerts inertial and elastic forces on the surrounding air, while the air acts on the fabric through pressure and viscosity \cite{huang2010three}. As we seek to measure both the cloth's intrinsic material properties and the external wind force, our physical model couples a non-linear cloth model \cite{wang2011data} with external wind force \cite{wejchert1991animation}. 

\begin{figure}	
	\centering	
    \includegraphics[width=\columnwidth,trim={0 17.5cm 17cm 0},clip]{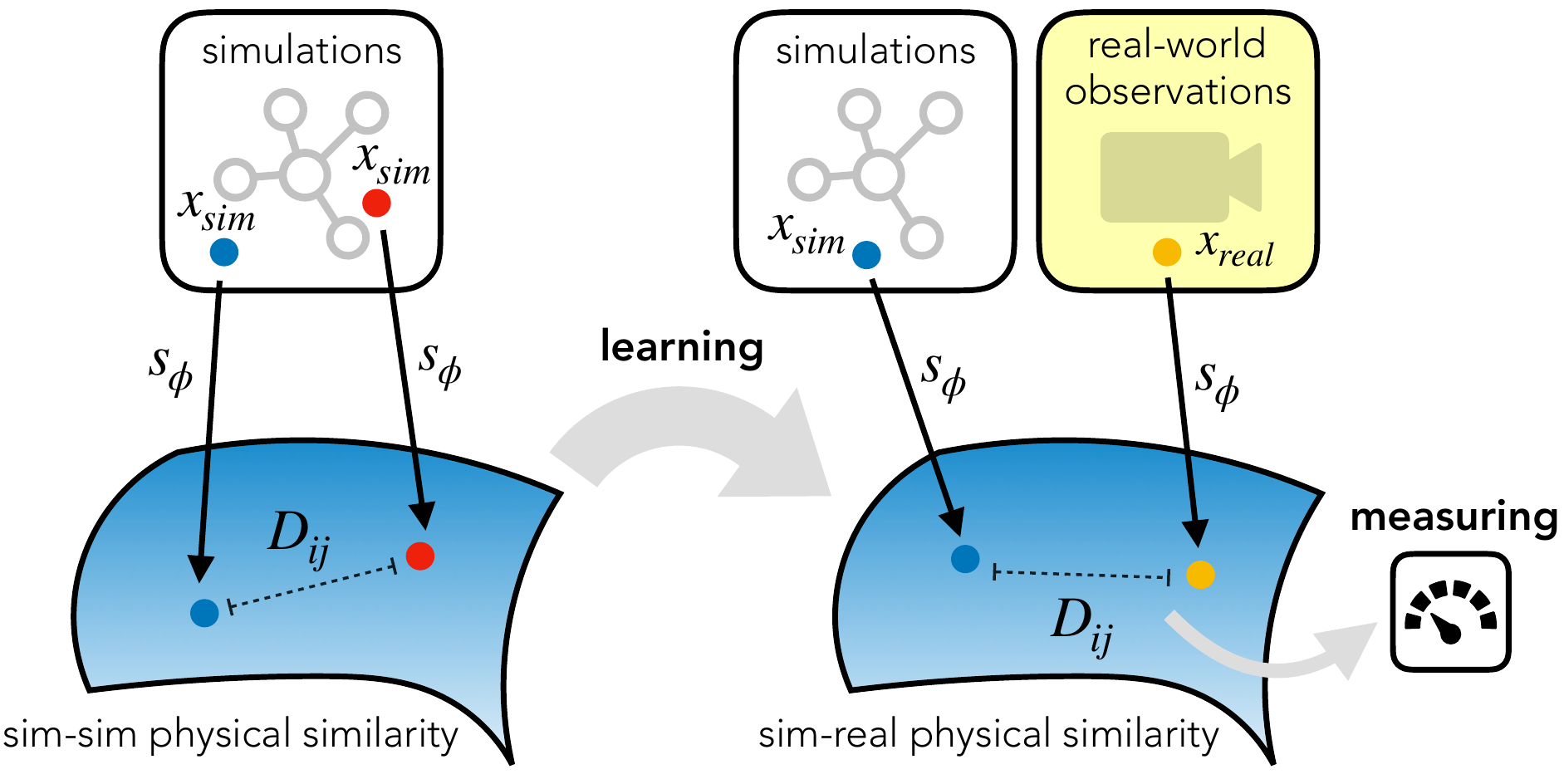}	
    \vspace{-5mm}	
    \caption{We propose to measure real-world physical cloth parameters without ever having seen the phenomena before. From cloth simulations only, we learn a distance metric that encodes both intrinsic and extrinsic physical properties. After learning, we use the embedding function to measure physical parameters from real-world video by comparison to its simulated counterpart. \label{fig:intro-figure}}	
    \vspace{-4mm}	
\end{figure}

The task is challenging, as physical models of cloth tend to have high numbers of unknown parameters and bear intricate coupling of intrinsic and external forces. Our solution is to compare pairs of real and simulated observations and measure their physical similarity. As there is a fundamental caveat in the use of simulation and rendering for learning: ``visually appealing'' does not necessarily imply the result is realistic, the main question is how to assess the similarity of the causally underlying physical parameters rather than visual correspondence. It might be the case that the image looks real but never occurs in reality. %

At the core of our measurement is a cloth simulation engine with unknown parameters $\btheta$ to be determined. The outcome of a simulation (\eg 3D meshes, points clouds, flow vectors) is converted to the image space using a render engine. We then compare the simulated visual data with a real-world observation of the particular phenomenon (\Cref{fig:intro-figure}). Accordingly, we propose to learn a \emph{physical similarity} metric from simulations only, without ever perceiving a real-world example. In the learned embedding space, observations with similar physical parameters will wind up close, while dissimilar example pairs will be further away. Guided by the physical similarity, the simulation's parameters are refined in each step. As a result, we obtain a complete computational solution for the refined measurements of physical parameters.

\begin{figure}[t]
    \centering
    \includegraphics[width=\columnwidth,trim={0 19.4cm 16.5cm 0},clip]{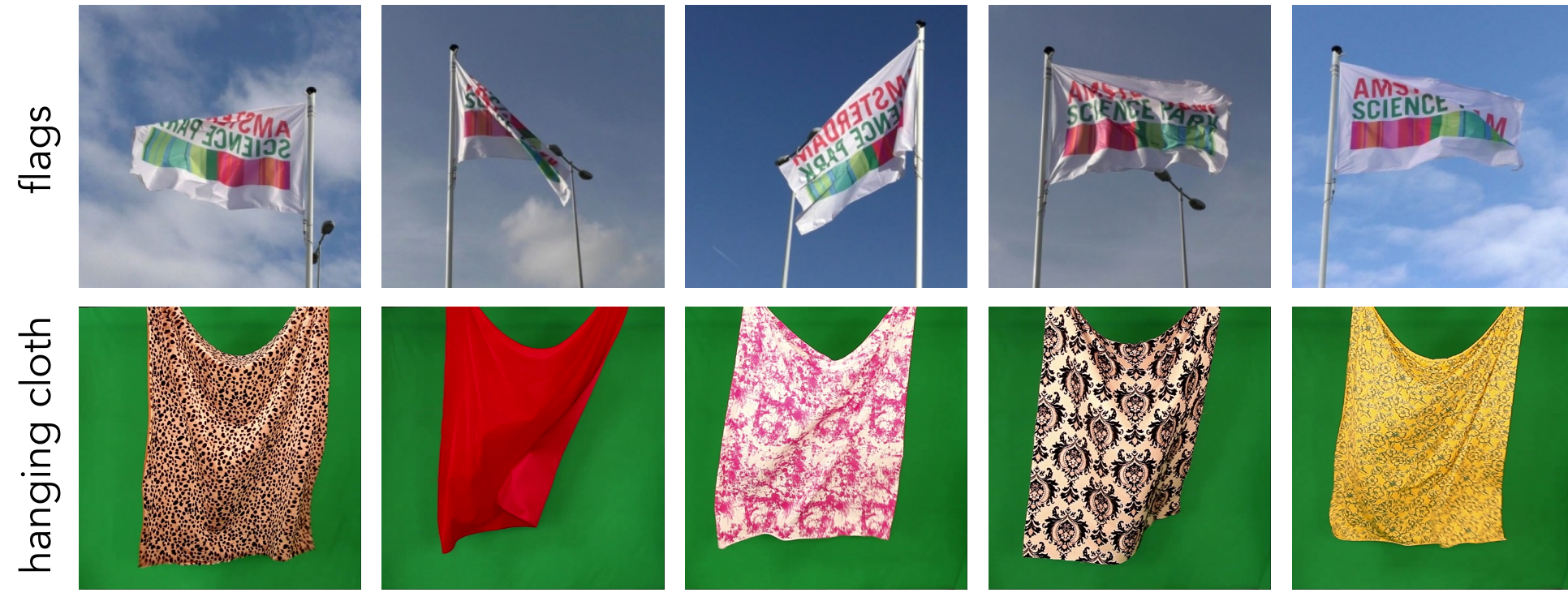}
    \vspace{-5mm}
    \caption{We consider two cases of cloth in the wind. Top row: random still images from our video recordings of real flags. Bottom row: examples from the Bouman \etal hanging cloth dataset \cite{bouman2013estimating}.}\label{fig:flagreal-hanging-cloth}
    \vspace{-5mm}
\end{figure}

Our contributions are as follows: (1) We propose to train a perception-based physical cloth measurement device from simulations only, without ever observing a real-world manifestation of the phenomena. Our measurement device is formulated as a comparison between two visual observations implemented as a Siamese network that we train with contrastive loss. (2) In a case study of cloth, we propose a specific instantiation of the physical embedding function. At its core is a new spectral decomposition layer that measures the spectral power over the cloth's surface. Our solution compares favorably to existing work that recovers intrinsic and extrinsic physical properties from visual observations. (3) To evaluate our method, we record real-world video of flags with the ground-truth wind speed gauged using an anemometer. (4) Finally, we iteratively refine physics simulations from a single real-world observation towards maximizing the physical similarity between the real-world and its simulation.

\begin{figure*}[t]
    \centering
    \includegraphics[width=\textwidth,trim={0 18.6cm 10.4cm 0},clip]{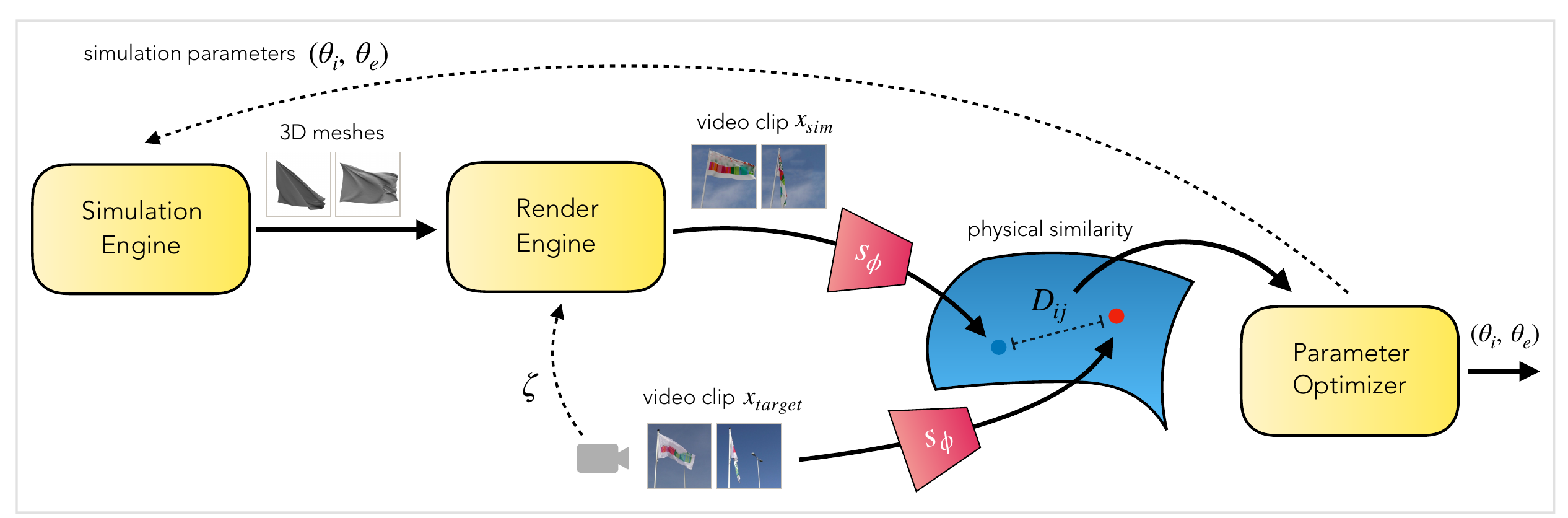}
    \vspace{-6mm}
    \caption{We propose the perception-based measurement of physical scene properties. Given an observation of a real-world physical phenomenon, here represented as video clip $\*x_{\text{target}}$, our algorithm measures the underlying parameters of the physical scene. Central is a simulation engine implementing the physical model, parametrized by intrinsic material properties $\btheta_i$ and the characterization of external forces $\btheta_e$. A render engine, with render parameters $\*\zeta$, maps the simulator's output to the image space producing video clip $\*x_{\text{sim}}$. Using an embedding function $s_{\phi}(\*x)$, both real and simulated examples are mapped to a manifold on which physically similar examples are assigned to nearby points. To measure the similarity between both clips, we evaluate a distance metric $D_{ij}(\cdot,\cdot)$ in the embedding space. Its result serves as the objective for an optimization module that refines the physical parameters $\btheta$ towards the actual observation.} \label{fig:method-overview}
    \vspace{-2mm}
\end{figure*}

\section{Related Work}
\label{sec:related-work}

 Previous work has measured physical properties by perceiving real-world objects or phenomena -- including material properties \cite{davis2015visual}, cloth stiffness and bending parameters \cite{bouman2013estimating,yang2017learning}, mechanical features \cite{wu2015galileo,mottaghi2016newtonian,mottaghi2016happens,li2016fall}, fluid characteristics \cite{wu2016physics,spencer2004water,sakaino2008fluid} and surface properties \cite{meka2018lime}. The primary focus of the existing literature has been on estimating intrinsic material properties from visual input. However, physical phenomena are often described by the interaction between intrinsic and extrinsic properties. Therefore, we consider the more complex scenario of jointly estimating intrinsic material properties and extrinsic forces from a single real-world video through the iterative refinement of physics simulations. 

Our case study focuses on the physics of cloth and flags, both of which belong to the broader category of wind-excited bodies. The visual manifestation of wind has received modest attention in computer vision, \eg the oscillation of tree branches \cite{xue2018seeing,sun2003video}, water surfaces \cite{spencer2004water}, and hanging cloth \cite{bouman2013estimating,yang2017learning,wang2019learning,cardona2019seeing}. Our leading example of a flag curling in the wind may appear simple at first, but its motion is highly complex. Its dynamics are an important and well-studied topic in the field of fluid-body interactions \cite{shelley2011flapping,taneda1968waving,tian2013role}. Inspired by this work and existing visual cloth representations that characterize wrinkles, folds and silhouette \cite{bhat2003estimating,haddon1998shading,white2007capturing,yang2018physics}, we propose a novel spectral decomposition layer which encodes the frequency distribution over the cloth's surface. 

Previous work has considered the task of measuring intrinsic cloth parameters \cite{bhat2003estimating,bouman2013estimating,yang2017learning} or external forces \cite{cardona2019seeing} from images or video. Notably, Bouman \etal \cite{bouman2013estimating} use complex steerable pyramids to describe hanging cloth in a video, while both Yang \etal \cite{yang2017learning} and Cardona \etal \cite{cardona2019seeing} propose a learning-based approach by combining a convolutional network and recurrent network. In our experiments we will compare our cloth frequency-based representations with Cardona \etal \cite{cardona2019seeing} on flags while Yang \etal \cite{yang2017learning} is a reference on the hanging cloth dataset of Bouman \etal \cite{bouman2013estimating}. 

Our approach of measuring physical parameters by iterative refinement of simulations shares similarity to the Monte Carlo-based parameter optimization of \cite{wu2015galileo} and the particle swarm refinement of clothing parameters from static images \cite{yang2018physics}. In particular, the work of \cite{yang2018physics} resembles ours as they infer garment properties from images for the purpose of virtual clothing try-on. However, our work is different in an important aspect: we estimate intrinsic and extrinsic physical parameters from video while their work focuses on estimating intrinsic cloth properties from static equilibrium images. Recently, Liang \etal \cite{liang2019differentiable} have proposed a differentiable cloth simulator which could potentially be used as an alternative to our approach for cloth parameter estimation.

\section{Method}
\label{sec:method}

We consider the scenario in which we make an observation of some phenomena with a physical model explaining its manifestation available to us. Based on the perception of reality, our goal is to measure the $D_p$ unknown continuous parameters of the physical model $\btheta \in \mathbb{R}^{D_p}$, consisting of intrinsic parameters $\btheta_i$ and extrinsic parameters $\btheta_e$ through an iterative refinement of a computer simulation that implements the physical phenomena at hand. In particular, we consider observations in the form of short video clips $\*x_{\text{target}} \in \mathbb{R}^{C\times N_t \times H \times W}$, with $C$ denoting the number of image channels and $N_t$ the number of $H \times W$ frames. In each iteration, the simulator runs with current model parameters $\btheta$ to produce some intermediate representation (\eg 3D meshes, point clouds or flow vectors), succeeded by a render engine with parameters $\*\zeta$ that yields a simulated video clip $\*x_{\text{sim}} \in \mathbb{R}^{C\times N_t \times H \times W}$. Our insight is that the physical similarity between real-world observation and simulation can be measured in some embedding space using pairwise distance:
\begin{align}
    D_{i,j} = D\left(s_{\phi}(\*x_i), s_{\phi}(\*x_j)\right) : \mathbb{R}^{D_e}\times \mathbb{R}^{D_e} \rightarrow \mathbb{R}
    \label{eq:distance-function}
\end{align}
where $s_{\phi}(\*x) : \mathbb{R}^{C\times N_t \times H \times W} 
\rightarrow \mathbb{R}^{D_e}$ an embedding function parametrized by $\phi$ that maps the data manifold $\mathbb{R}^{C\times N_t \times H \times W}$ to some embedding manifold $\mathbb{R}^{D_e}$ on which physically similar examples should lie close. In each iteration, guided by the pairwise distance \eqref{eq:distance-function} between real and simulated instance, the physical model is refined to maximize physical similarity. This procedure ends whenever the physical model parameters have been measured accurately enough or when the evaluation budget is finished. The output comprises the measured physical parameters $\btheta^*$ and corresponding simulation $\*x_{\text{sim}}^*$ of the real-world phenomenon. An overview of the proposed method is presented in \mbox{\Cref{fig:method-overview}}.

\subsection{Physical Similarity}
\label{subsec:similarity-function}

For the measurement to be successful, it is crucial to measure the similarity between simulation $\*x_{\text{sim}}$ and real-world observation $\*x_{\text{target}}$. The similarity function must reflect correspondence in physical dynamics between the two instances. The prerequisite is that the physical model must describe the phenomenon's behavior at the scale that coincides with the observational scale. For example, the quantum mechanical understanding of a pendulum will be less meaningful than its formulation in classical mechanics when capturing its appearance using a regular video camera. 

Given the physical model and its implementation as a simulation engine, we generate a dataset of simulations with its parameters $\btheta$ randomly sampled from some predefined search space. For each of these simulated representations of the physical phenomenon, we use a 3D render engine to generate multiple video clips $\*x_{\text{sim},}^i$ with different render parameters $\*\zeta^i$. As a result, we obtain a dataset with multiple renders for each simulation instance. Given this dataset we propose the following training strategy to learn a distance metric quantifying the physical similarity between observations.

We employ a \emph{contrastive loss} \cite{hadsell2006dimensionality} and consider positive example pairs to be rendered video clips originating from the same simulation (\ie sharing physical parameters) while negative example pairs have different physical parameters. Both rendered video clips of an example pair are mapped to the embedding space through $s_{\phi}(\*x)$ in Siamese fashion \cite{bromley1994signature}. In the embedding space, the physical similarity will be evaluated using the squared Euclidean distance: $D_{i,j} = D\left(s_{\phi}(\*x_i), s_{\phi}(\*x_j)\right) = \Vert s_{\phi}(\*x_i) - s_{\phi}(\*x_j) \Vert_2^2$. If optimized over a collection of rendered video clips, the contrastive loss asserts that physically similar examples are pulled together, whereas physically dissimilar points will be pushed apart. As a result, by training on simulations only, we can learn to measure the similarity between simulations and the real-world pairs.

\subsection{Simulation Parameter Optimization}
\label{subsec:parameter-optimization}

We will arrive at a measurement through gradual refinement of the simulations (\Cref{fig:method-overview}). To optimize the physical parameters we draw the parallel with the problem of hyperparameter optimization \cite{snoek2012practical,bergstra2012random}. In light of this correspondence, our collection of model parameters is analogous to the hyperparameters involved by training deep neural networks (\eg learning rate, weight decay, dropout). Formally, we seek to find the global optimum of physical parameters: 
\begin{align}
    \btheta^* = \argmin_{\btheta} \, D\left(s_{\phi}(\*x_\text{target}), s_{\phi}(\*x_\text{sim}(\btheta))\right),
    \label{eq:hyperparam-optimum}
\end{align}
where the target example is fixed and the simulated example depends on the current set of physical parameters $\btheta$. Adjusting the parameters $\btheta$ at each iteration is challenging as it is hard to make parametric assumptions on \eqref{eq:hyperparam-optimum} as function of $\btheta$ and accessing the gradient is costly due to the simulations' computational complexity. Our goal is, therefore, to estimate the global minimum with as few evaluations as possible. Considering this, we adopt Bayesian optimization \cite{snoek2012practical} for updating parameters $\btheta$. Its philosophy is to leverage all available information from previous observations of \eqref{eq:hyperparam-optimum} and not only use local gradient information. We treat the optimization as-is and use a modified implementation of Spearmint \cite{snoek2012practical} with the Mat\'ern52 kernel and improved initialization of the acquisition function \cite{oh2018bock}. Note that the embedding function $s_{\phi}(\*x)$ is fixed throughout this optimization. 

\section{Physics, Simulation and Appearance of Cloth}
\label{sec:physical-model}

Up until now, we have discussed the proposed method in general terms and made no assumptions on physical phenomena. In this paper, we will consider two cases of cloth exposed to the wind: curling flags and hanging cloth (\Cref{fig:flag-illustration}). To proceed, we need to confine the parameters $\btheta$ and design an appropriate embedding function $s_{\phi}(\*x)$. 

\subsection{Physical Model}
\label{subsec:physical-model-of-flags}

The physical understanding of cloth and its interaction with external forces has been assimilated by the computer graphics community. Most successful methods treat cloth as a mass-spring model: a dense grid of point masses organized in a planar structure, inter-connected with different types of springs which properties determine the fabric's behavior \cite{baraff1998large,provot1995deformation,wang2011data,baraff2003untangling,narain2012adaptive}. We adopt Wang's \etal \cite{wang2011data} non-linear and anisotropic mass-spring model for cloth. This model uses a piecewise linear bending and stretching formulation. The stretching model is a generalization of Hooke's law for continuous media \cite{slaughter2012linearized}. As our experiments focus on flags in the wind for which the stretching properties are of minimal relevance, our experiments will focus on flags in the wind, typically made of strong weather-resistant material such as polyester and nylon. Therefore, the material's stretching properties are of minimal relevance and we will emphasize on the cloth's bending model \cite{wang2011data} and external forces \cite{wejchert1991animation}.

\vspace{2mm}

\begin{figure}
	\centering
    \includegraphics[width=\columnwidth,trim={0 16.5cm 20.5cm 0},clip]{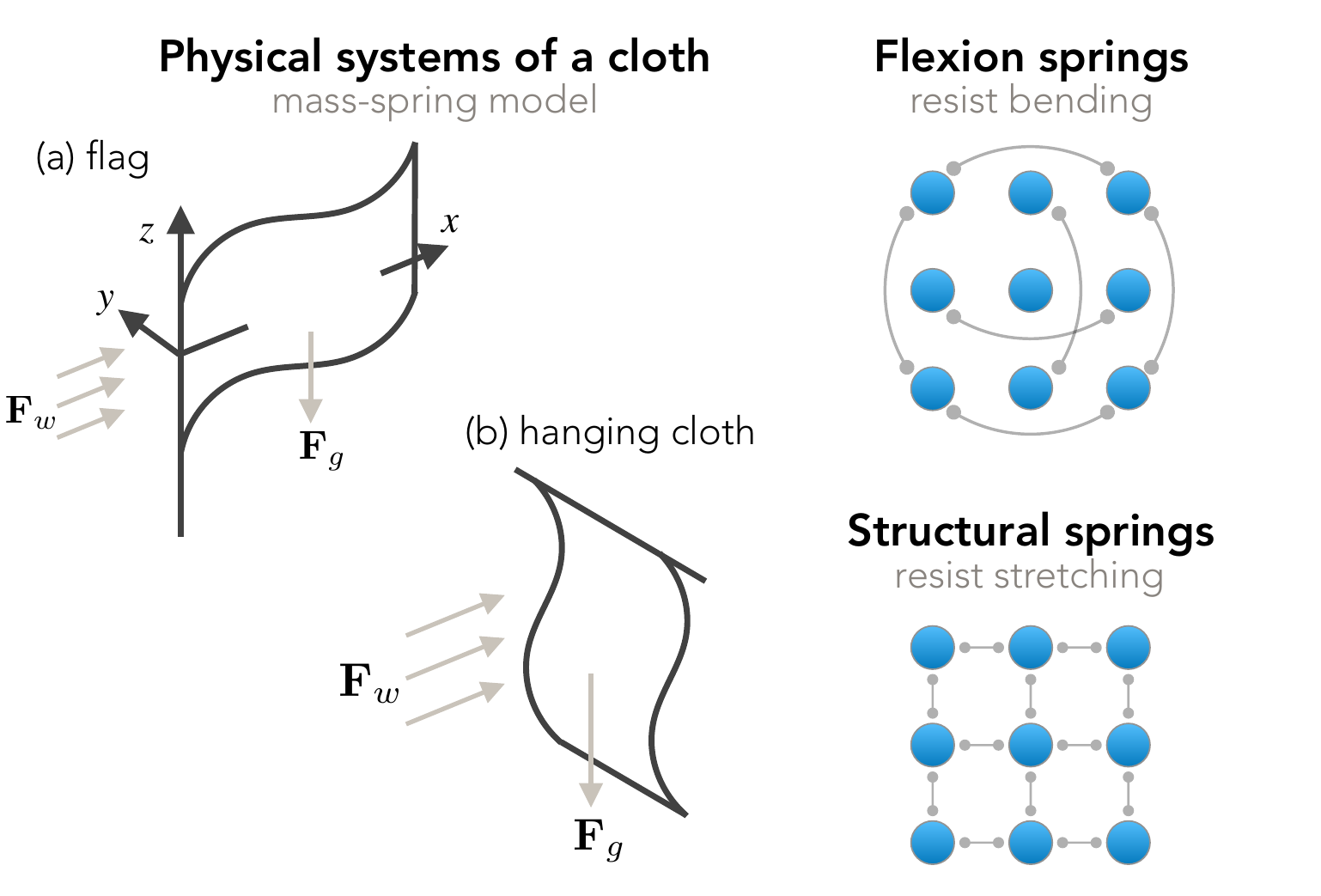}
    \vspace{-7mm}
    \caption{\emph{Left:} we consider two cases of cloth exposed in the wind: (a) a flag curling in the wind; and (b) cloth fabric hanging from a rod. In both cases, the fabric fabric is treated as a mass-spring model in which a dense grid of point masses is inter-connected with multiple springs. \emph{Right:} the bending and stretching springs determine the materials behavior. Flexion springs act over shared edges whereas structural springs connect to direct neighbors. \label{fig:flag-illustration}}
    \vspace{-3mm}
\end{figure}

\noindent \textbf{Bending Model ($\btheta_i$).} The bending model is based on the linear bending force equation first proposed in \cite{bridson2005simulation}. The model formulates the elastic bending force $\*{F}_e$ over triangular meshes sharing an edge (\Cref{fig:flag-illustration}). For two triangles separated by the dihedral angle $\varphi$, the bending force reads:
\begin{align}
    \*{F}_e = k_e \sin(\varphi / 2)(N_1 + N_2)^{-1} \vert \*{E} \vert \*{u},
    \label{eq:bending-equation}
\end{align}
where $k_e$ is the material dependent bending stiffness, $N_1, N_2$ are the weighted surface normals of the two triangles, $\*{E}$ represents the edge vector and $\*{u}$ is the bending mode (see Figure~1 in \cite{bridson2005simulation}). The bending stiffness $k_e$ is non-linearly related to the dihedral angle $\varphi$. This is realized by treating $k^e$ as piecewise linear function of the reparametrization $\alpha = \sin(\varphi/2)(N_1+N_2)^{-1}$. After this reparametrization, for a certain fabric, the parameter space is sampled for $N_b$ angles yielding a total of $3N_b$ parameters across the three directions. Wang \etal \cite{wang2011data} empirically found that $5$ measurements are sufficient for most fabrics, producing $15$ bending parameters. 

\vspace{2mm}

\noindent \textbf{External Forces ($\btheta_e$).} For the dynamics of cloth, we consider two external forces acting upon its planar surface. First, the Earth's gravitational acceleration (\mbox{$\*F_g = m\*a_g$}) naturally pushes down the fabric. The total mass is defined by the cloth's area weight $\rho_A$ multiplied by surface area. More interestingly, we consider the fabric exposed to a constant wind field. Again, modeling the cloth as a grid of point masses, the drag force on each mass is stipulated by Stokes's equation $\*F_d = 6 \pi R \eta \*v_w$ in terms of the surface area, the air's dynamic viscosity and the wind velocity $\*v_w$ \cite{wejchert1991animation,narain2012adaptive}. By all means, this is a simplification of reality. Our model ignores terms associated with the Reynolds number (such as the cloth's drag coefficient), which will also affect a real cloth's dynamics. However, it appears that the model is accurate enough to cover the spectrum of cloth dynamics. %

\subsection{Simulation Engine}
\label{subsec:simulation-engine}

We employ the non-differentiable ArcSim simulation engine \cite{narain2012adaptive} which efficiently implements the complex physical model described in \Cref{subsec:physical-model-of-flags}. On top of the physical model, the simulator incorporates anisotropic remeshing to improve detail in densely wrinkled regions while coarsening flat regions. As input, the simulator expects the cloth's initial mesh, its material properties and the configuration of external forces. At each time step, the engine solves the system for implicit time integration using a sparse Cholesky-based solver. This produces a sequence of 3D cloth meshes based on the physical properties of the scene. As our goal is to learn a physical distance metric in image space between simulation and a real-world observation, we pass the sequence of meshes through a 3D render engine \cite{blender2018}. Given render parameters $\*\zeta$ comprising of camera position, scene geometry, lighting conditions and the cloth's visual texture, the renderer produces a simulated video clip ($\*x_{\text{sim}}$) which we can compare directly to the real-world observation ($\*x_{\text{target}}$). We emphasize that our focus is neither on inferring render parameters $\*\zeta$ from observations nor on attaining visual realism for our renders.

\vspace{2mm}

\noindent \textbf{Parameter Search Space ($\btheta_i, \btheta_e$).} The ArcSim simulator \cite{narain2012adaptive} operates in metric units, enabling convenient comparison with real-world dynamics. As the base material for our flag experiments, we use ``Camel Ponte Roma'' from \cite{wang2011data}. Made of $60\%$ polyester and $40\%$ nylon, this material closely resembles widely used flag fabrics \cite{wang2011data}. The fabric's bending coefficients, stretching coefficients, and area weight were accurately measured in a mechanical setup by the authors. We adopt and fix their stretching parameters and use the bending stiffness and area weight as initialization for our cloth material. Specifically, using their respective parameters we confine a search space that is used during our parameter refinement. We determine $\rho_A \sim \text{Uniform}(0.10, 0.17)$~\kgmm{} after consulting various flag materials at online retailers. And, we restrict the range of the bending stiffness coefficients by multiplying the base material's $\overbar{k}_e$ in \eqref{eq:bending-equation} by $10^{-1}$ and $10$ to obtain the most flexible and stiffest material respectively. As the bending coefficients have a complex effect on the cloth's appearance, we independently optimize the $15$ bending coefficients instead of only tuning the one-dimensional multiplier. The full parameter search space is listed in \Cref{tab:parameter-search-space}. %

\begin{table}
    \centering
    \caption{The predefined parameter range for optimization of $\btheta = (\btheta_i, \btheta_e)$ given the physical model of a flag curling in the wind. The bending parameters $\overbar{k}_e$ correspond to the ``Camel Ponte Roma'' base material from \cite{wang2011data}. \label{tab:parameter-search-space} } %
    \ra{1.1}
    \small
    \vspace{-2mm}
    \scalebox{0.96}[1.0]{
        \begin{tabular}{llcl}
            \toprule
            & Parameter & Params & Search space \\ 
            \midrule
            $\theta_i$ & Bending stiffness  & $15$ & $k_e \in [10^{-1}\overbar{k}_e, 10\overbar{k}_e]$ \\ 
            $\theta_i$  & Fabric area weight & $1$  & $\rho_A \in [0.10, 0.17]$ \kgmm{} \\ 
            $\theta_e$ & Wind velocity      & $1$  & $v_w \in [0,10]$ \ms{} \\ 
            \bottomrule
        \end{tabular}}
    \vspace{-3mm}
\end{table}

\begin{figure*}
    \centering
    \includegraphics[width=0.95\textwidth,trim={0 6cm 19.8cm 0},clip]{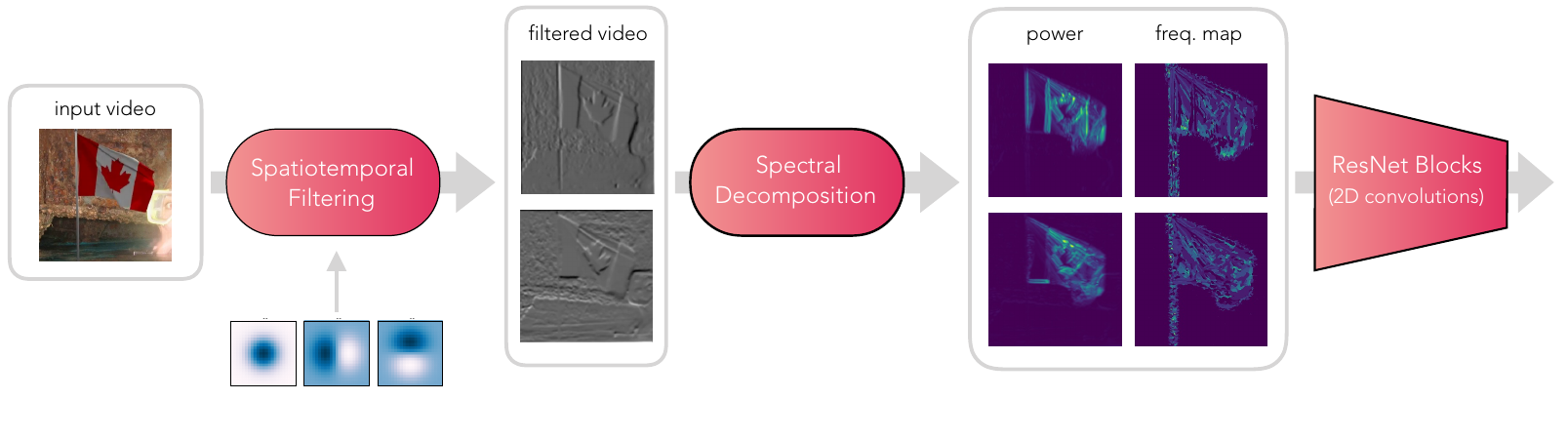}
    \vspace{-3mm}
    \caption{Overview of our SDN architecture $s_{\phi}(\*x)$ for learning the physical correspondence between the simulation and real-world observation of dynamic flags. Given a 3D video volume as input, we first apply a $0^{\text{th}}$-order temporal Gaussian filter followed by two directional $1^{\text{st}}$-order Gaussian derivative filters and then spatially subsample both filtered video volumes by a factor two. The proposed spectral decomposition layer then applies the Fourier transform and selects the maximum power and corresponding frequencies densely for all spatial locations. This produces 2D multi-channel feature maps which we process with 2D ResNet blocks to learn the embedding. \label{fig:network-architecture}}
    \vspace{-3mm}
\end{figure*}

\subsection{Spectral Decomposition Network}
\label{subsec:spectral-decomposition-network}

The dominant source of variation is in the geometry of the waves in cloth rather than in its texture. Therefore, we seek a perceptual model that can encode the cloth's dynamics such as high-frequent streamwise waves, the number of nodes in the fabric, violent flapping at the trailing edge, rolling motion of the corners and its silhouette \cite{shelley2011flapping,taneda1968waving,eloy2008aeroelastic}. As our goal is to measure sim-to-sim and sim-to-real similarity, a crucial underpinning is that our embedding function is able to disentangle and extract the relevant signal for domain adaptation \cite{peng2018visda,james2019sim}. Therefore, we propose modeling the \emph{spatial distribution of temporal spectral power} over the cloth's surface. Together with direction awareness, this effectively characterizes the traveling waves and flapping behavior from visual observations.

\vspace{2mm}

\noindent \textbf{Spectral Decomposition Layer.} The proposed solution is a novel spectral decomposition layer that distills temporal frequencies from a video. Specifically, similar to \cite{runia2019repetition}, we treat an input video volume as a collection of signals for each spatial position (\ie $H\times W$ signals) and map the signals into the frequency domain using the Discrete Fourier Transform (DFT) to estimate the videos' spatial distribution of temporal spectral power. The DFT maps a signal $f[n]$ for $n \in [0, N_t-1]$ into the frequency domain \cite{oppenheim1999discrete} as formalized by:
\begin{equation}
    F(j\omega) = \sum_{n=0}^{N_t-1} f[n] e^{-j \omega n T}.
    \label{eq:DFT}
\end{equation}
We proceed by mapping the DFT's complex output to a real-valued representation. The periodogram of a signal is a representation of its spectral power and is defined as $I({\omega}) = \frac{1}{N_t} \vert F(j\omega) \vert^2$ with $F(j \omega)$ as defined in \eqref{eq:DFT}. This provides the spectral power magnitude at each sampled frequency. To effectively reduce the dimensionality and emphasize on the videos' discriminative frequencies, we select the top-$k$ strongest frequencies and corresponding spectral power from the periodogram. Given a signal of arbitrary length, this produces $k$ pairs containing $I(\omega_{\max_i})$ and $\omega_{\max_i}$ for $i \in [0,k]$ yielding a total of $2k$ scalar values.

Considering an input video volume, treated as a collection of $H \times W$ signals of length $N_t$, the procedure extracts the discriminative frequency and its corresponding power at each spatial position. In other words, the spectral decomposition layer performs the mapping $\mathbb{R}^{C \times N_t \times H\times W } \rightarrow \mathbb{R}^{2kC \times H \times W}$. The videos' temporal dimension is squeezed and the result can be considered a multi-channel feature map -- to be further processed by any 2D convolutional layer. We reduce spectral leakage using a Hanning window before applying the DFT. The batched version of the proposed layer is formalized as algorithm in the supplementary material.

\newpage

\noindent \textbf{Embedding Function}. The specification of $s_{\phi}(\*x)$, with the spectral decomposition layer at its core, is illustrated in \Cref{fig:network-architecture}. First, our model convolves the input video $\*x$ with a temporal Gaussian filter followed by two spatially oriented first-order derivative filters. Both resulting video volumes are two-times spatially subsampled by means of max-pooling. Successively, the filtered video representations are fed through the spectral decomposition layer to produce spectral power and frequency maps. The outputs are stacked into a multi-channel feature map to be further processed by a number of 2D convolutional filters with trainable weights $\phi$. We use $3$ standard ResNet blocks \cite{he2016deep} and a final linear layer that maps to the $\mathbb{R}^{D_e}$ embedding space. We refer to our network as \emph{Spectral Decomposition Network (SDN)}.

\vspace{2mm}

\noindent \textbf{Network Details.} Our network is implemented in PyTorch \cite{paszke2017automatic} and is publicly available\footnote{\url{https://tomrunia.github.io/projects/cloth/}}. Unless mentioned otherwise, all network inputs are temporally sampled at $25$ fps. After that, we use a temporal Gaussian with $\sigma_t = 1$ and first-order Gaussian derivative filters with $\sigma_{x,y} = 2$. For training the embedding function with the contrastive loss, we adopt a margin of $1$ and use the \emph{BatchAll} sampling strategy \cite{hermans2017defense,ding2015deep}. The spectral decomposition layer selects the single most discriminative frequency (\ie $k=1$). Adding secondary frequency peaks to the feature maps did not yield substantial performance gains. The size of our embeddings is fixed ($D_e = 512$) for the paper. Input video clips of size $224 \times 224$ are converted to grayscale. We optimize the weights of the trainable ResNet blocks using Adam \cite{kingma2015adam} with mini-batches of $32$, learning rate $10^{-2}$ and a weight decay of $2\cdot 10^{-3}$.

\begin{figure}[b]
    \centering
    \begin{minipage}{0.2\columnwidth}
    \centering
        \includegraphics[width=\textwidth,trim={1.15cm 0 1.15cm 0},clip]{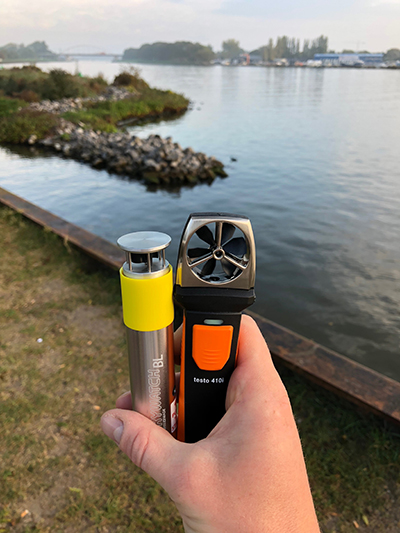}
    \end{minipage}
    \begin{minipage}{0.79\columnwidth}
        \begin{subfigure}{.19\columnwidth}
            \centering
            \includegraphics[width=\textwidth]{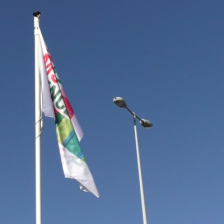}
        \end{subfigure}
        \begin{subfigure}{.19\columnwidth}
            \centering
            \includegraphics[width=\textwidth]{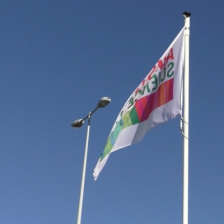}
        \end{subfigure}
        \begin{subfigure}{.19\columnwidth}
            \centering
            \includegraphics[width=\textwidth]{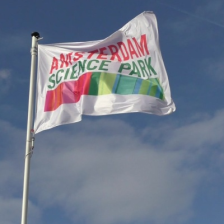}
        \end{subfigure}
        \begin{subfigure}{.19\columnwidth}
            \centering
            \includegraphics[width=\textwidth]{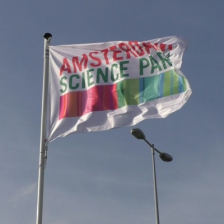}
        \end{subfigure}
        \begin{subfigure}{.19\columnwidth}
            \centering
            \includegraphics[width=\textwidth]{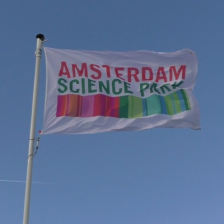}
        \end{subfigure}
        \\[1mm]
        \begin{subfigure}{.19\columnwidth}
            \centering
            \includegraphics[width=\textwidth]{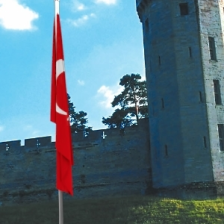}
        \end{subfigure}
        \begin{subfigure}{.19\columnwidth}
            \centering
            \includegraphics[width=\textwidth]{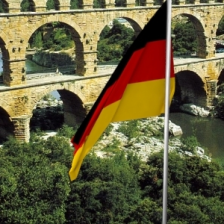}
        \end{subfigure}
        \begin{subfigure}{.19\columnwidth}
            \centering
            \includegraphics[width=\textwidth]{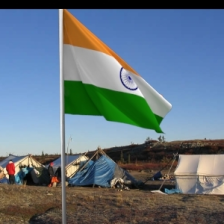}
        \end{subfigure}
        \begin{subfigure}{.19\columnwidth}
            \centering
            \includegraphics[width=\textwidth]{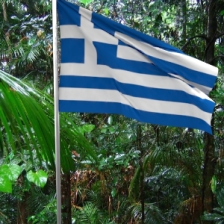}
        \end{subfigure}
        \begin{subfigure}{.19\columnwidth}
            \centering
            \includegraphics[width=\textwidth]{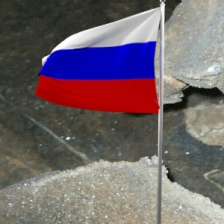}
        \end{subfigure}
    \end{minipage}
    \vspace{-2mm}
    \caption{\emph{Left:} Two anemometers used for gauging the wind speed. \emph{Right top:} Real flag recordings with corresponding wind speeds measured by the anemometer hoisted in the flagpole. \emph{Right bottom:} simulated examples from our FlagSim dataset.} \label{fig:flag-datset-examples}
\end{figure}

\section{Real and Simulated Datasets}
\label{sec:datasets}

\noindent \textbf{Real-world Flag Videos.} To evaluate our method's ability to infer physical parameters from real-world observations, we have set out to collect video recordings of real-world flags with ground-truth wind speed. We used two anemometers (\Cref{fig:flag-datset-examples}) to measure the wind speed at the flag's position. After calibration and verification of the meters, we hoisted one of them in the flagpole to the center height of the flag to ensure accurate and local measurements. A Panasonic HC-V770 camera was used for video recording. In total, we have acquired more than an hour of video over the course of $5$ days in varying wind and weather conditions. We divide the dataset in $2.7${\sc K} train and $1.3${\sc K} non-overlapping test video clips and use 1-minute average wind speeds as ground-truth. The train and test video clips are recorded on different days with varying weather conditions. Examples are displayed in \Cref{fig:flag-datset-examples} and the dataset is available through our website.

\vspace{2mm}

\noindent \textbf{FlagSim Dataset.} To train the embedding function $s_{\phi}(\*x)$ as discussed in \Cref{subsec:similarity-function}, we introduce the FlagSim dataset consisting of flag simulations and their rendered animations. We simulate flags by random sampling a set of physical parameters $\btheta$ from \Cref{tab:parameter-search-space} and feed them to ArcSim. For each flag simulation, represented as sequence of 3D meshes, we use Blender \cite{blender2018} to render multiple flag animations $\*x_{\text{sim}}^i$ at different render settings $\*\zeta^i$. We position the camera at a varying distance from the flagpole and assert that the cloth surface is visible by keeping a minimum angle of $15^{\circ}$ between the wind direction and camera axis. From a collection of $12$ countries, we randomly sample a flag texture. Background images are selected from the SUN397 dataset \cite{xiao2010sun}. Each simulation produces $60$ cloth meshes at step size $\Delta T = 0.04$~\si{\second} (\ie 25 fps) which we render at $300\times 300$ resolution. Following this procedure, we generate $1,000$ mesh sequences and render a total of $14,000$ training examples. We additionally generate validation and test sets of $150/3,800$ and $85/3,500$ mesh sequences/renders respectively. Some examples are visualized in \Cref{fig:flag-datset-examples}. %

\begin{table}
    \centering
    \caption{External wind speed prediction from real-world flag observations on the dataset of Cardona \etal \cite{cardona2019seeing}. We regress the wind speed ($v_w \in \btheta_e$) in the range $0$~\ms{} to $15.5$~\ms{} and report numbers on the evaluation set. \label{tab:cardona-wind-speed-regression} }
    \ra{1.1}
    \vspace{-2mm}
    \small
    \begin{tabular}{lccc}
        \toprule
        Model & Input Modality & RMSE $\downarrow$ & Acc@$0.5$ $\uparrow$ \\ 
        \midrule
        Cardona \etal \cite{cardona2019seeing} & $30\times 224 \times 224$ & $1.458$ & $0.301$ \\ 
        ResNet-18 & $\hphantom{0}1\times 224 \times 224$ & $1.390$ & $0.274$ \\
        ResNet-18 & $10 \times 224 \times 224$  & $1.237$ & $0.314$ \\
        ResNet-18 & $20 \times 224 \times 224$  & $1.347$ & $0.296$ \\
        \midrule
        SDN (ours) & $30\times 224 \times 224$ & $\mathbf{1.179}$ & $\mathbf{0.337}$ \\
        \bottomrule
    \end{tabular}
    \vspace{-4mm}
\end{table}

\section{Results and Discussion}
\label{sec:experiments}

\noindent \textbf{Real-world Extrinsic Wind Speed Measurement ($\btheta_e$).} We first assess the effectiveness of the proposed spectral decomposition network by measuring the wind speed on the recently proposed real-world flag dataset by Cardona \etal \cite{cardona2019seeing}. Their method, consisting of an ImageNet-pretrained ResNet-18 \cite{he2016deep} with LSTM, will be the main comparison. We also train ResNet-18 with multiple input frames, followed by temporal average pooling of the final activations \cite{karpathy2014large}. After training all methods, we report the root mean squared error (RMSE) and accuracy within $0.5$ \ms{} (Acc@0.5) in \Cref{tab:cardona-wind-speed-regression}. While our method has significantly fewer parameters ($2.6${\sc M} versus $11.2${\sc M} and $42.1${\sc M}), the SDN outperforms the existing work on the task of real-world wind speed regression. This indicates the SDN's effectiveness in modeling the spatial distribution of spectral power over the cloth's surface and its descriptiveness for the task at hand. The supplementary material contains the results on our FlagSim dataset.

\vspace{2mm}

\begin{figure}
	\centering
    \includegraphics[width=1.0\columnwidth]{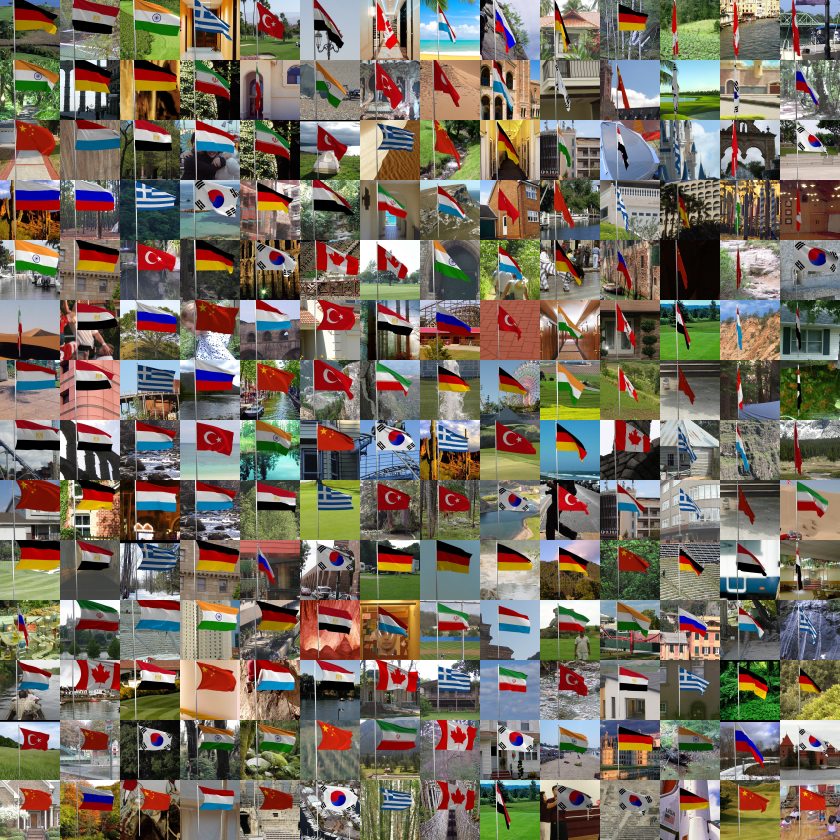}
    \vspace{-6mm}
    \caption{Barnes-Hut t-SNE \cite{van2014accelerating} visualization of the learned flag embedding space. For visualization purpose we only display examples with wind from the left. Top-right examples exhibit flags at low wind speeds while bottom-left corresponds to strong winds.} \label{fig:tsne-embeddings}
    \vspace{-4mm}
\end{figure}

\begin{table}[b]
    \centering
    \ra{1.2}
    \caption{Evaluation of our physical similarity $s_\phi(\*x)$ for FlagSim test examples. We report average triplet accuracies \cite{veit2017conditional}. \label{tab:triplet-accuracies}}
    \vspace{-2mm}
    \small
    \begin{tabular}{llllll}
        \toprule
        Input Frames & $10$ & $20$ & $30$ & $40$ & $50$ \\
        \midrule
        FlagSim Accuracy & $89.3$ & $92.1$ & $\mathbf{96.3}$ & $90.1$ & $92.4$ \\
        \bottomrule
    \end{tabular}
\end{table}

\noindent \textbf{SDN's Physical Similarity Quality ($\btheta_i, \btheta_e$).} We evaluate the physical similarity embeddings after training with contrastive loss. To quantify the ability to separate examples with similar and dissimilar physical parameters, we report the triplet accuracy \cite{veit2017conditional}. We construct $3.5$\si{K} FlagSim triplets from the test set as described in \Cref{subsec:similarity-function}. We consider the SDN trained for video clips of a varying number of input frames and report its accuracies in \Cref{tab:triplet-accuracies}. The results indicate the effectiveness of the learned distance metric to quantify the physical similarity between different observations. When considering flags, we conclude that $30$ input frames are best with a triplet accuracy of $96.3\%$ and therefore use $30$ input frames in the remainder of this paper. In \Cref{fig:tsne-embeddings} we visualize a subset of the embedding space and observe that the flag instances with low wind speeds are clustered in the top-right corner whereas strong wind speeds live in the bottom-left. 

\vspace{2mm}

\begin{figure*}

    \fboxsep=0mm  %
    \fboxrule=2pt %
    
    \centering
    \begin{minipage}[c]{0.85\textwidth}
        \centering
        \begin{subfigure}{.19\textwidth}
            \centering
            \fcolorbox{greencustom}{white}{\includegraphics[width=\textwidth,trim={1.2cm 1cm 0.3cm 0.4cm},clip]{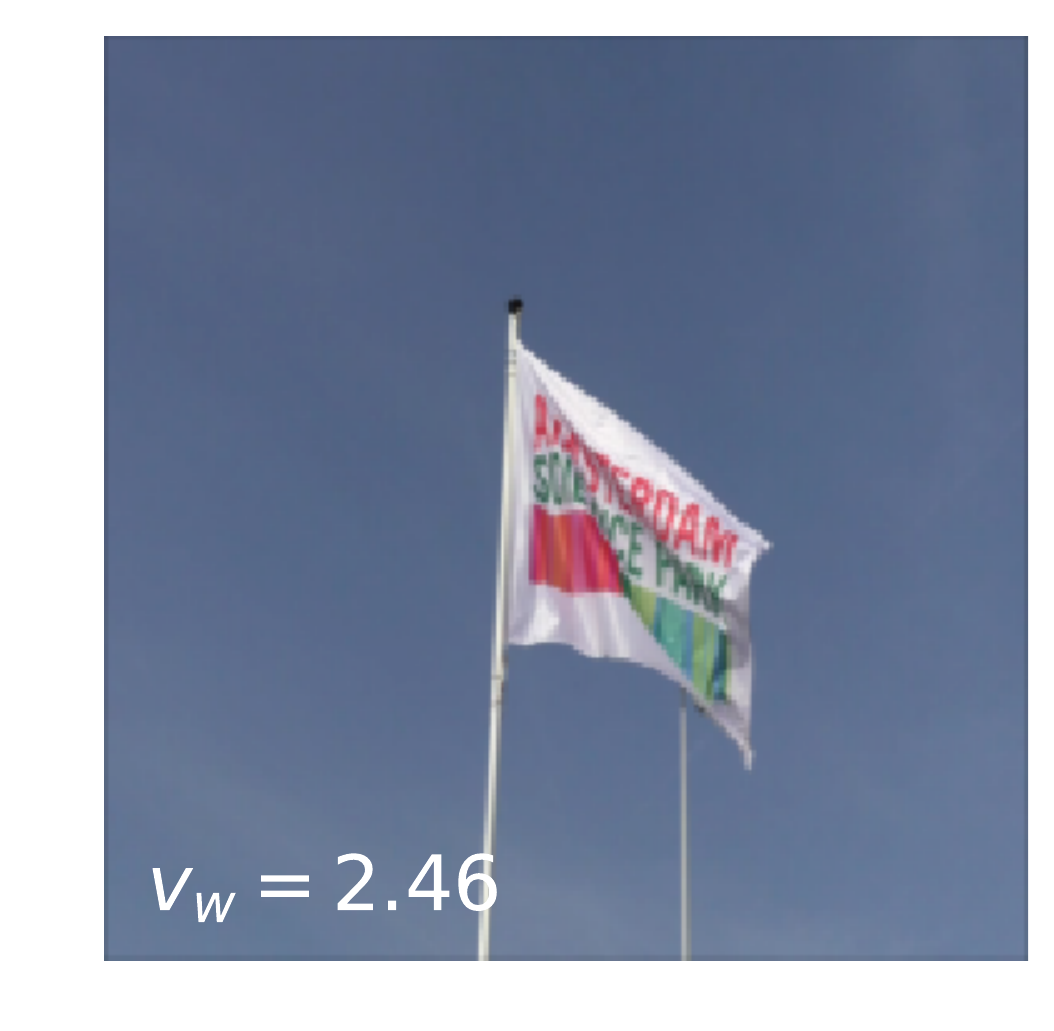}}
        \end{subfigure}
        \hfill
        \begin{subfigure}{.19\textwidth}
            \centering
            \fcolorbox{lightgray}{white}{\includegraphics[width=\textwidth,trim={1.3cm 1.1cm 0.35cm 0.8cm},clip]{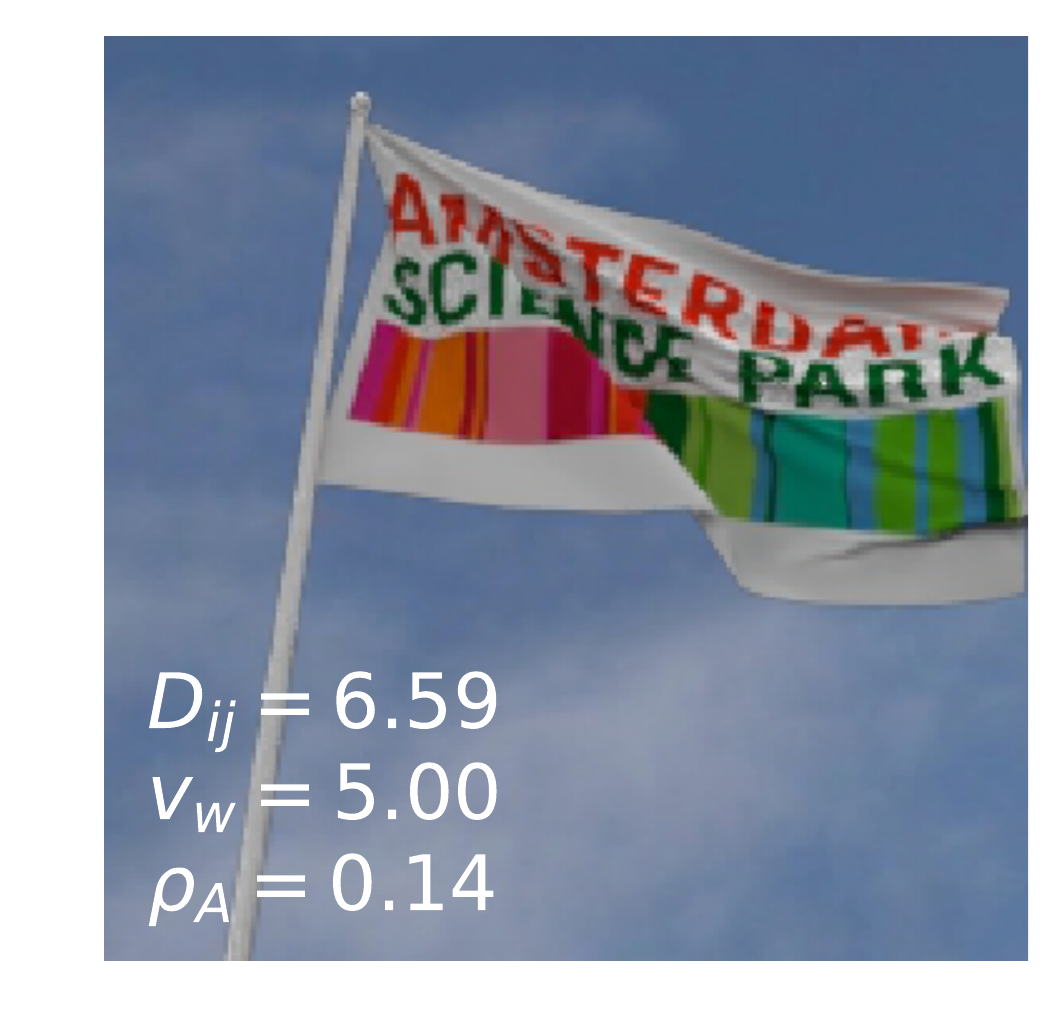}}
        \end{subfigure}
        \hfill
        \begin{subfigure}{.19\textwidth}
            \centering
            \fcolorbox{lightgray}{white}{\includegraphics[width=\textwidth,trim={1.3cm 1.1cm 0.35cm 0.8cm},clip]{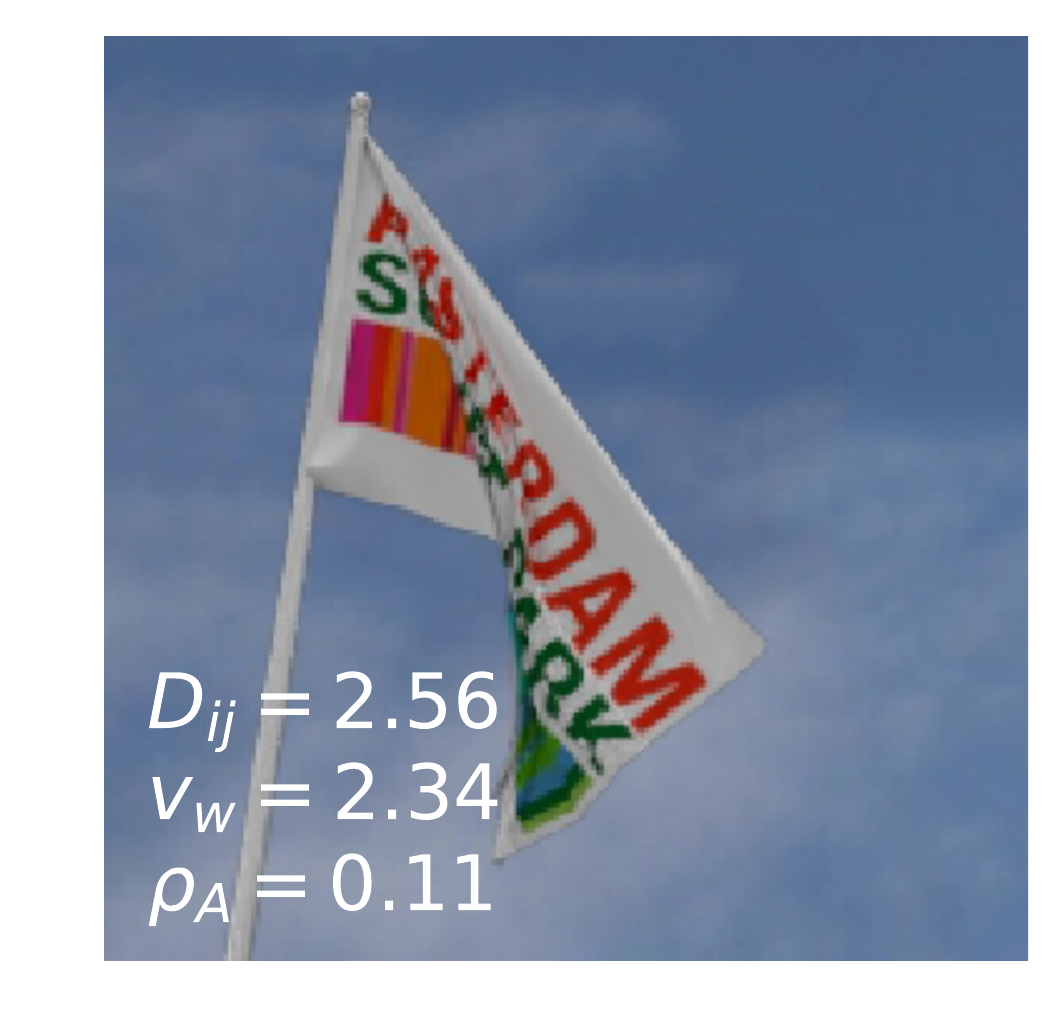}}
        \end{subfigure}
        \hfill
        \begin{subfigure}{.19\textwidth}
            \centering
            \fcolorbox{lightgray}{white}{\includegraphics[width=\textwidth,trim={1.3cm 1.1cm 0.35cm 0.8cm},clip]{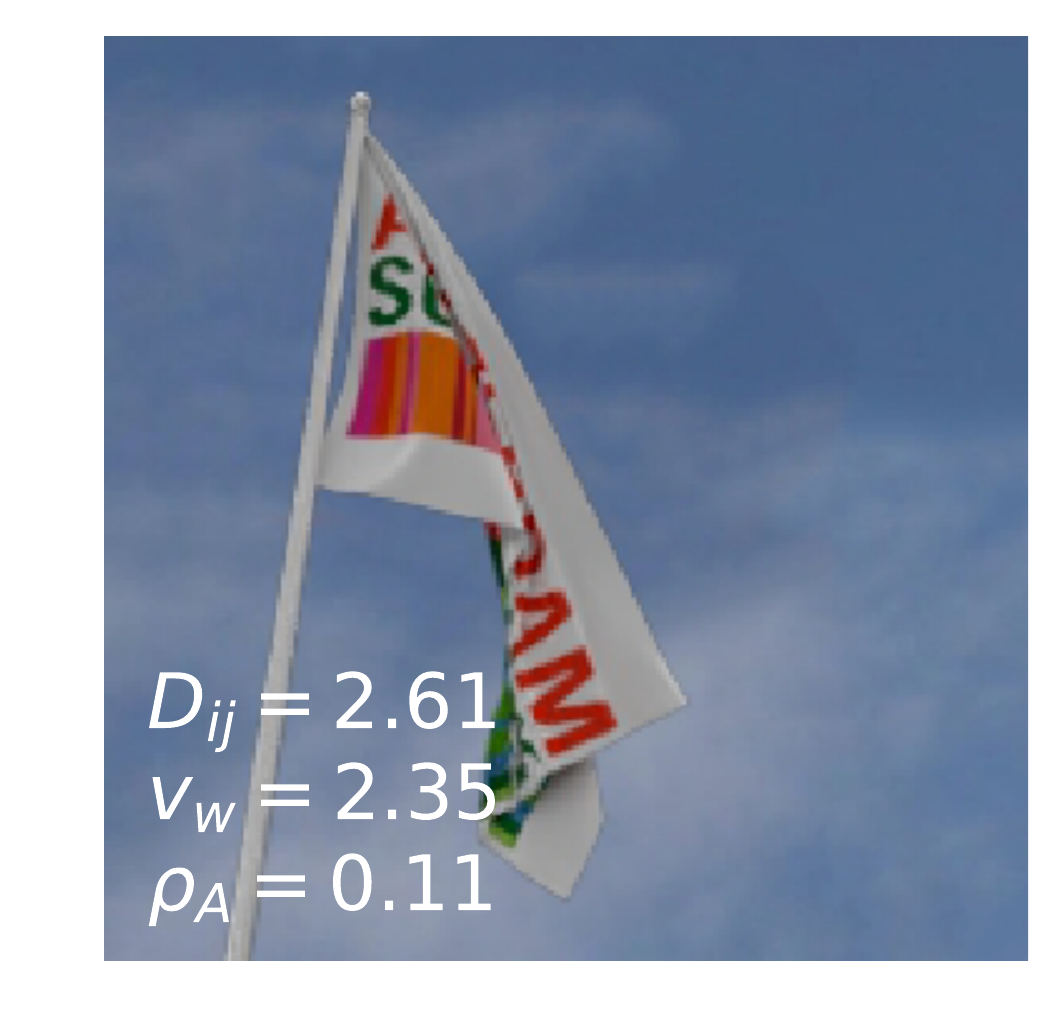}}
        \end{subfigure}
        \hfill
        \begin{subfigure}{.19\textwidth}
            \centering
            \fcolorbox{lightgray}{white}{\includegraphics[width=\textwidth,trim={1.3cm 1.1cm 0.35cm 0.8cm},clip]{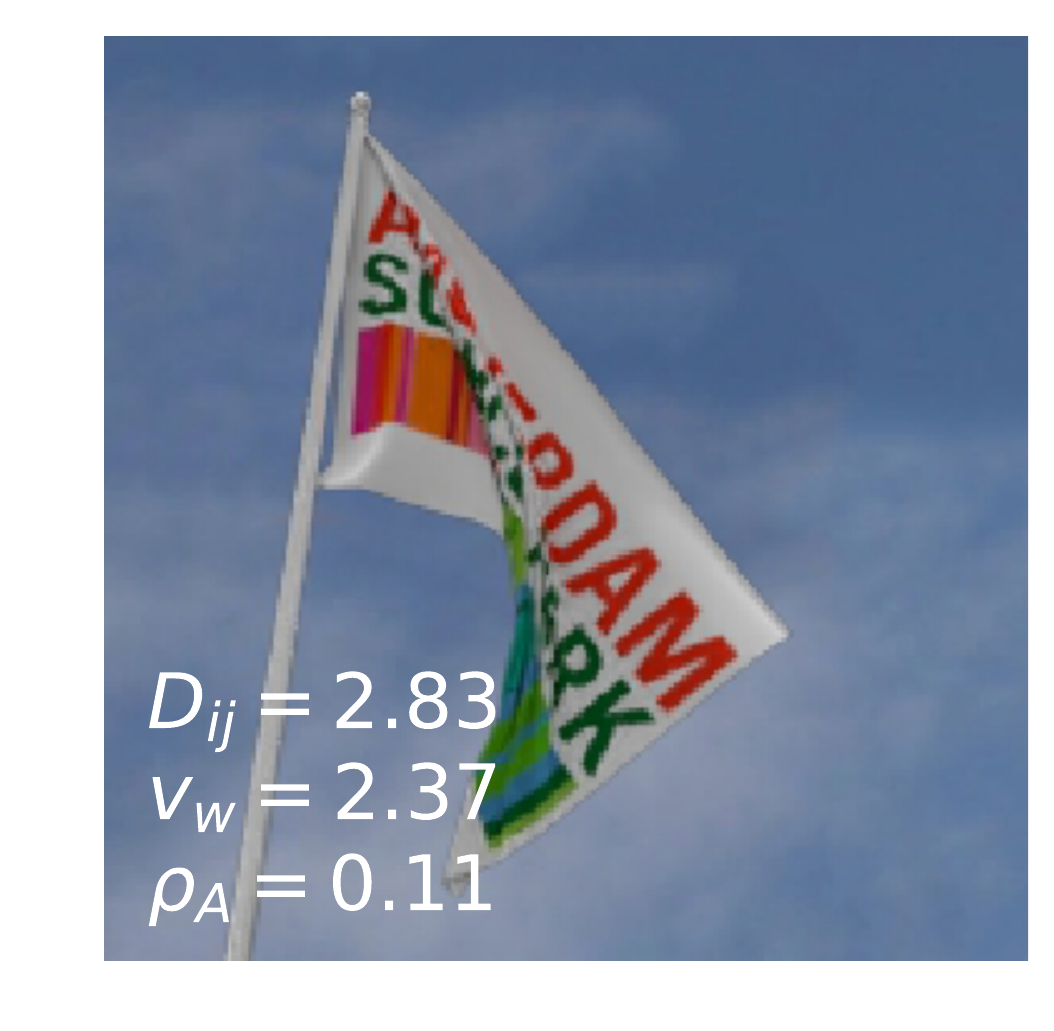}}
        \end{subfigure}
    \end{minipage}
    \\
    \begin{subfigure}{\textwidth}
        \centering
        \includegraphics[width=\textwidth]{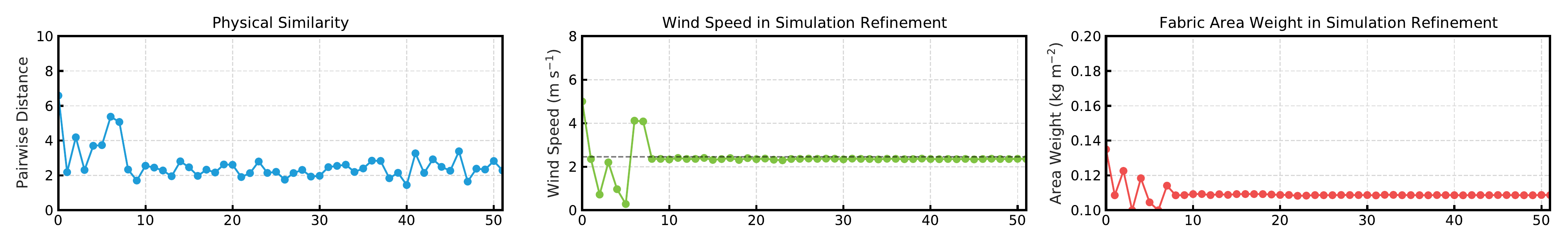}
    \end{subfigure}
    \caption{Result of our iterative measurement for a target video capturing a flag in the wind. \emph{Top left:} frame from the real-world target video clip with the ground-truth wind speed measured using an anemometer. \emph{Top remaining:} simulated examples throughout the refinement process with corresponding simulation parameters. \emph{Bottom:} development throughout the refinement process for $50$ iteration steps. We plot the distance between simulation and target instance in the embedding space and the estimated wind speed (\ms). We annotate the ground-truth wind speed with a dashed line. As the plot indicates, the refinement process converges towards the real wind speed.} \label{fig:simrefine_results}
    \vspace{-2mm}
  \end{figure*}

\begin{figure}[b]
    \centering
    \vspace{-7mm}
    \includegraphics[width=\columnwidth,trim={3mm 5mm 3mm 0},clip]{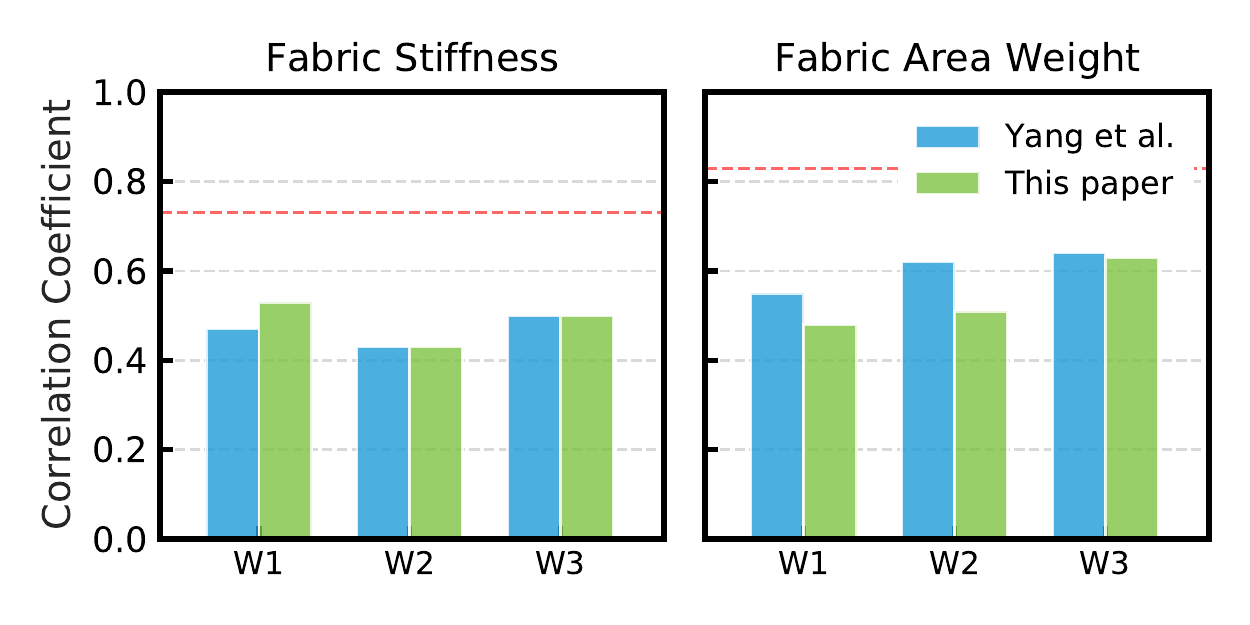}
    \vspace{-5mm}
    \caption{Intrinsic cloth material measurements from real videos. We report the Pearson correlation coefficients (higher is better) between predicted material type and both ground-truth stiffness/density on the Bouman \etal \cite{bouman2013estimating} hanging cloth dataset. The dashed red line indicates human performance as determined by \cite{bouman2013estimating}.} \label{fig:barchart-bouman}
    \vspace{-4mm}
\end{figure}

\noindent \textbf{Real-world Intrinsic Cloth Parameter Recovery ($\btheta_i$).} In this experiment, we assess the effectiveness of our SDN for estimating intrinsic cloth material properties from a real-world video. We compare against Yang \etal \cite{yang2017learning} on the hanging cloth dataset of Bouman \etal \cite{bouman2013estimating} (\Cref{fig:flagreal-hanging-cloth}). Each of the $90$ videos shows one of $30$ cloth types hanging down while being excited by a fan at $3$ wind speeds (W1-3). The goal is to infer the cloth's stiffness and area weight. From our SDN trained on FlagSim with contrastive loss, we extract the embedding vectors for the $90$ videos and project them into a $50$-dimensional space using PCA. Then we train a linear regression model using leave-one-out following \cite{bouman2013estimating}. The results are displayed in \Cref{fig:barchart-bouman}. While not outperforming the specialized method of \cite{yang2017learning}, we find that our flag-based features generalize to intrinsic cloth material recovery. This is noteworthy, as our SDN was trained on flags of lightweight materials exhibiting predominantly horizontal motion. %

\vspace{1mm}

\noindent \textbf{Real-world Combined Parameter Refinement ($\btheta_i, \btheta_e$).} Putting everything together, our goal is measuring physics parameters based on real-world observations. We demonstrate the full measurement procedure (\Cref{fig:method-overview}) by optimizing over intrinsic and extrinsic model parameters $(\btheta_i, \btheta_e)$ from real-world flag videos (and present hanging cloth refinement results in the supplementary material). First, we randomly sample a real-world flag recording as subject of the measurement. The parameter range of the intrinsic ($16\times$) and extrinsic ($1\times$) is normalized to the domain $[-1,+1]$ and are all initialized to $0$, \ie their center values. We fix the render parameters $\*\zeta$ manually as our focus is not on inferring those from real-video. However, these parameters are not carefully determined as the residual blocks in the embedding function can handle such variation (\Cref{fig:tsne-embeddings}). In each step, we simulate the cloth meshes with current parameters $\btheta_i, \btheta_e$ and render its video clip with fixed render parameters $\*\zeta$. Both the simulation and real-world video clips are then projected onto the embedding space using $s_\phi(\*x)$, and we compute their pairwise distance \eqref{eq:distance-function}. Finally, the Bayesian optimization's acquisition function (\Cref{subsec:parameter-optimization}) determines where to make the next evaluation $\btheta_i, \btheta_e \in [-1,+1]$ to maximize the expected improvement, \ie improving the measurement. The next iteration starts by denormalizing the parameters and running the simulation. We run the algorithm for $50$ refinement steps. In \Cref{fig:simrefine_results}, we demonstrate our method's measurements throughout optimization. Most importantly, we observe a gradual decrease in the pairwise distance between simulation and real-world example, indicating a successful measurement of the physical parameters. Importantly, we note that the wind speed converges towards the ground-truth wind speed within a few iterations, as indicated with a dashed line. More examples are given in the supplementary material.

\section{Conclusion}
\label{sec:conclusion}

We have presented a method for measuring intrinsic and extrinsic physical parameters for cloth in the wind without perceiving real cloth before. The iterative measurement gradually improves by assessing the similarity between the current cloth simulation and the real-world observation. By leveraging only simulations, we have proposed a method to train a physical similarity function. This enables measuring the physical correspondence between real and simulated data. To encode cloth dynamics, we have introduced a spectral decomposition layer that extracts the relevant features from the signal and generalizes from simulation to real observations. We compare the proposed method to prior work that considers flags in the wind and hanging cloth and obtain favorable results. For future work, given an appropriate physical embedding function, our method could be considered for other physical phenomena such as fire, smoke, fluid or mechanical problems. \\

\ifcvprfinal
    \noindent \textbf{Acknowledgements.} We would like to thank Rik Holsheimer for his support with the real-world flag dataset acquisition. Changyong Oh helped setting up the Bayesian optimization code and William Thong provided code for the contrastive loss and the sampling.
\fi

{\small
  \bibliographystyle{ieee_fullname}
  \bibliography{bibliography}
}

\clearpage

\appendix

\twocolumn[  
    \begin{@twocolumnfalse}
        \begin{center}

             {\Large \bf Supplementary Material}
             \vspace{10mm}

         \end{center}
     \end{@twocolumnfalse}
]

\section{Table of Contents}
\label{sec:supp-document-overview}

\noindent \emph{The supplementary material has the following content:}
\vspace{1mm}
\begin{itemize}

  \item \textbf{\Cref{sec:supplementary-spectral-decomposition-layer}} and \textbf{\Cref{alg:supplementary-spectral-decomposition-layer}}: Formalization of the batched version of the spectral decomposition layer. 

  \item \textbf{\Cref{sec:supp-dataset-acquisition}} and \textbf{\Cref{fig:supp-wind-speed-measurements}}: Details on the collection of real wind speed measurements and video recordings.
  
  \item \textbf{\Cref{sec:supp-training-details}}: Details on the video data augmentation and optimization for the training of all networks. 
  
  \item \textbf{\Cref{sec:supp-experiments-hanging-cloth}} and \textbf{\Cref{fig:supp_simrefine_bouman_results}}: Additional experiments on refined measurements for the hanging cloth dataset and details on our ClothSim dataset.
  
  \item \textbf{\Cref{tab:supplementary-flagsim-wind-strength}}: Supplement to Table~2 in the main paper with wind speed regression results on our FlagSim dataset.
  
  \item \textbf{\Cref{fig:supp_simrefine_flags_results}}: Supplement to Figure~8 in the main paper to include more refined measurement examples.

  \item \textbf{\Cref{code:json-arcsim-config}}, \textbf{\Cref{code:json-arcsim-config-cloth}} and \textbf{\Cref{code:json-material-file}}: Specification of ArcSim scene and material configuration files for flag and cloth simulations.
  
  \item \textbf{\Cref{tab:supp-blender-render-parameters}}: Exhaustive list of Blender's rendering parameters for generating the FlagSim and ClothSim datasets.
  
\end{itemize}

\section{Spectral Decomposition Layer}
\label{sec:supplementary-spectral-decomposition-layer}

To support Section~4.3 in the main paper, we formalize the batched version of the spectral decomposition layer in \Cref{alg:supplementary-spectral-decomposition-layer}. Given a batch of video clips as input, our spectral layer applies the discrete Fourier transform along the temporal dimension to compute the temporal frequency spectrum. From the periodogram, we can select the top-$k$ strongest frequency responses and their corresponding spectral power. The resulting frequency maps and power maps all have the same dimensions and can, therefore, be stacked as a multi-channel image. These tensors can be further processed by standard 2D convolution layers to learn frequency-based feature representations. The proposed layer is efficiently implemented in PyTorch \cite{paszke2017automatic} to run on the GPU using the \texttt{torch.irfft} operation. The source code is available through the project website.

\begin{algorithm}[t]
    \caption{Spectral Decomposition Layer}
    \label{alg:supplementary-spectral-decomposition-layer}
    \begin{algorithmic}[1] %

        \State \textbf{Input.} Video tensor $\*x$ of shape $[N_b, C, N_t, H, W]$
        \State \textbf{Input.} Number of frequency peaks to select, $k$
        \State \textbf{Output.} Decomposition of shape $[N_b,2kC,H,W]$
        \vspace{3mm}

        \Procedure{SpectralDecompositionLayer}{$\*x$}
            \State Reshape $\*x$ to $[N_bCHW, N_t]$ to obtain batch of signals
            \State Apply a Hanning window to signals
            \State Compute the DFT of signals using Eq.~4 (main paper)
            \State Compute periodogram of signals $I(\omega)$
            \State Select top-$k$ peaks of $I(\omega)$ and corresponding $\omega$'s
            \State $P \leftarrow$ top-$k$ peaks of $I(\omega)$ reshaped to $[N_b,kC,H,W]$
            \State $\Omega \leftarrow$ corresponding $\omega$'s reshaped to $[N_b,kC,H,W]$
            \State \textbf{return} $P, \Omega$
        \EndProcedure
    \end{algorithmic}
\end{algorithm}

\section{Real-World Flag Dataset Acquisition}
\label{sec:supp-dataset-acquisition}

\begin{figure*}[h!]
    \centering
    \begin{subfigure}{\textwidth}
        \centering
        \includegraphics[width=\textwidth]{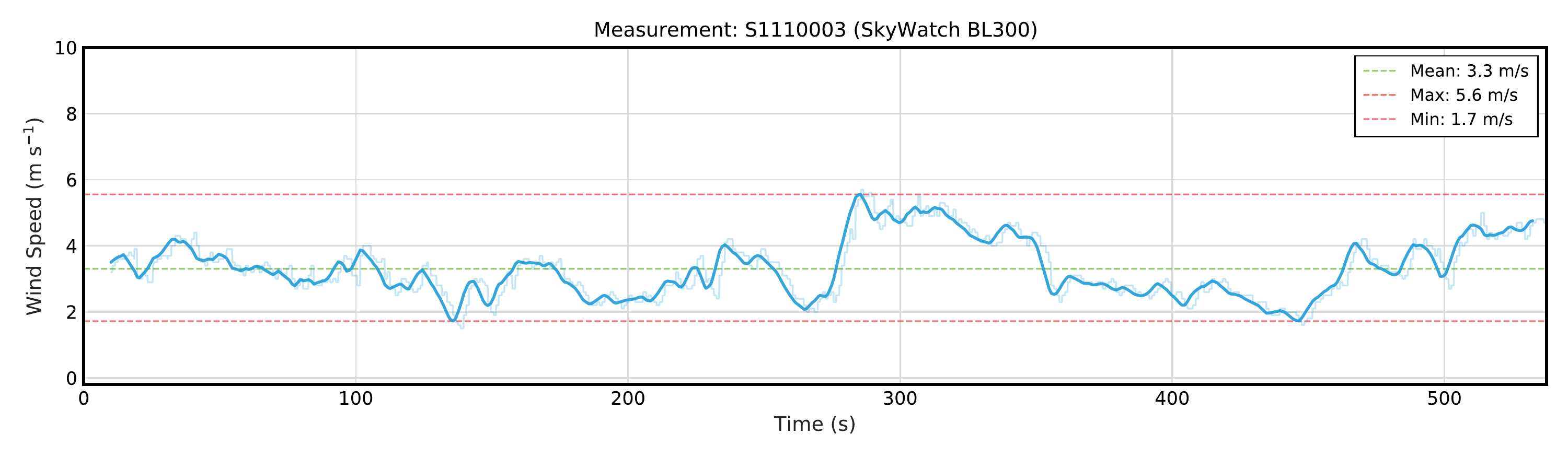}
    \end{subfigure}
    \begin{subfigure}{\textwidth}
        \centering
        \includegraphics[width=0.7\textwidth]{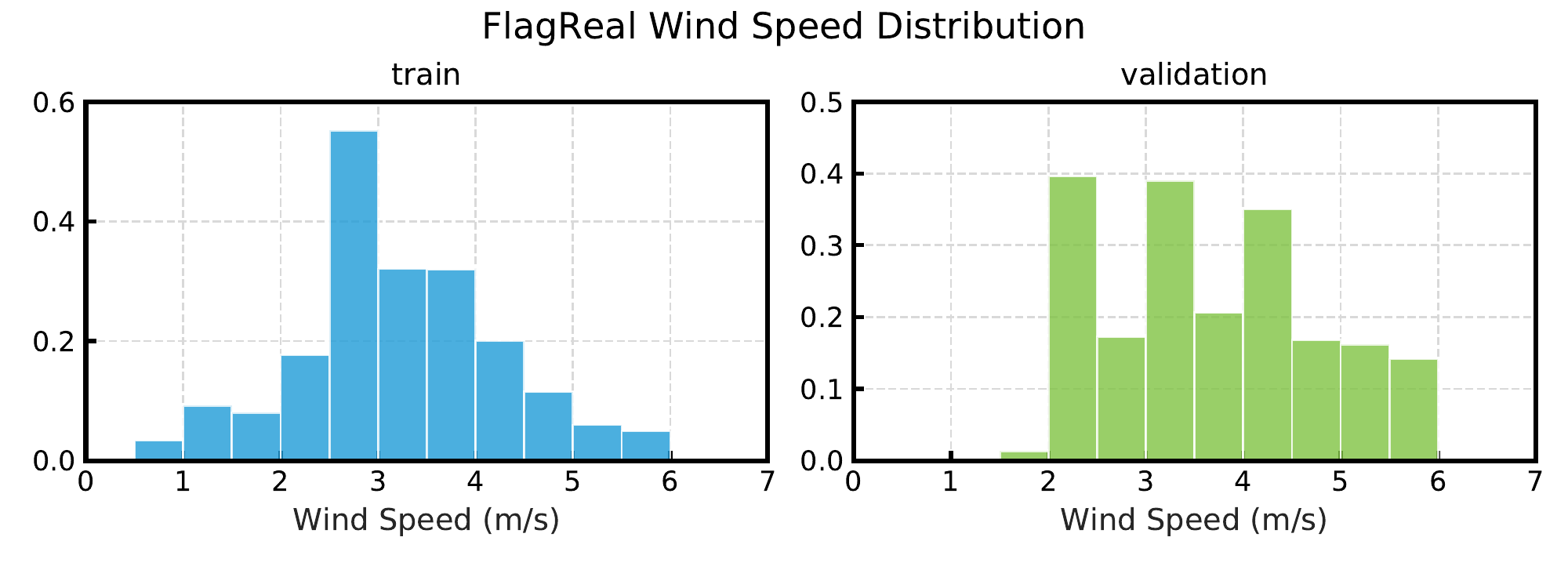}
    \end{subfigure}
    \vspace{-3mm}
    \caption{\emph{Top:} Example of the time-varying wind speed as obtained by the SkyWatch BL-400 anemometer positioned directly next to the video-recorded flag. The wind speed is sampled at $1$ \si{\hertz} and interfaced to a computer using bluetooth. For our final experiment, we sample video clips of $30$ frames and consider the ground-truth wind speed to be the average wind speed over the last minute. \emph{Bottom:} Distribution statistics of the dataset we collected. Over all $4$K non-overlapping videos the average wind speed is $3.2$\ms{} while the minimum and maximum wind speeds are $0.5$\ms{} and $6.0$\ms{} respectively.} \label{fig:supp-wind-speed-measurements}
\end{figure*}

We here describe our data acquisition to obtain real-world wind speed measurements serving as ground-truth for our final experiment. To accurately gauge the wind speed next to the flag, we have obtained two anemometers:
\begin{itemize}
 \item SkyWatch BL-400: windmill-type anemometer
 \item Testo 410i: vane-type anemometer
\end{itemize}
The measurement accuracy of both anemometers is $0.2$~\ms{}. To verify the correctness of both anemometers, we have checked that both wind meters report the same wind speeds before usage. After that, we use the \mbox{SkyWatch BL-400} anemometer for our measurements as it measures omnidirectional which is more convenient. We hoisted the anemometer in a flag pole such that the wind speeds are measured at the same height as the flag. Wind speed measurements are recorded at $1$ second intervals and interfaced to the computer. In \Cref{fig:supp-wind-speed-measurements} we display an example measurement and report the dataset's wind speed distribution. For the experiments (Section~6, main paper, last section), we randomly sample video clips of $30$ consecutive frames from our video recordings and consider the ground-truth wind speed to be the average over the last minute. This procedure ensures that small wind speed deviations and measurement errors are averaged out over time.

To capture the videos, we use a Panasonic HC-V770 video camera. The camera records at $1920\times1080$ at $60$ frames per second. We perform post-processing of the videos in the following ways. Firstly, we temporally subsample the video frames at $25$ fps such that the clips are in accordance with the frame step size in the physics simulator. Moreover, we assert that the video recordings are temporally aligned with the wind speed measurements using their timestamps. Secondly, we manually crop the videos such that the curling flag appears in the approximate center of the frame. After this, the frames are spatially subsampled to $300 \times 300$, again in agreement with animations obtained from the render engine.

\section{Training Details}
\label{sec:supp-training-details}

\noindent \textbf{Data Augmentation.} The examples in the FlagSim dataset are stored as a sequence of $60$ JPEG frames of size $300 \times 300$. During training, when using less than $60$ input frames ($30$ is used in all experiments), we randomly sample $N_t$ successive frames from each video clip. This is achieved by uniform sampling of a temporal offset within the video. After this, for the sampled sequence of frames, we convert images to grayscale, perform multi-scale random cropping and apply random horizontal flipping \cite{wang2015towards} to obtain a $N_t \times 1 \times 224 \times 224$ input clip. Finally, we subtract the mean and divide by the standard deviation for each video clip.

\vspace{3mm}

\noindent \textbf{Optimization Details.} We train all networks using stochastic gradient descent with Adam \cite{kingma2015adam}. We initialize training with a learning rate of $10^{-2}$ and decay the learning rate with a factor $10$ after $20$ epochs. To prevent overfitting, we utilize weight decay of $2\cdot 10^{-3}$ for all networks. Training continues until validation loss plateaus -- typically around $40$ epochs. Total training time for our spectral decomposition network is about $4$ hours on a single Nvidia GeForce GTX Titan X. When training the recurrent models \cite{cardona2019seeing,yang2017learning} we also perform gradient clipping (max norm of 10) to improve training stability.

\section{Experiments on Hanging Cloth Video}
\label{sec:supp-experiments-hanging-cloth}

Our real-world flag dataset enables us to evaluate our method's measurement performance of external parameters ($v_w \in \btheta_e$). However, the cloth's internal parameters are unknown and cannot be evaluated beyond visual inspection. Therefore, we also perform experiments on the hanging cloth dataset of Bouman \etal \cite{bouman2013estimating}. The authors have carefully determined the internal cloth material properties, which we can leverage for quantitative evaluation of our simulated-refined measurements. Specifically, we assess our method's ability to measure the cloth's \emph{area weight} (\kgmm{}). The method is identical to that explained in the main paper with its results presented in the final experiment of Section~6. However, we retrain the embedding function $s_{\phi}(\*x)$ on a dataset of hanging cloth simulations, which we refer to as ClothSim. In this section, we will briefly discuss the characteristics of this dataset and report experimental results.

\vspace{2mm}

\noindent \textbf{ClothSim Dataset.} Following the same procedure as for the FlagSim dataset, we additionally generate a dataset of simulated hanging cloth excited by a constant wind force. The main difference between the FlagSim dataset is the wider variety of cloth material. Specifically, we use all the materials presented in \cite{wang2011data} available in ArcSim. The increased diversity allows us to model the dynamics in real-world hanging cloth recording \cite{bouman2013estimating}. Our dataset shares similarity with the simulated hanging cloth dataset of \cite{yang2017learning}. However, in their work, the dataset is employed to train a classifier for predicting the material class. In \Cref{code:json-arcsim-config-cloth} and \Cref{tab:supp-blender-render-parameters} we present an exhaustive overview of the simulation and render parameters that were used for generating the dataset.

\vspace{2mm}

\noindent \textbf{Real-world Parameter Refinement ($\btheta_i, \btheta_e$).} Given the embedding function $s_{\phi}(\*x)$ trained on ClothSim using contrastive loss, we run our refined measurement experiment on the hanging cloth dataset of Bouman \etal \cite{bouman2013estimating}. Our goal is to measure the cloth's area weight as we have access to its ground-truth measurement. Unlike for our real-world flag dataset, we do not know the true wind speed beyond the setting of the industrial fan that was used for exciting the fabric artificially. In \Cref{fig:supp_simrefine_bouman_results} we report the results for $3$ randomly sampled real-world videos.

\begin{table}
    \centering
    \caption{External wind speed prediction from simulation. We regress the wind speed ($v_w \in \btheta_e$) on our \textbf{FlagSim dataset}. The metrics are computed over the $3.5${\sc{K}} test examples. Target velocities range from $0$~\ms{} (no wind) to $10$~\ms{} (strong wind). Experimental setup is identical to Table~2 in the main paper. \label{tab:supplementary-flagsim-wind-strength}}
    \vspace{-2mm}
    \ra{1.1}
    \small
    \begin{tabular}{lccc}
        \toprule
        Model & Input Modality & RMSE $\downarrow$ & Acc@$0.5$ $\uparrow$ \\ 
        \midrule
        Yang \etal \cite{yang2017learning} & $10\times 227 \times 227$ & $0.380$ & $0.620$ \\
        Cardona \etal \cite{cardona2019seeing} & $30\times 227 \times 227$ & $0.271$ & $0.580$ \\
        ResNet-18 & $\hphantom{0}1\times 224 \times 224$ & $0.381$ & $0.615$ \\
        ResNet-18 & $10 \times 224 \times 224$  & $0.264$ & $0.734$ \\
        ResNet-18 & $20 \times 224 \times 224$  & $0.207$ & $0.775$ \\
        \midrule
        SDN (ours) & $20\times 224 \times 224$ & $0.183$ & $0.813$ \\ 
        SDN (ours) & $30\times 224 \times 224$ & $\mathbf{0.180}$ & $\mathbf{0.838}$ \\
        \bottomrule
    \end{tabular}
    \vspace{-3mm}
\end{table}

\clearpage

\begin{figure*}[h!]
    \fboxsep=0mm  %
    \fboxrule=2pt %
    \vspace{3mm}
    \centering
    \begin{minipage}[c]{0.85\textwidth}
        \centering
        \begin{subfigure}{.19\textwidth}
            \centering
            \fcolorbox{greencustom}{white}{\includegraphics[width=\textwidth,trim={1.2cm 1cm 0.3cm 0.4cm},clip]{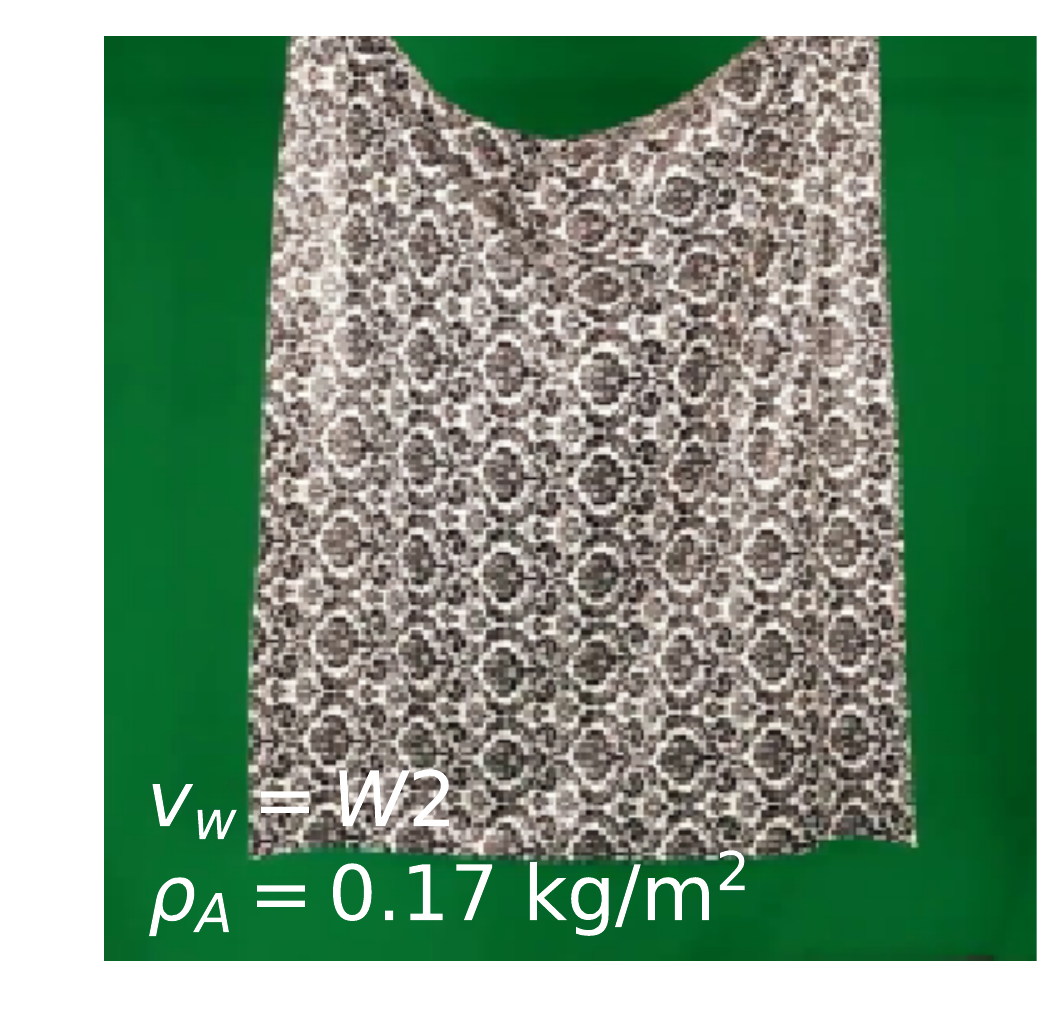}}
        \end{subfigure}
        \hfill
        \begin{subfigure}{.19\textwidth}
            \centering
            \fcolorbox{lightgray}{white}{\includegraphics[width=\textwidth,trim={1.3cm 1.1cm 0.35cm 0.8cm},clip]{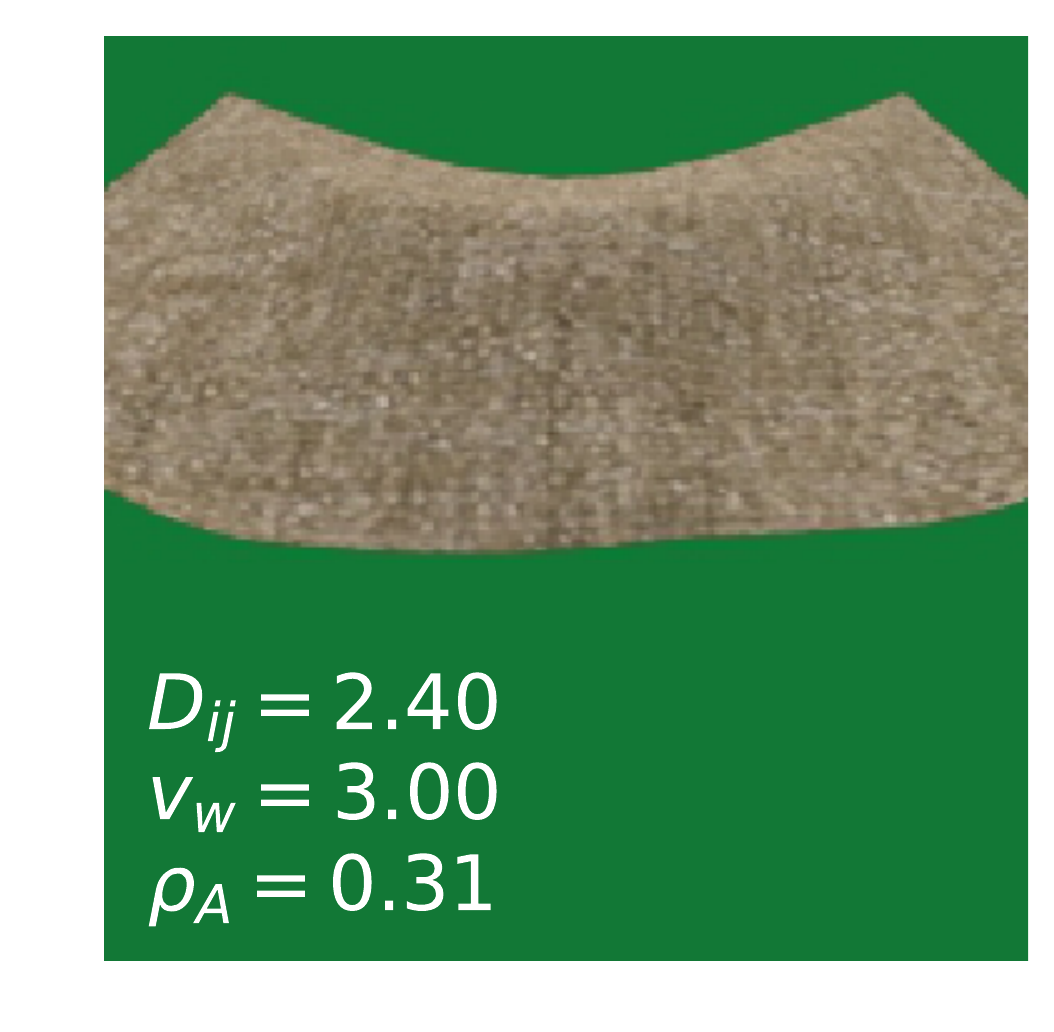}}
        \end{subfigure}
        \hfill
        \begin{subfigure}{.19\textwidth}
            \centering
            \fcolorbox{lightgray}{white}{\includegraphics[width=\textwidth,trim={1.3cm 1.1cm 0.35cm 0.8cm},clip]{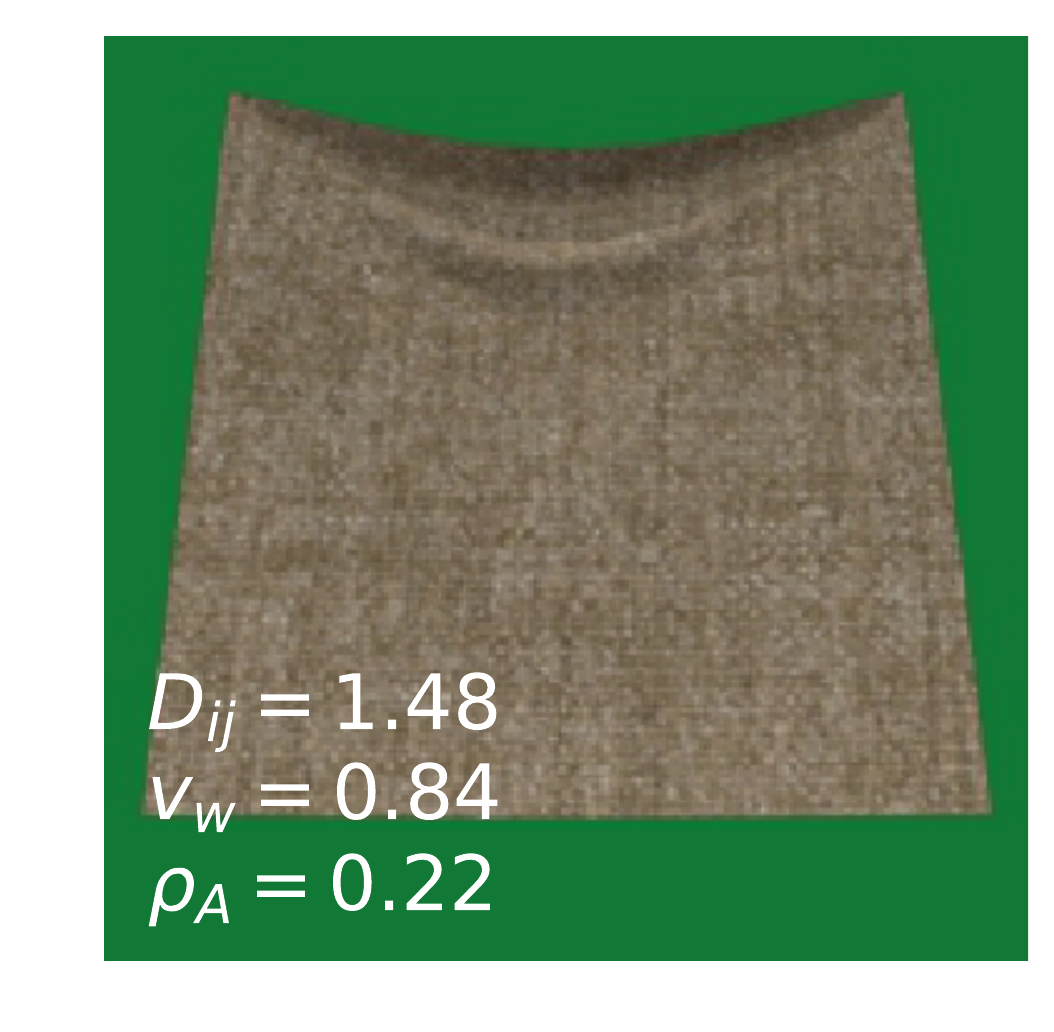}}
        \end{subfigure}
        \hfill
        \begin{subfigure}{.19\textwidth}
            \centering
            \fcolorbox{lightgray}{white}{\includegraphics[width=\textwidth,trim={1.3cm 1.1cm 0.35cm 0.8cm},clip]{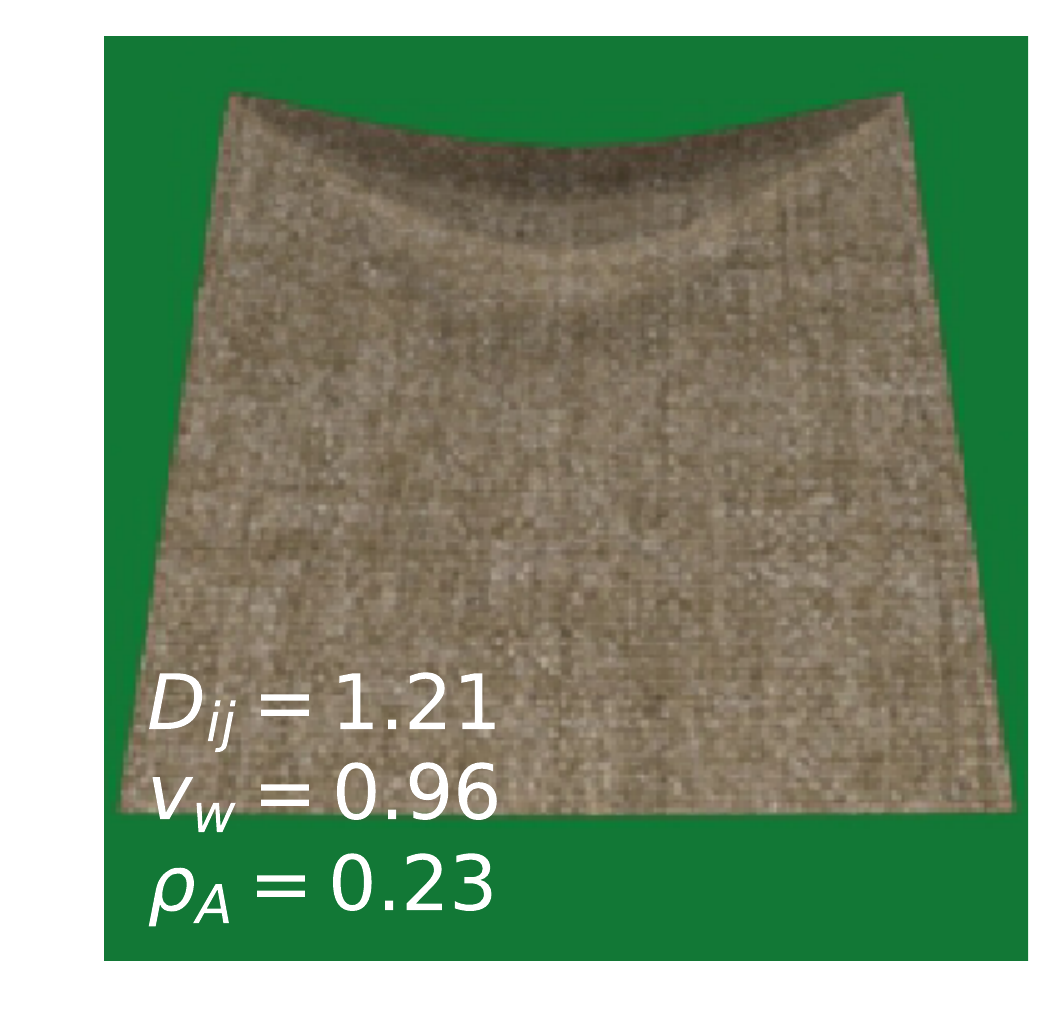}}
        \end{subfigure}
        \hfill
        \begin{subfigure}{.19\textwidth}
            \centering
            \fcolorbox{lightgray}{white}{\includegraphics[width=\textwidth,trim={1.3cm 1.1cm 0.35cm 0.8cm},clip]{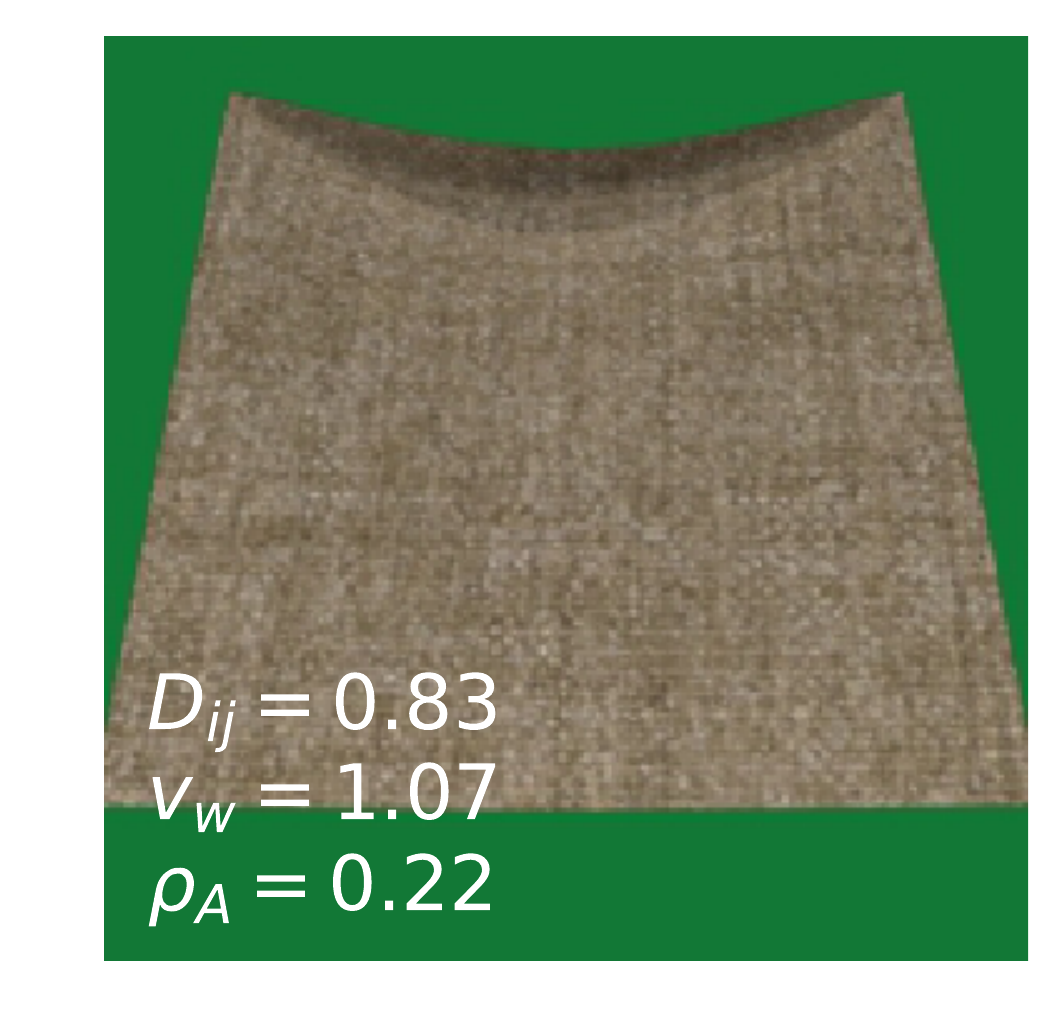}}
        \end{subfigure}
    \end{minipage}
    \\
    \begin{subfigure}{\textwidth}
        \centering
        \includegraphics[width=0.9\textwidth]{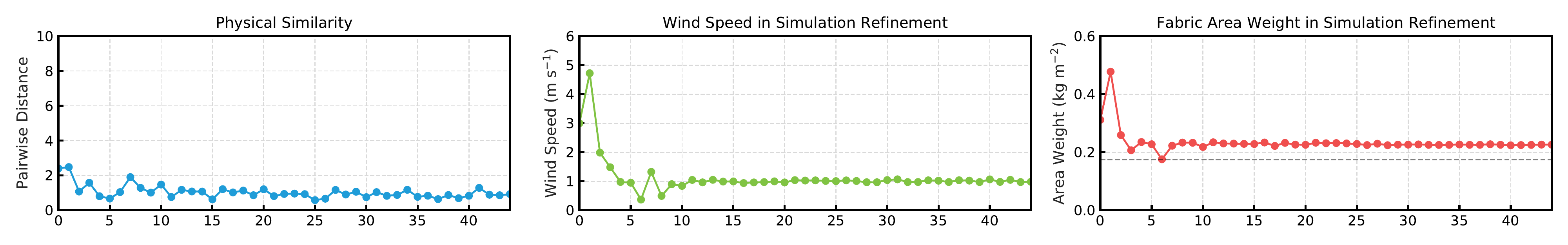}
    \end{subfigure}
  \end{figure*}
  \begin{figure*}
    \fboxsep=0mm  %
    \fboxrule=2pt %
    \centering
    \begin{minipage}[c]{0.85\textwidth}
        \centering
        \begin{subfigure}{.19\textwidth}
            \centering
            \fcolorbox{greencustom}{white}{\includegraphics[width=\textwidth,trim={1.2cm 1cm 0.3cm 0.4cm},clip]{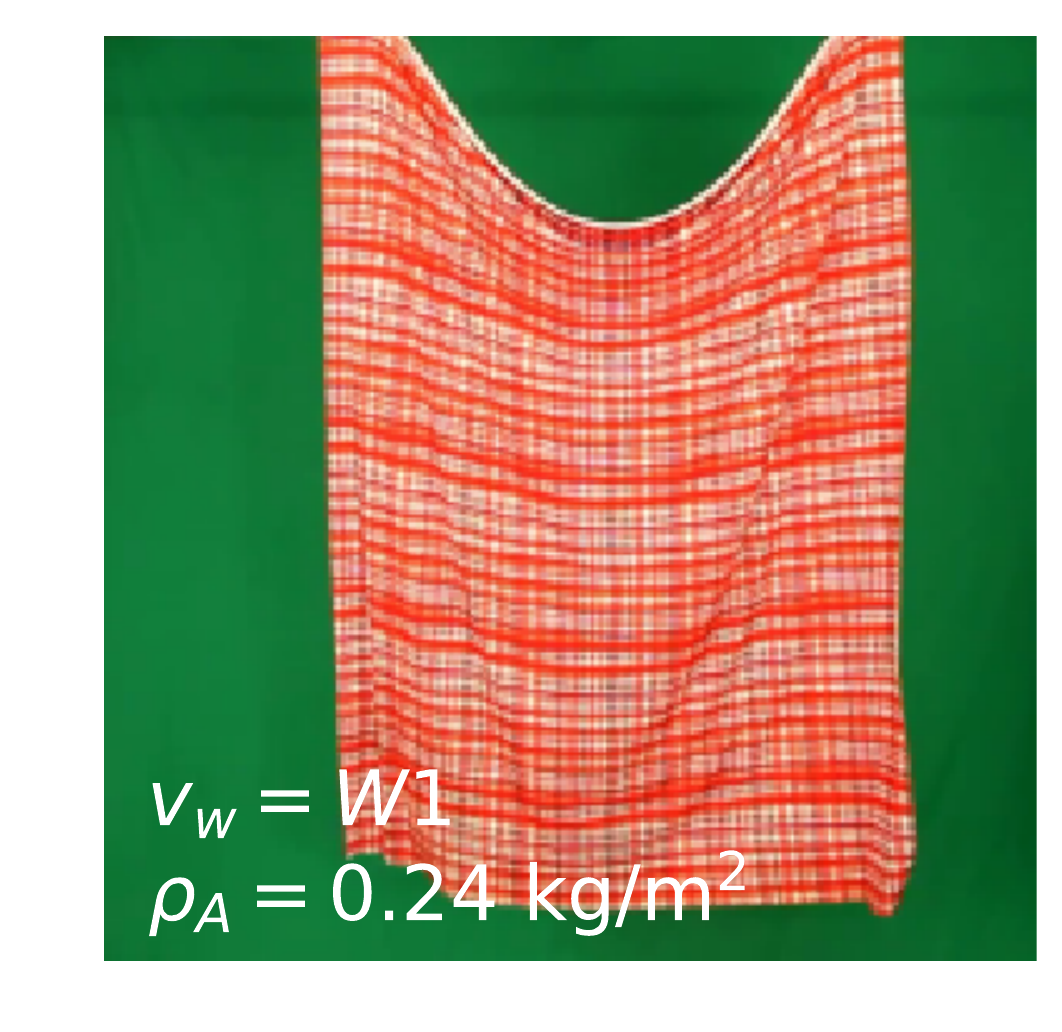}}
        \end{subfigure}
        \hfill
        \begin{subfigure}{.19\textwidth}
            \centering
            \fcolorbox{lightgray}{white}{\includegraphics[width=\textwidth,trim={1.3cm 1.1cm 0.35cm 0.8cm},clip]{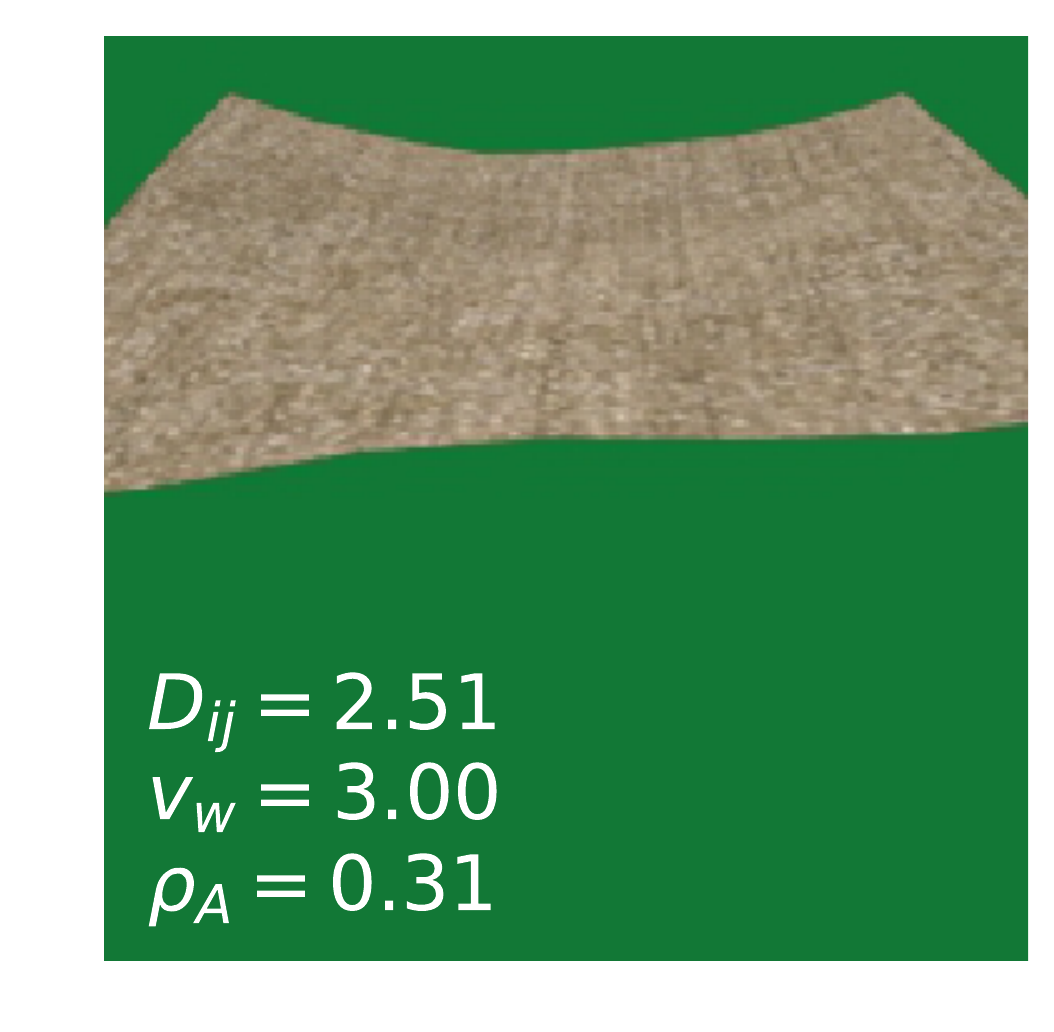}}
        \end{subfigure}
        \hfill
        \begin{subfigure}{.19\textwidth}
            \centering
            \fcolorbox{lightgray}{white}{\includegraphics[width=\textwidth,trim={1.3cm 1.1cm 0.35cm 0.8cm},clip]{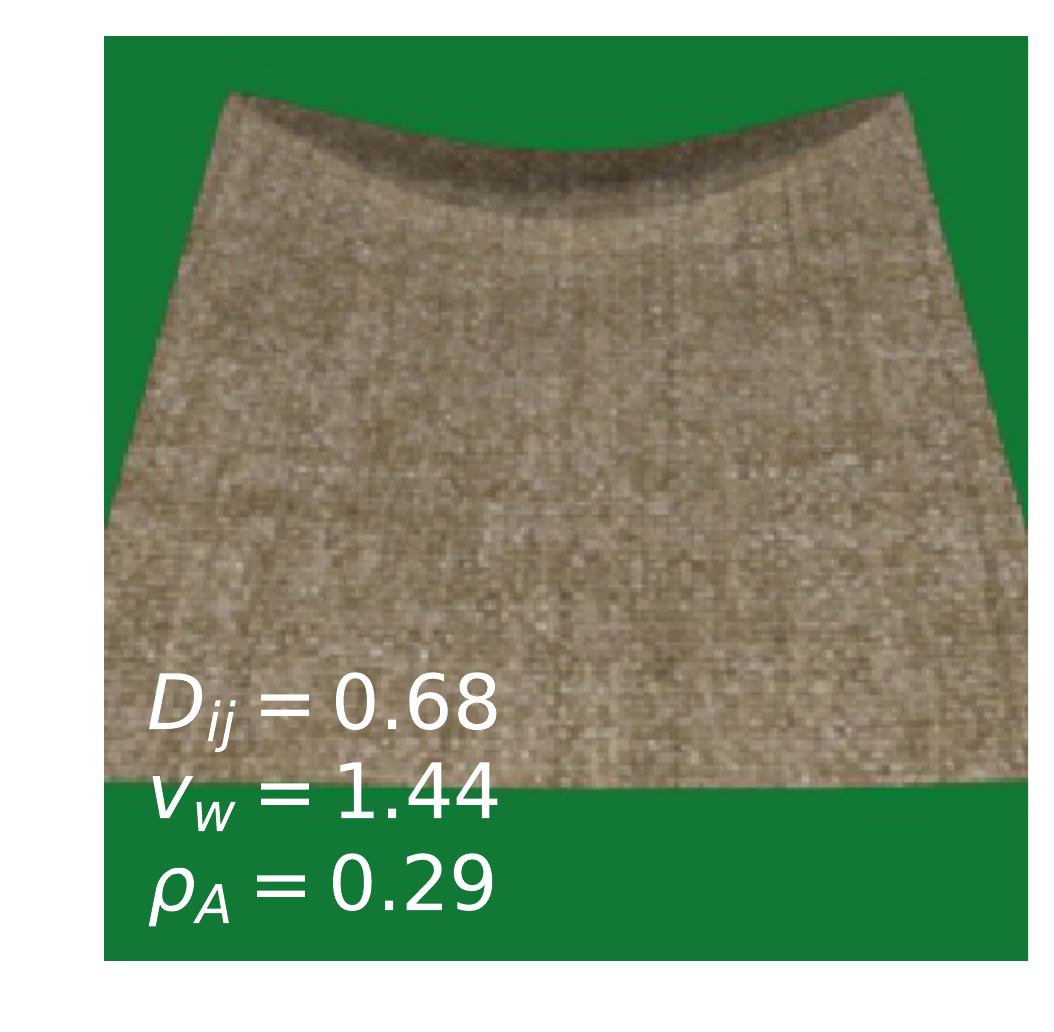}}
        \end{subfigure}
        \hfill
        \begin{subfigure}{.19\textwidth}
            \centering
            \fcolorbox{lightgray}{white}{\includegraphics[width=\textwidth,trim={1.3cm 1.1cm 0.35cm 0.8cm},clip]{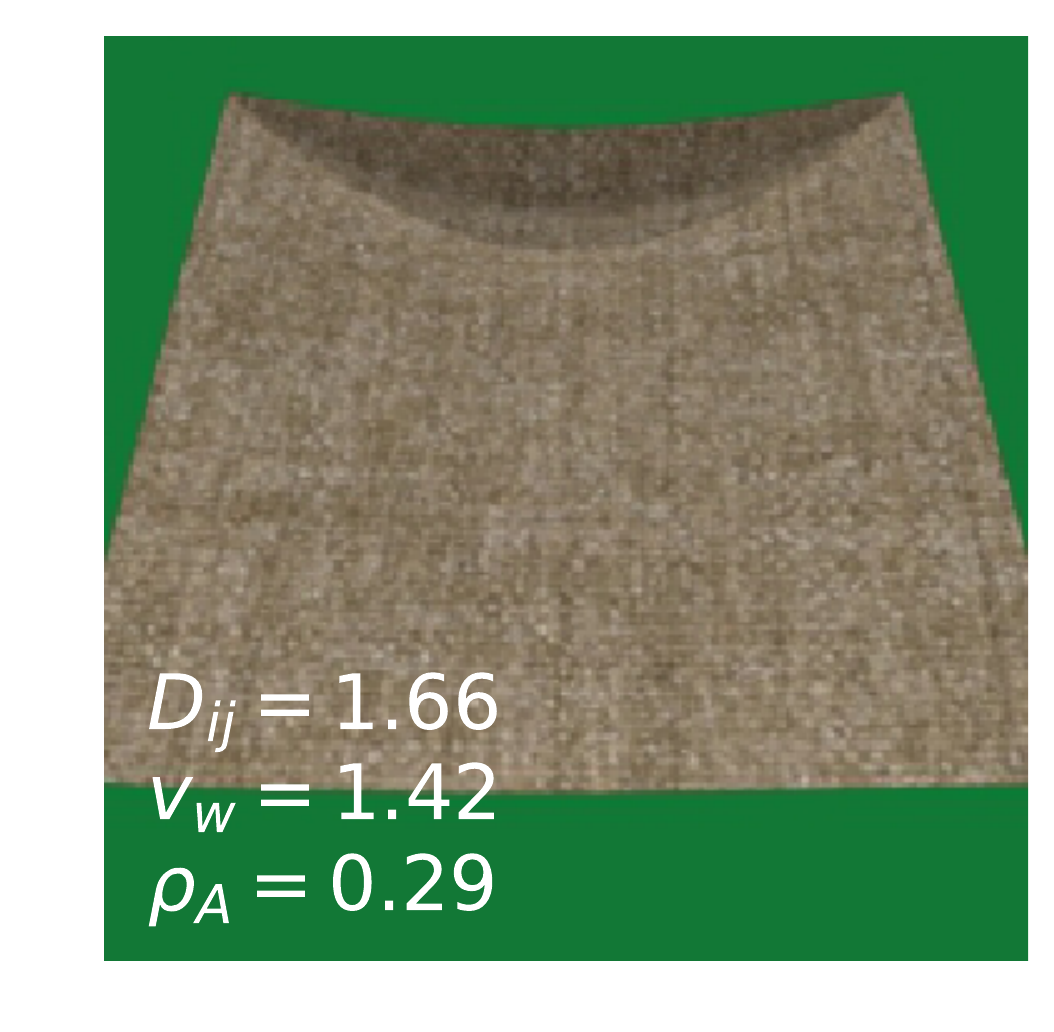}}
        \end{subfigure}
        \hfill
        \begin{subfigure}{.19\textwidth}
            \centering
            \fcolorbox{lightgray}{white}{\includegraphics[width=\textwidth,trim={1.3cm 1.1cm 0.35cm 0.8cm},clip]{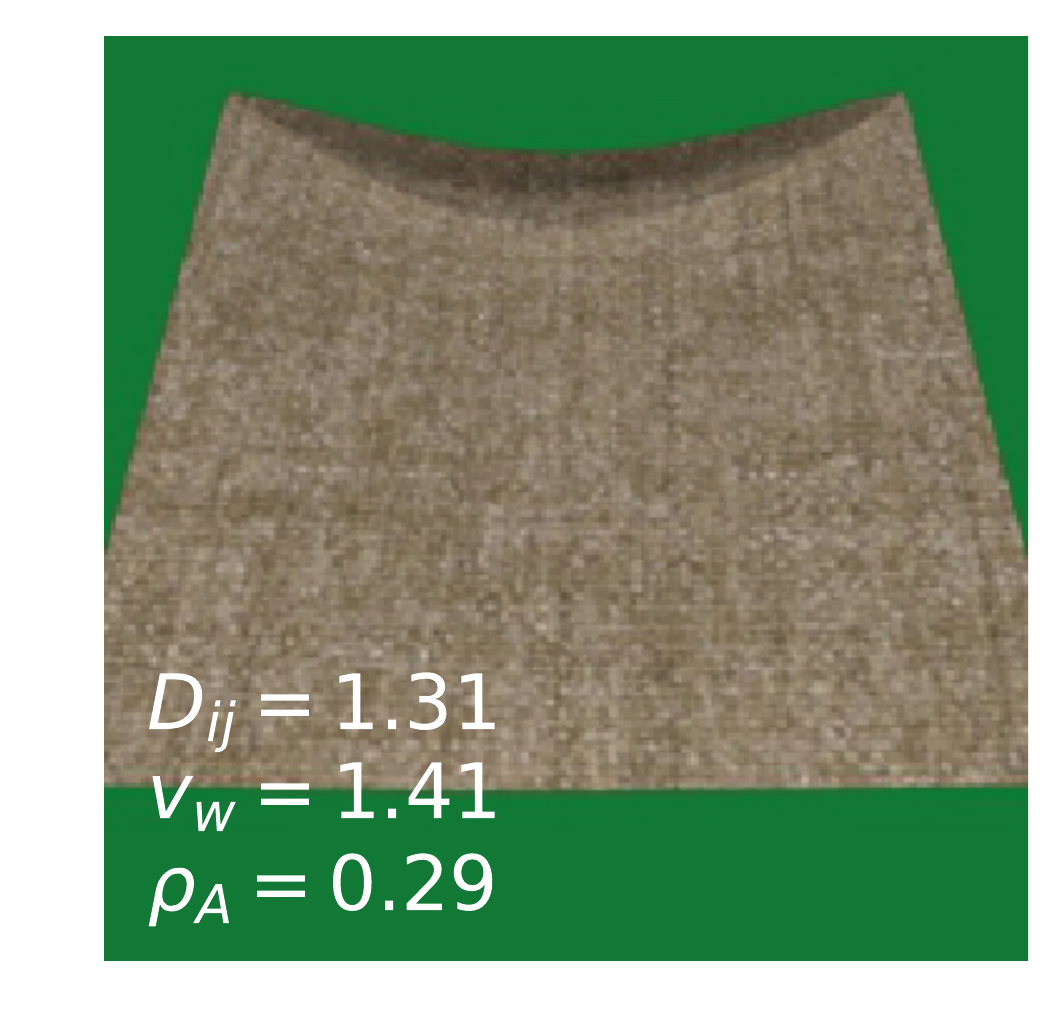}}
        \end{subfigure}
    \end{minipage}
    \\
    \begin{subfigure}{\textwidth}
        \centering
        \includegraphics[width=0.9\textwidth]{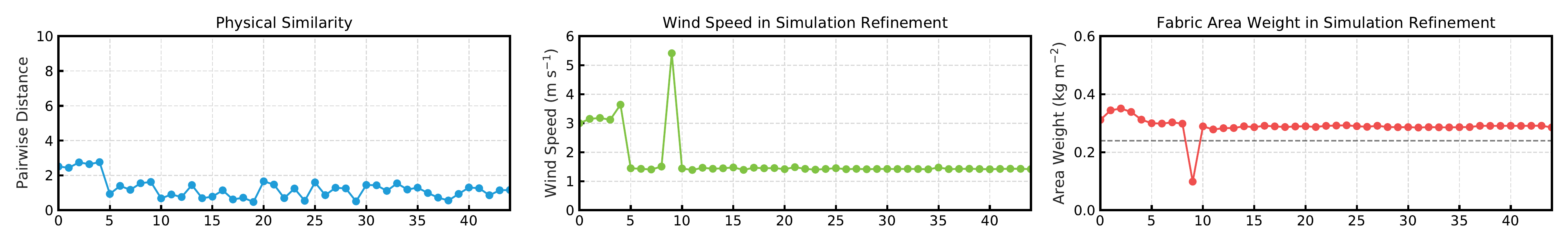}
    \end{subfigure}
  \end{figure*}
  \begin{figure*}
    \fboxsep=0mm  %
    \fboxrule=2pt %
    \centering
    \begin{minipage}[c]{0.85\textwidth}
        \centering
        \begin{subfigure}{.19\textwidth}
            \centering
            \fcolorbox{greencustom}{white}{\includegraphics[width=\textwidth,trim={1.2cm 1cm 0.3cm 0.4cm},clip]{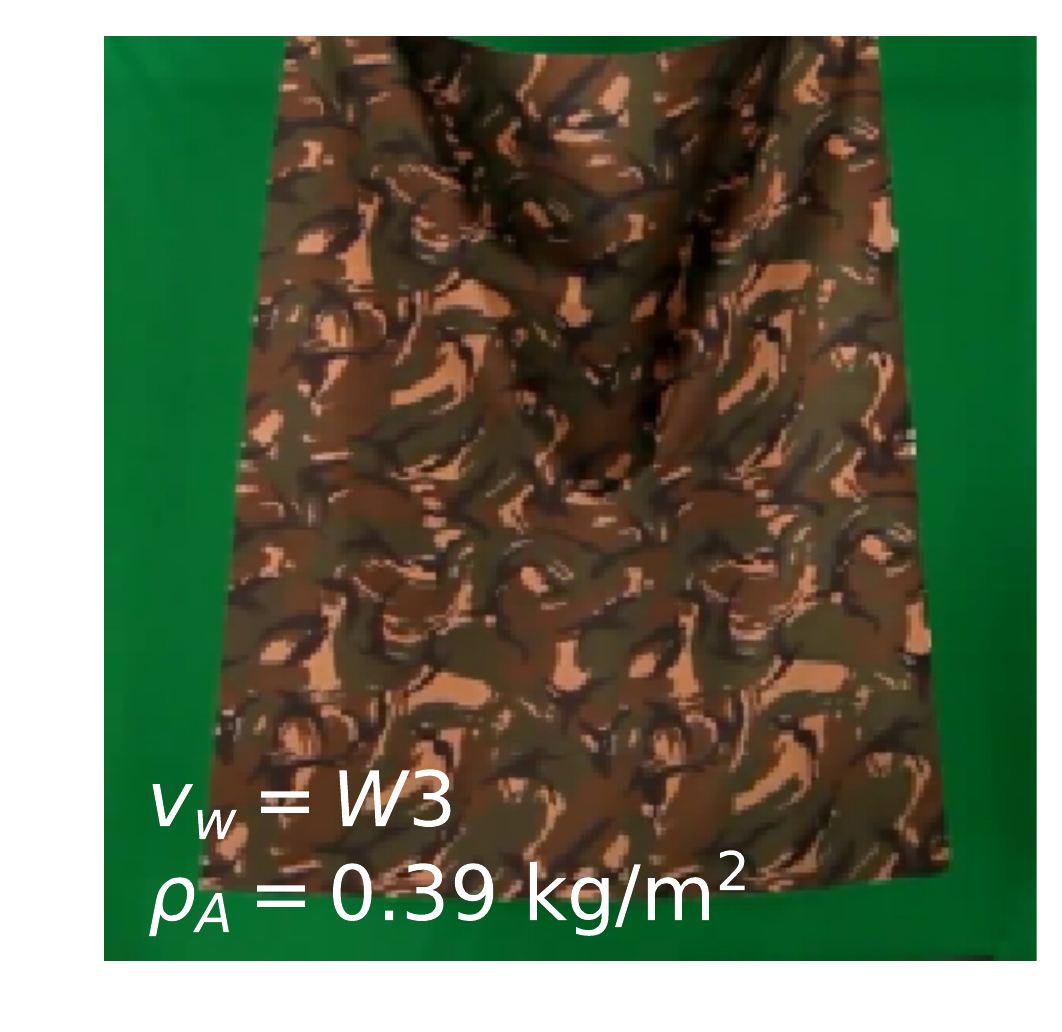}}
        \end{subfigure}
        \hfill
        \begin{subfigure}{.19\textwidth}
            \centering
            \fcolorbox{lightgray}{white}{\includegraphics[width=\textwidth,trim={1.3cm 1.1cm 0.35cm 0.8cm},clip]{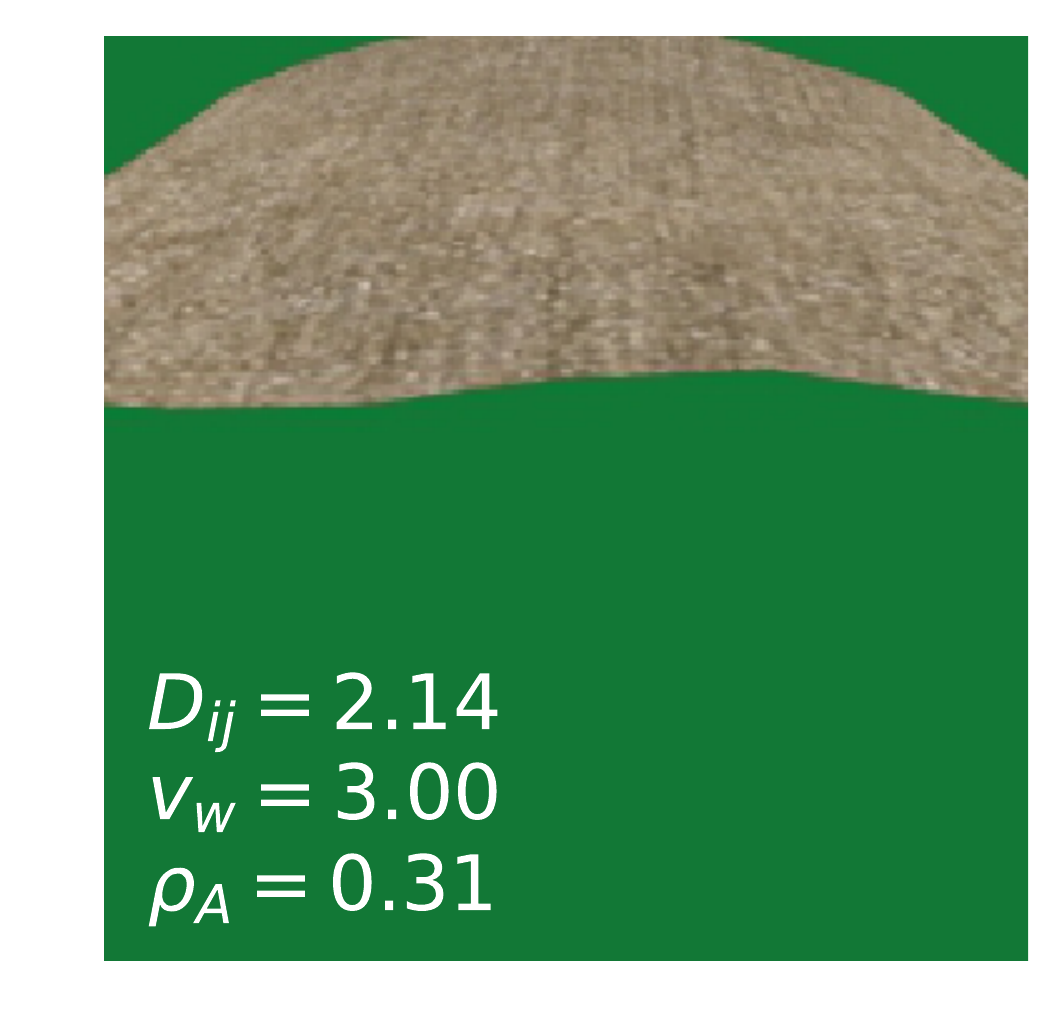}}
        \end{subfigure}
        \hfill
        \begin{subfigure}{.19\textwidth}
            \centering
            \fcolorbox{lightgray}{white}{\includegraphics[width=\textwidth,trim={1.3cm 1.1cm 0.35cm 0.8cm},clip]{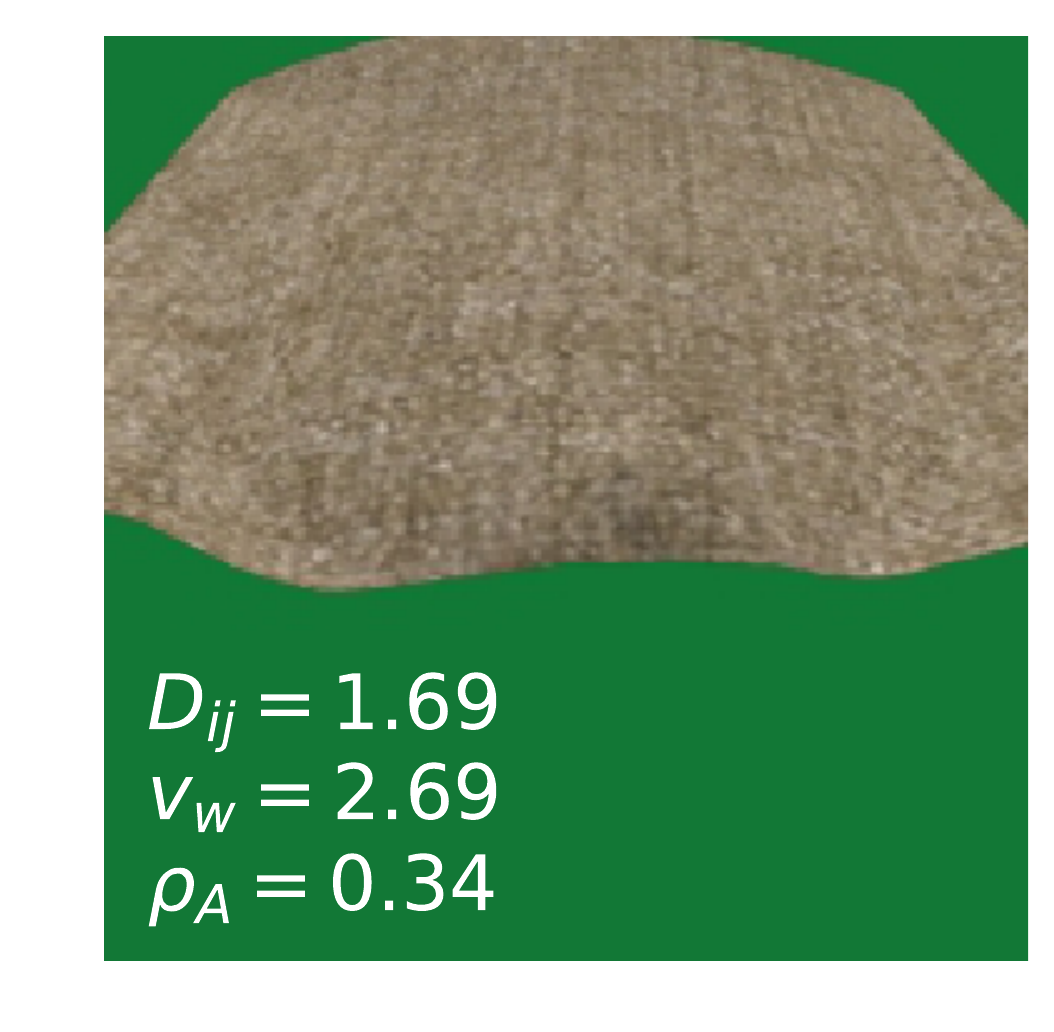}}
        \end{subfigure}
        \hfill
        \begin{subfigure}{.19\textwidth}
            \centering
            \fcolorbox{lightgray}{white}{\includegraphics[width=\textwidth,trim={1.3cm 1.1cm 0.35cm 0.8cm},clip]{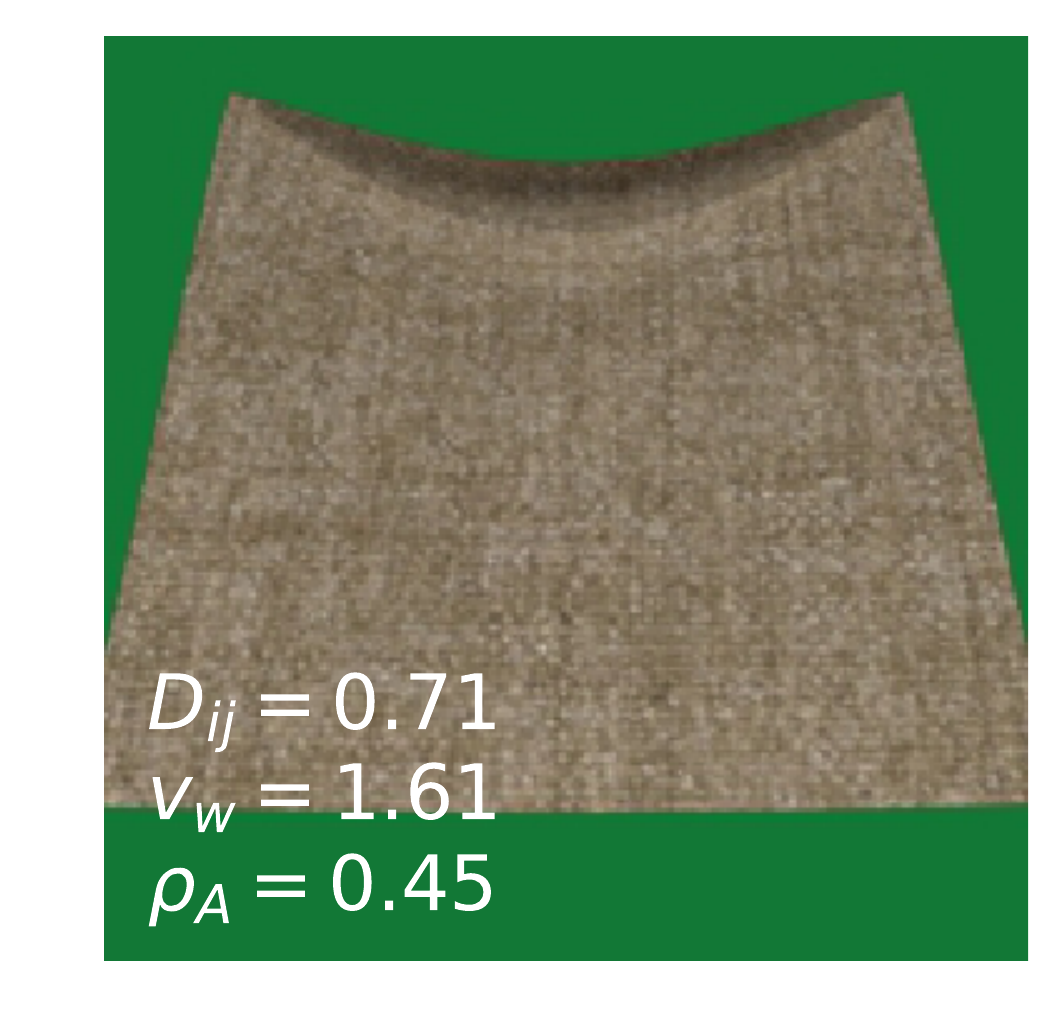}}
        \end{subfigure}
        \hfill
        \begin{subfigure}{.19\textwidth}
            \centering
            \fcolorbox{lightgray}{white}{\includegraphics[width=\textwidth,trim={1.3cm 1.1cm 0.35cm 0.8cm},clip]{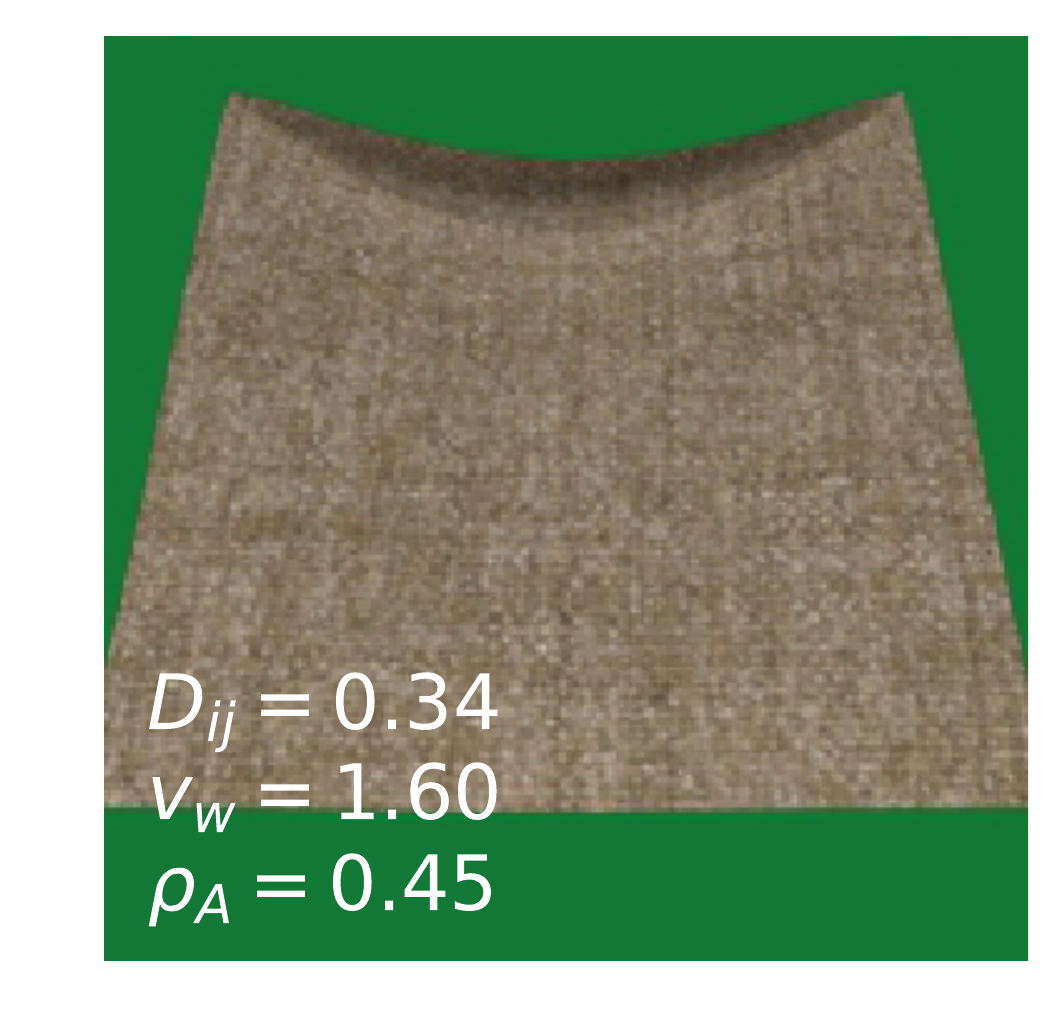}}
        \end{subfigure}
    \end{minipage}
    \\
    \begin{subfigure}{\textwidth}
        \centering
        \includegraphics[width=0.9\textwidth]{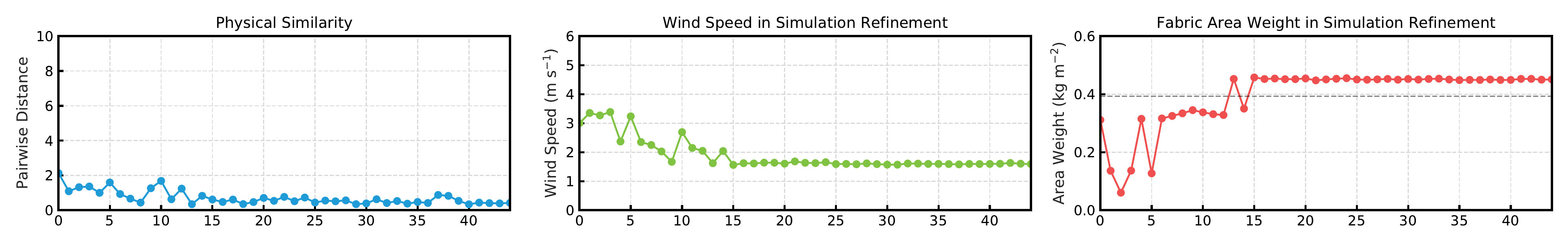}
    \end{subfigure}
    \vspace{3mm}
    \renewcommand\thefigure{4}
    \caption{Results for our \textbf{hanging cloth refined-measurements} for a random target video capturing the hanging cloth as recorded by Bouman \etal \cite{bouman2013estimating}. The five images are the center frames of the real-world target video (left) and simulations throughout the refinement process after $t = 0,10,20,40$ optimization steps. Optimization is performed over all $16$ intrinsic cloth parameters $\btheta_i$ and $1$ external wind speed $\btheta_e$. We plot the estimated cloth material area weight (\kgmm{}) and wind speed velocity (\ms{}), although we only have access to the true material area weight (dashed horizontal line). For this dataset, the wind speed has three settings of increasing wind speed: W1, W2 and W3. \emph{Top row:} The cloth's true area weight is $0.17$ \kgmm{} while the final measurement attains $0.22$ \kgmm{} after only $10$ iterations. \emph{Center row:} The cloth's true area weight is $0.24$ \kgmm{} while the prediction is $0.29$ \kgmm{}. While the ground-truth wind speed is not known, the wind speed in the simulations seems like an underestimate. \emph{Top:} A heavier cloth at $0.39$ \kgmm{} while the simulation measures $0.45$ \kgmm{}. \label{fig:supp_simrefine_bouman_results}}
  \end{figure*}

\begin{figure*}[h!]
  \fboxsep=0mm  %
  \fboxrule=2pt %
  \vspace{3mm}
  \centering
  \begin{minipage}[c]{0.85\textwidth}
      \centering
      \begin{subfigure}{.19\textwidth}
          \centering
          \fcolorbox{greencustom}{white}{\includegraphics[width=\textwidth,trim={1.2cm 1cm 0.3cm 0.4cm},clip]{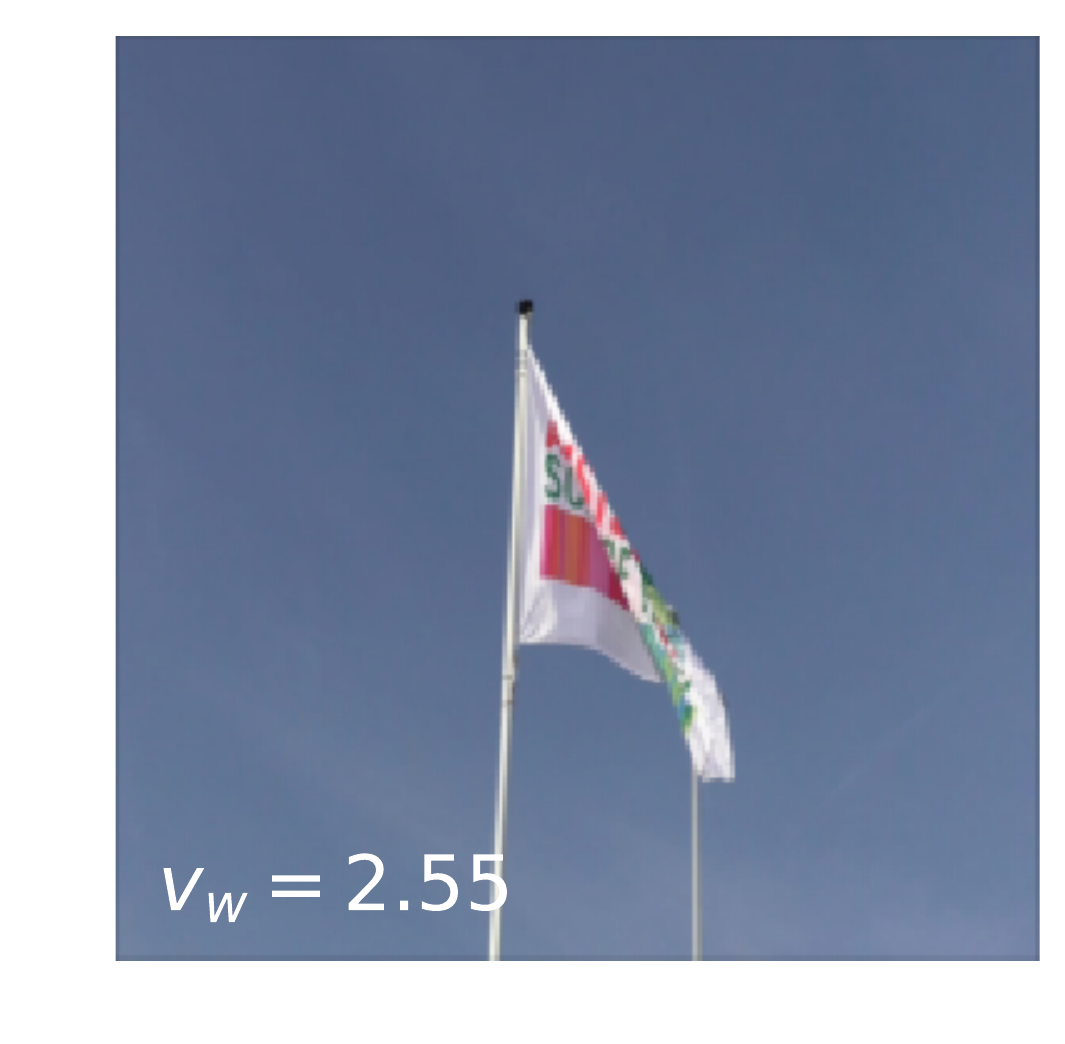}}
      \end{subfigure}
      \hfill
      \begin{subfigure}{.19\textwidth}
          \centering
          \fcolorbox{lightgray}{white}{\includegraphics[width=\textwidth,trim={1.3cm 1.1cm 0.35cm 0.8cm},clip]{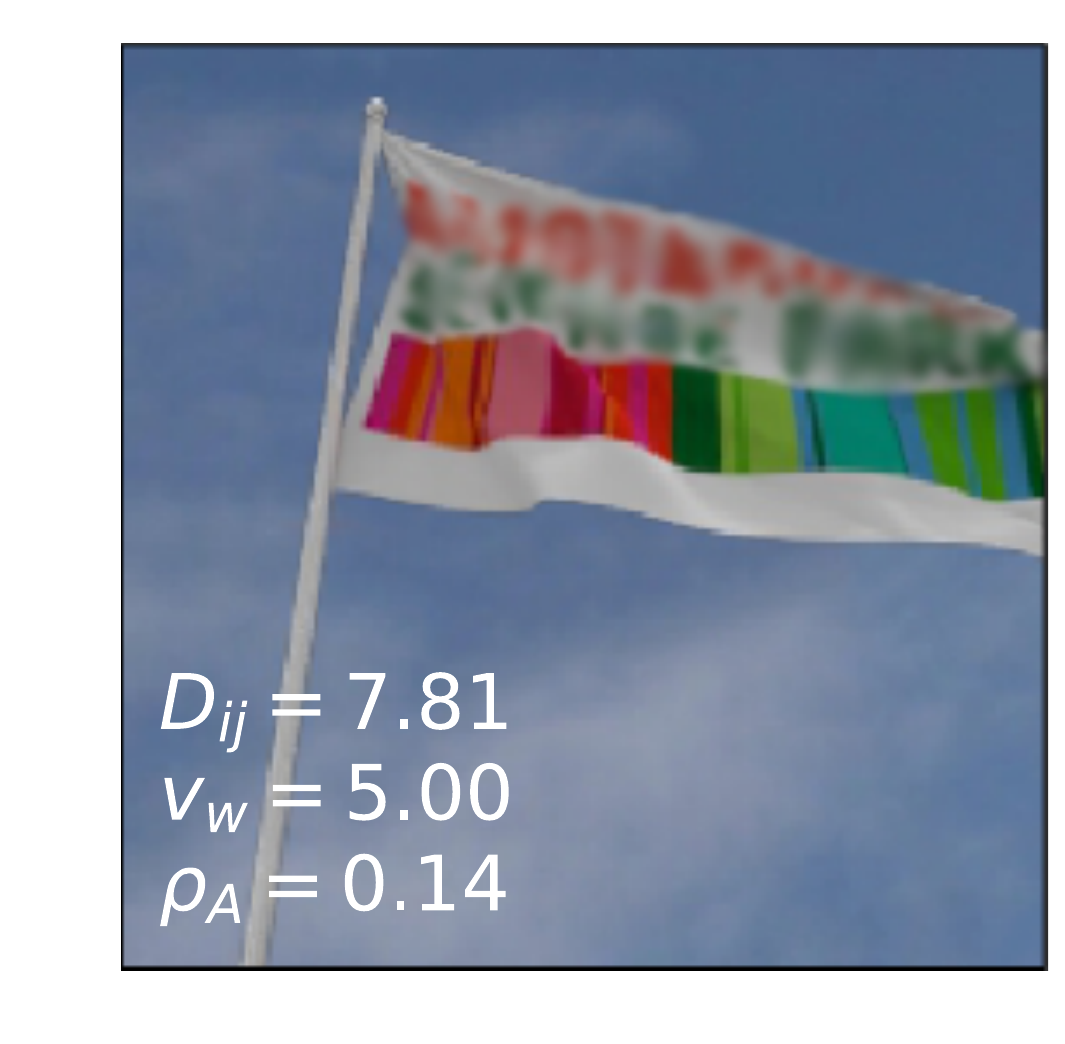}}
      \end{subfigure}
      \hfill
      \begin{subfigure}{.19\textwidth}
          \centering
          \fcolorbox{lightgray}{white}{\includegraphics[width=\textwidth,trim={1.3cm 1.1cm 0.35cm 0.8cm},clip]{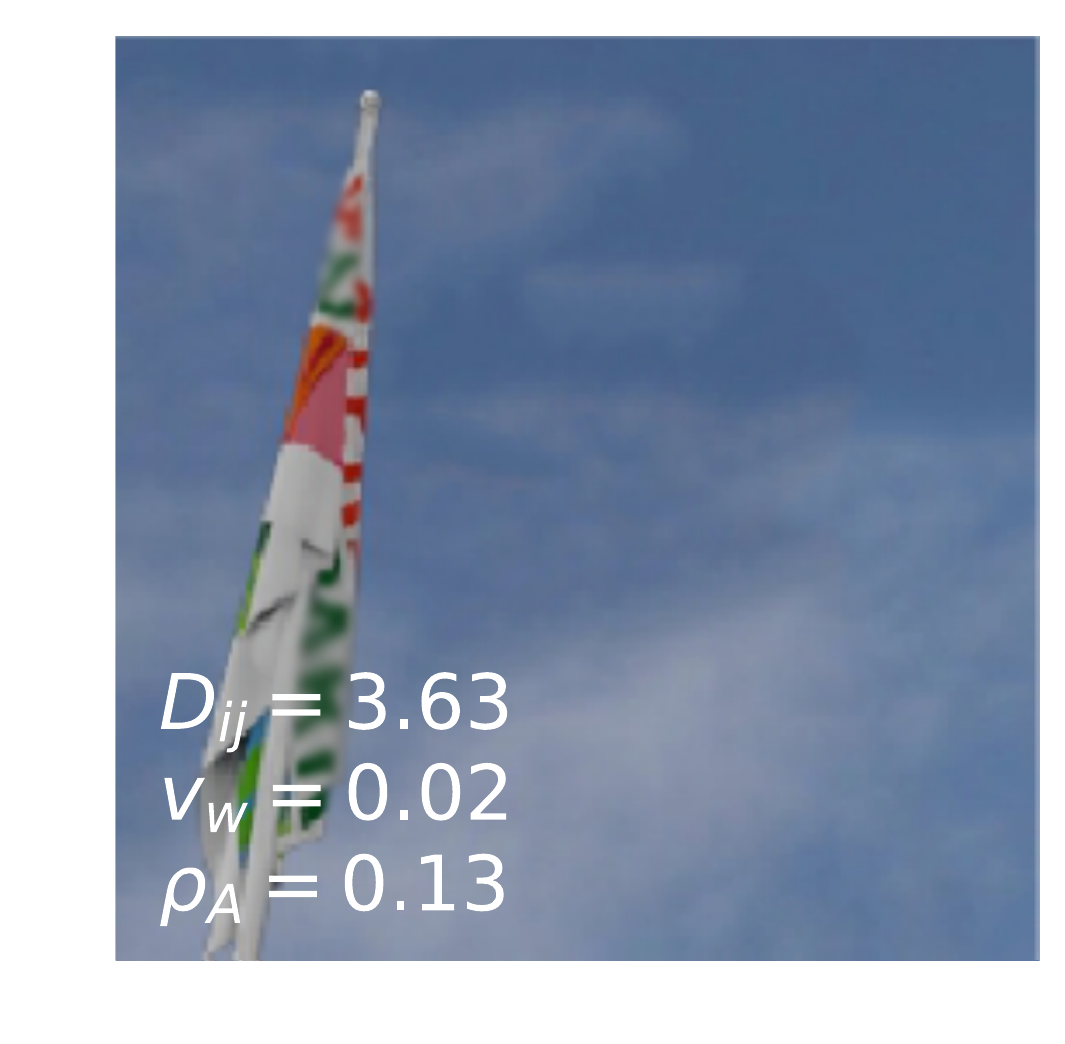}}
      \end{subfigure}
      \hfill
      \begin{subfigure}{.19\textwidth}
          \centering
          \fcolorbox{lightgray}{white}{\includegraphics[width=\textwidth,trim={1.3cm 1.1cm 0.35cm 0.8cm},clip]{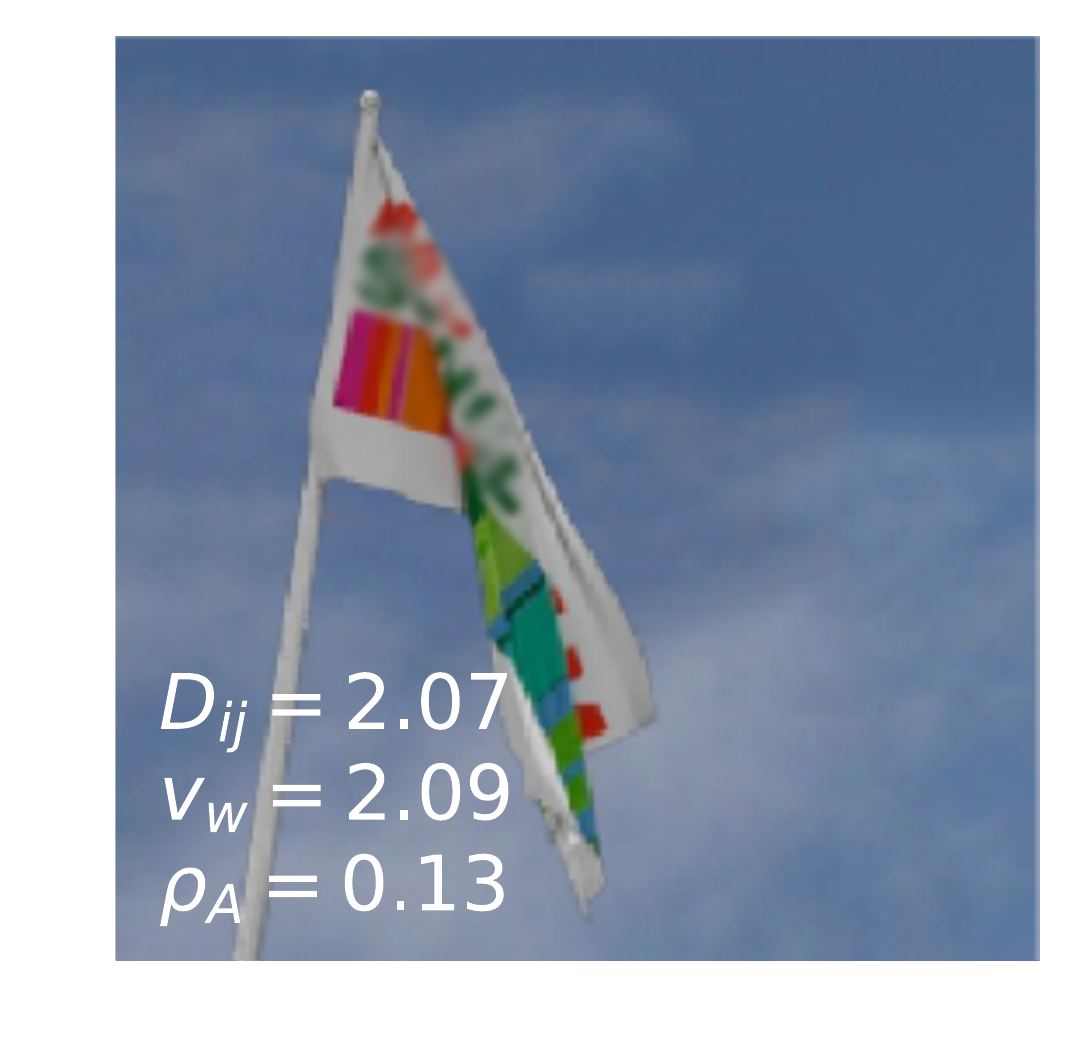}}
      \end{subfigure}
      \hfill
      \begin{subfigure}{.19\textwidth}
          \centering
          \fcolorbox{lightgray}{white}{\includegraphics[width=\textwidth,trim={1.3cm 1.1cm 0.35cm 0.8cm},clip]{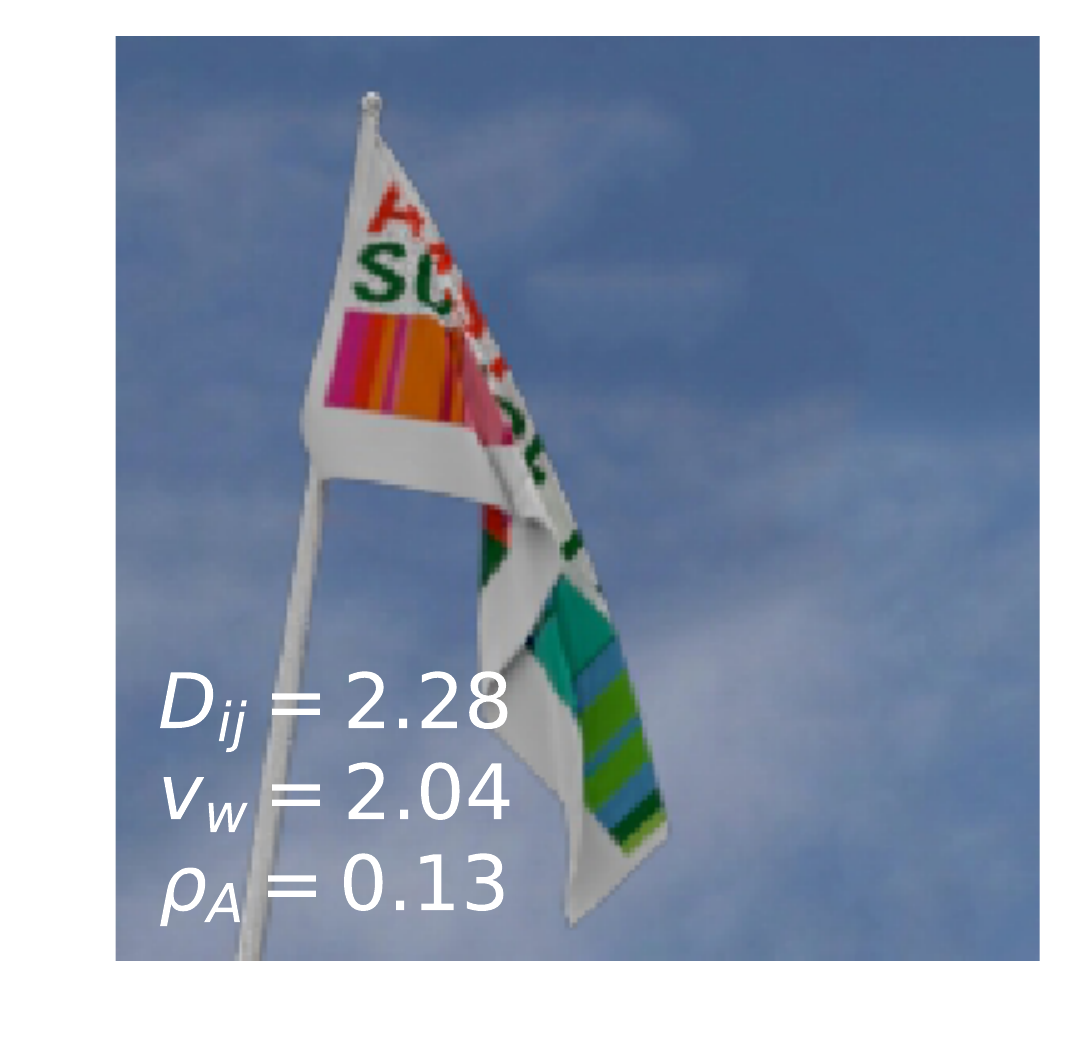}}
      \end{subfigure}
  \end{minipage}
  \\
  \begin{subfigure}{\textwidth}
      \centering
      \includegraphics[width=0.9\textwidth]{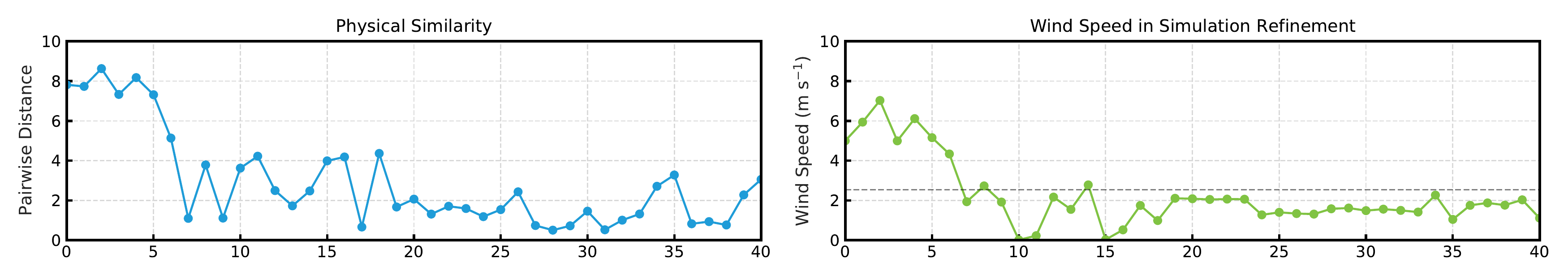}
  \end{subfigure}
\end{figure*}
\begin{figure*}
  \fboxsep=0mm  %
  \fboxrule=2pt %
  \centering
  \begin{minipage}[c]{0.85\textwidth}
      \centering
      \begin{subfigure}{.19\textwidth}
          \centering
          \fcolorbox{greencustom}{white}{\includegraphics[width=\textwidth,trim={1.2cm 1cm 0.3cm 0.4cm},clip]{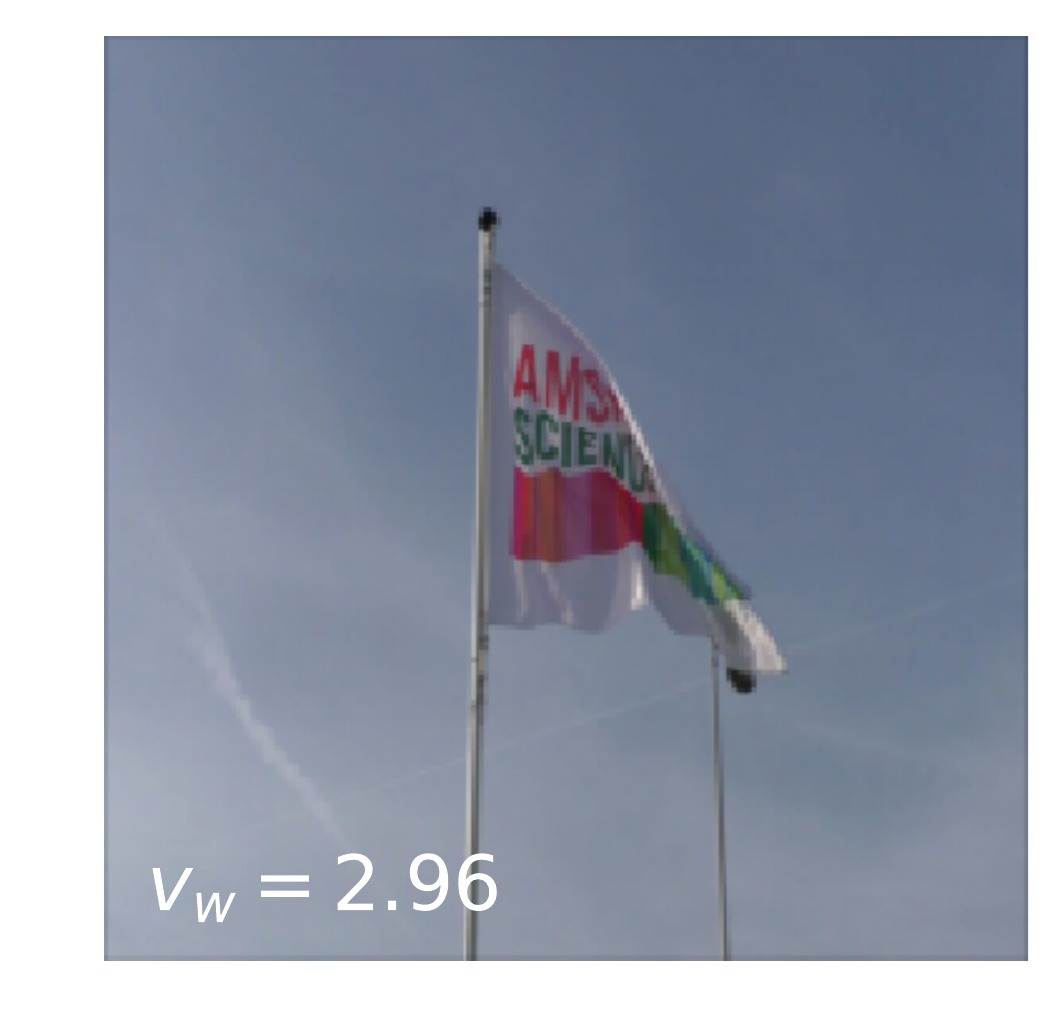}}
      \end{subfigure}
      \hfill
      \begin{subfigure}{.19\textwidth}
          \centering
          \fcolorbox{lightgray}{white}{\includegraphics[width=\textwidth,trim={1.3cm 1.1cm 0.35cm 0.8cm},clip]{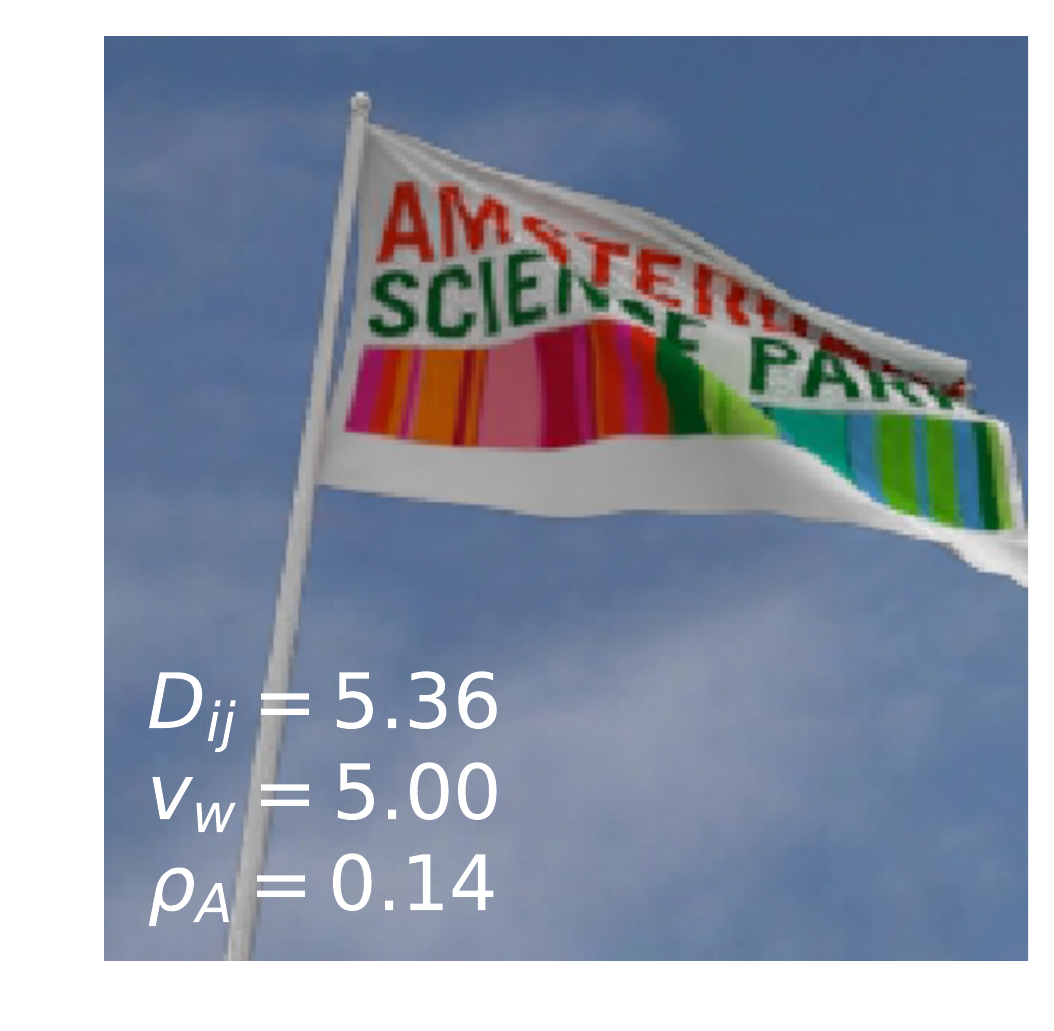}}
      \end{subfigure}
      \hfill
      \begin{subfigure}{.19\textwidth}
          \centering
          \fcolorbox{lightgray}{white}{\includegraphics[width=\textwidth,trim={1.3cm 1.1cm 0.35cm 0.8cm},clip]{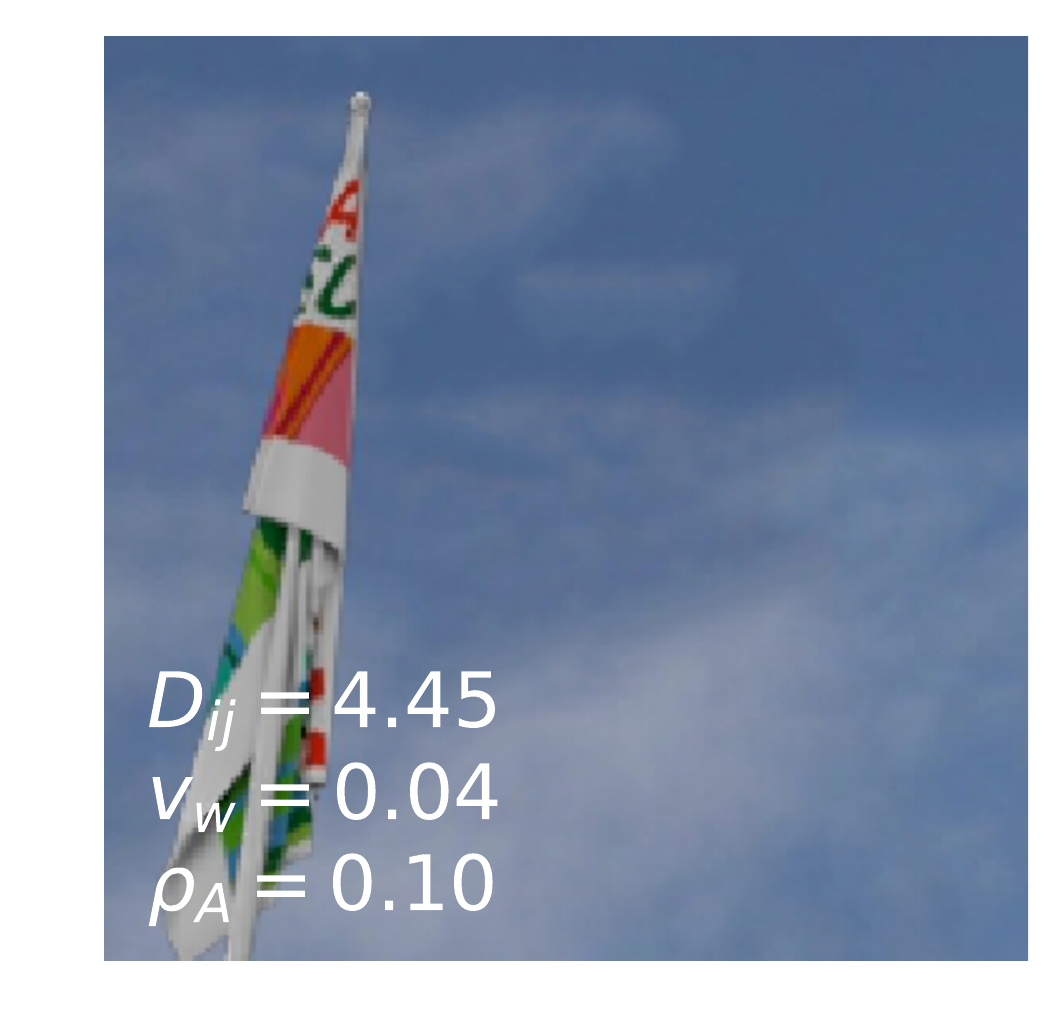}}
      \end{subfigure}
      \hfill
      \begin{subfigure}{.19\textwidth}
          \centering
          \fcolorbox{lightgray}{white}{\includegraphics[width=\textwidth,trim={1.3cm 1.1cm 0.35cm 0.8cm},clip]{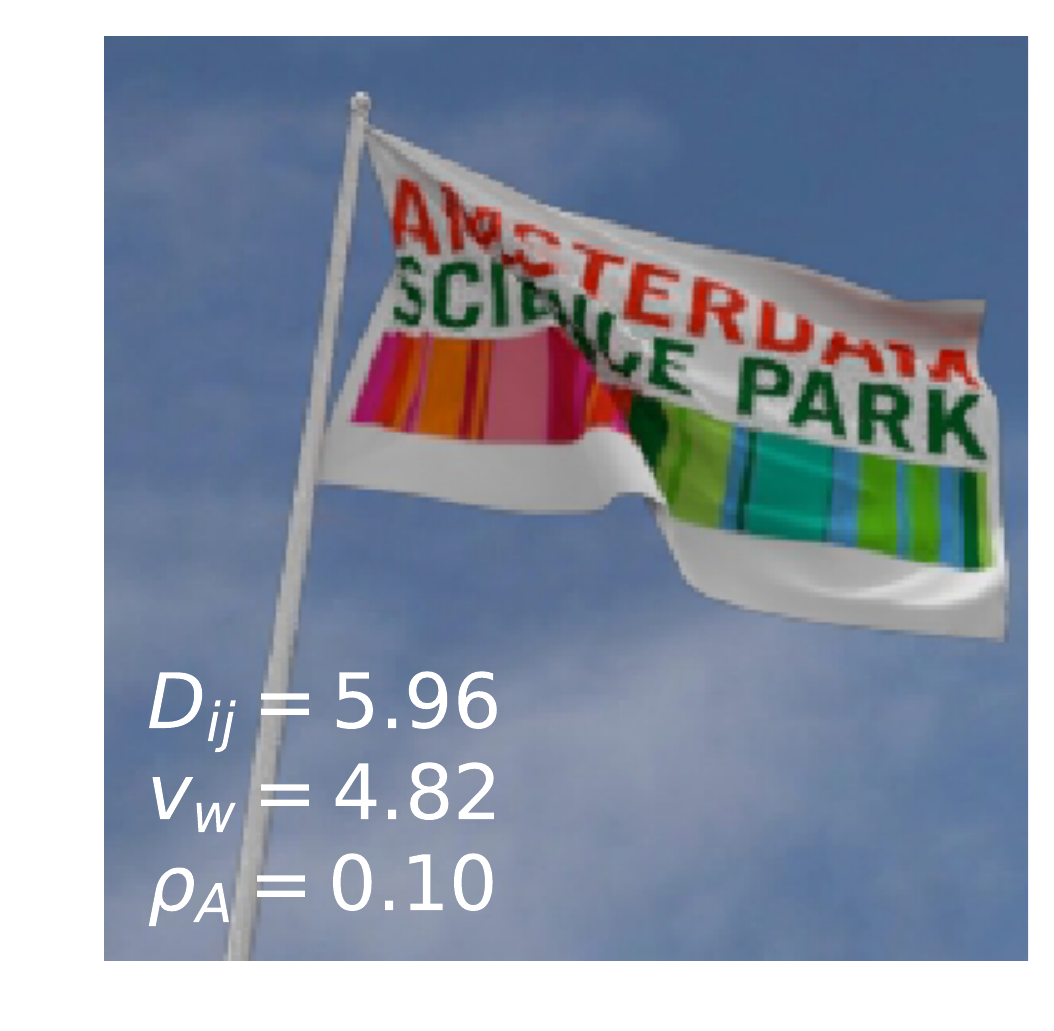}}
      \end{subfigure}
      \hfill
      \begin{subfigure}{.19\textwidth}
          \centering
          \fcolorbox{lightgray}{white}{\includegraphics[width=\textwidth,trim={1.3cm 1.1cm 0.35cm 0.8cm},clip]{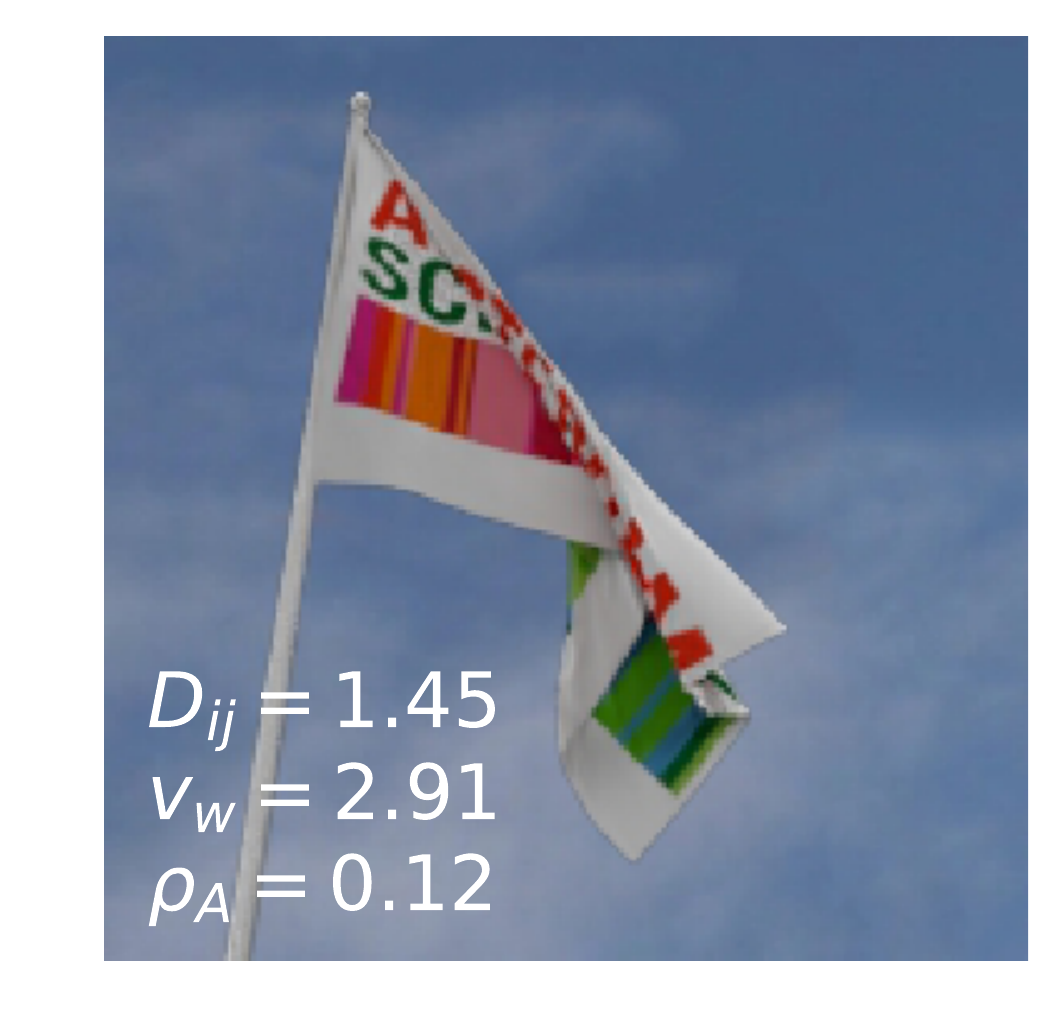}}
      \end{subfigure}
  \end{minipage}
  \\
  \begin{subfigure}{\textwidth}
      \centering
      \includegraphics[width=0.9\textwidth]{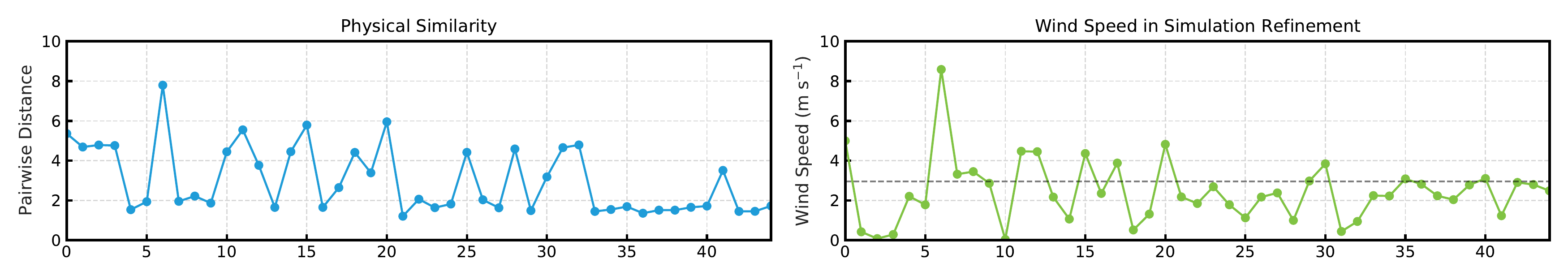}
  \end{subfigure}
\end{figure*}
\begin{figure*}
  \fboxsep=0mm  %
  \fboxrule=2pt %
  \centering
  \begin{minipage}[c]{0.85\textwidth}
      \centering
      \begin{subfigure}{.19\textwidth}
          \centering
          \fcolorbox{greencustom}{white}{\includegraphics[width=\textwidth,trim={1.2cm 1cm 0.3cm 0.4cm},clip]{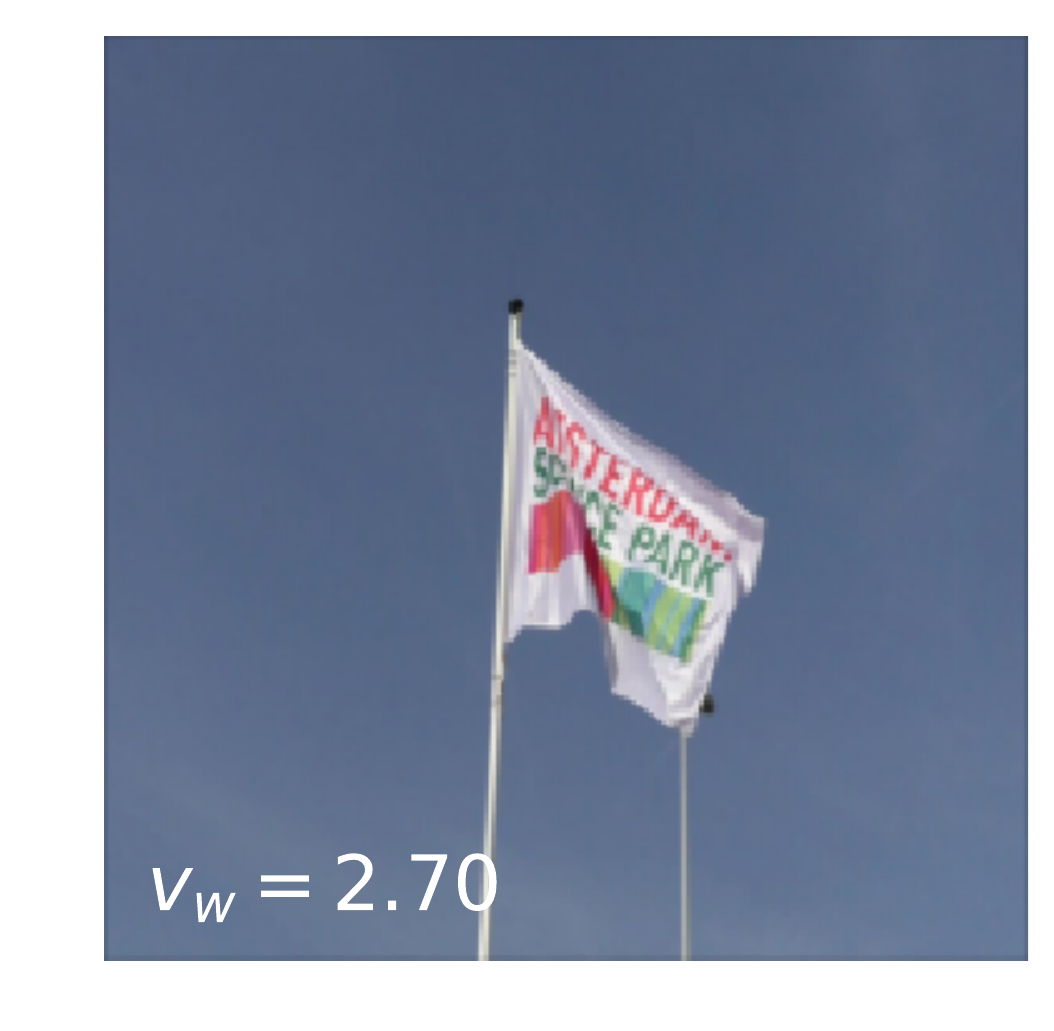}}
      \end{subfigure}
      \hfill
      \begin{subfigure}{.19\textwidth}
          \centering
          \fcolorbox{lightgray}{white}{\includegraphics[width=\textwidth,trim={1.3cm 1.1cm 0.35cm 0.8cm},clip]{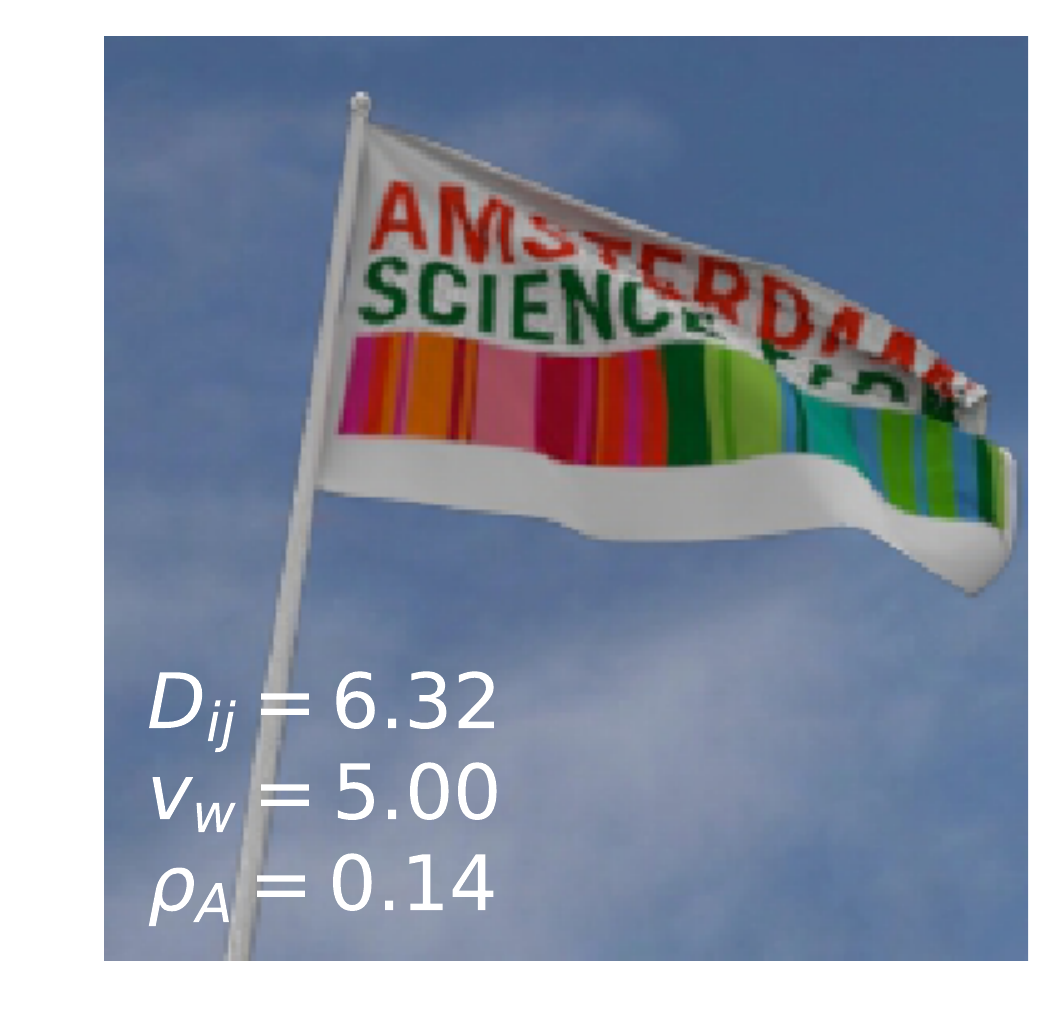}}
      \end{subfigure}
      \hfill
      \begin{subfigure}{.19\textwidth}
          \centering
          \fcolorbox{lightgray}{white}{\includegraphics[width=\textwidth,trim={1.3cm 1.1cm 0.35cm 0.8cm},clip]{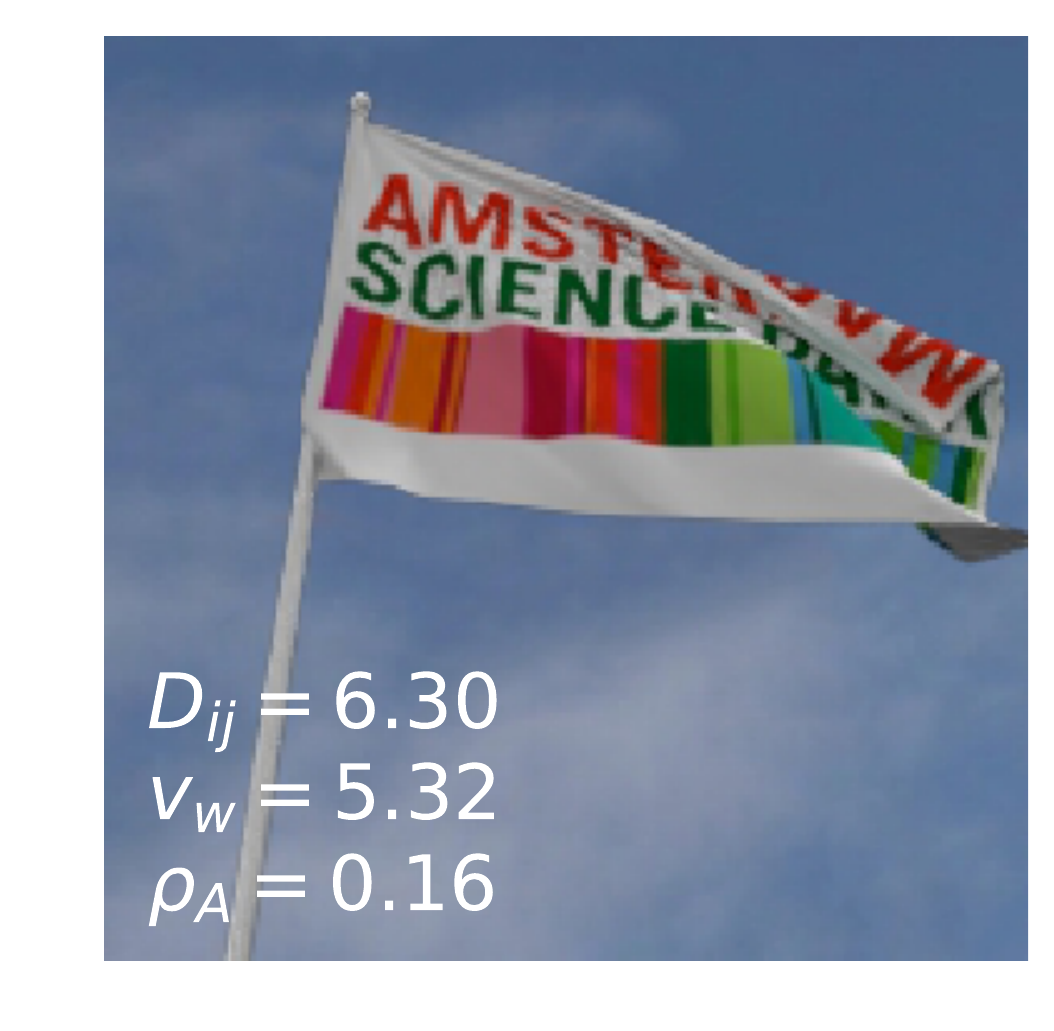}}
      \end{subfigure}
      \hfill
      \begin{subfigure}{.19\textwidth}
          \centering
          \fcolorbox{lightgray}{white}{\includegraphics[width=\textwidth,trim={1.3cm 1.1cm 0.35cm 0.8cm},clip]{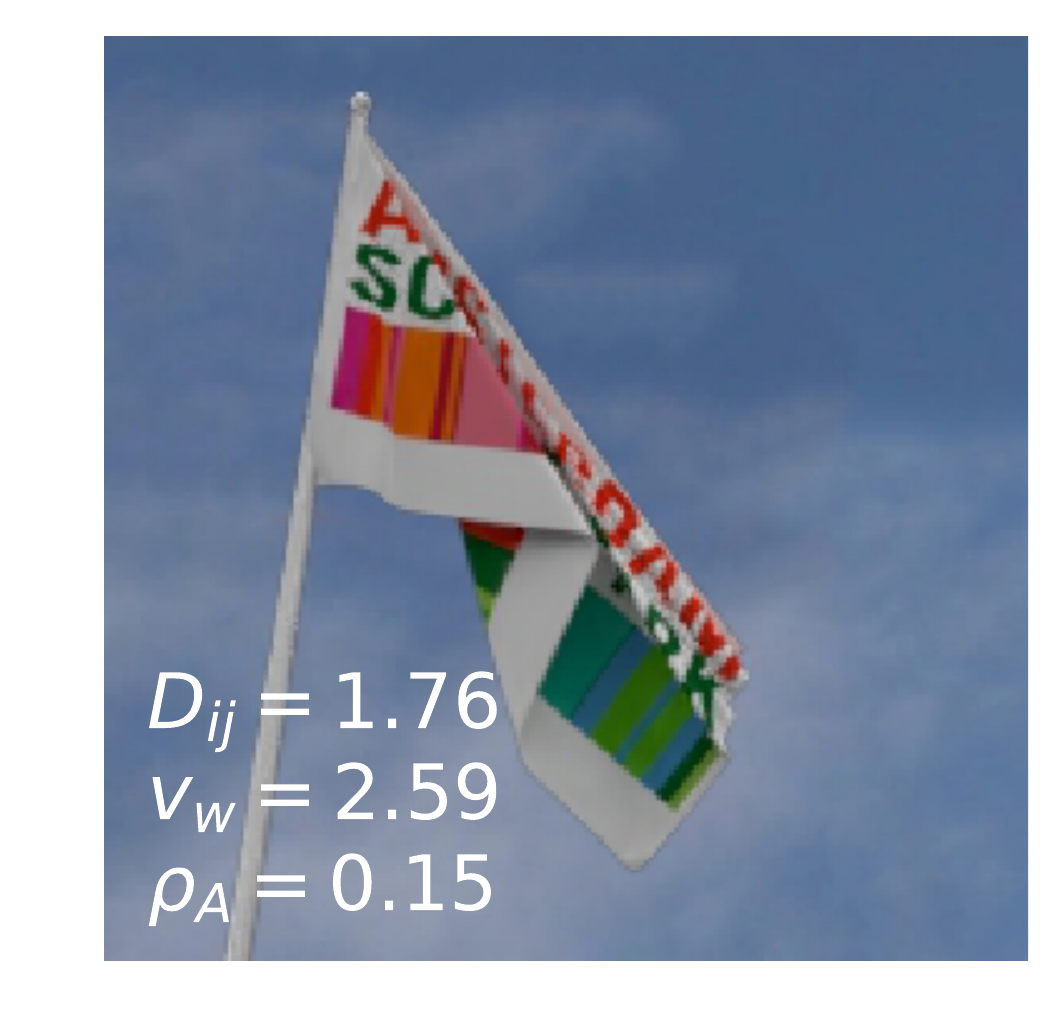}}
      \end{subfigure}
      \hfill
      \begin{subfigure}{.19\textwidth}
          \centering
          \fcolorbox{lightgray}{white}{\includegraphics[width=\textwidth,trim={1.3cm 1.1cm 0.35cm 0.8cm},clip]{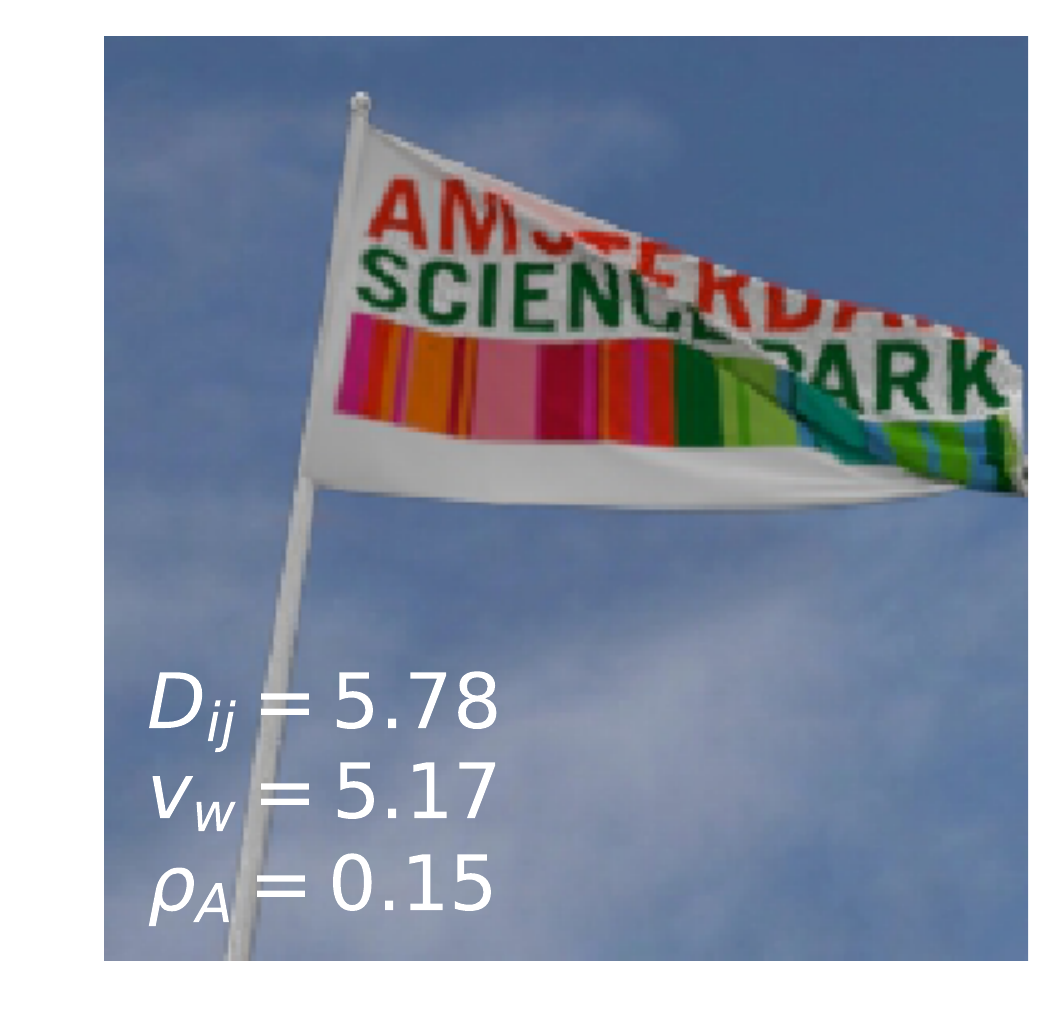}}
      \end{subfigure}
  \end{minipage}
  \\
  \begin{subfigure}{\textwidth}
      \centering
      \includegraphics[width=0.9\textwidth]{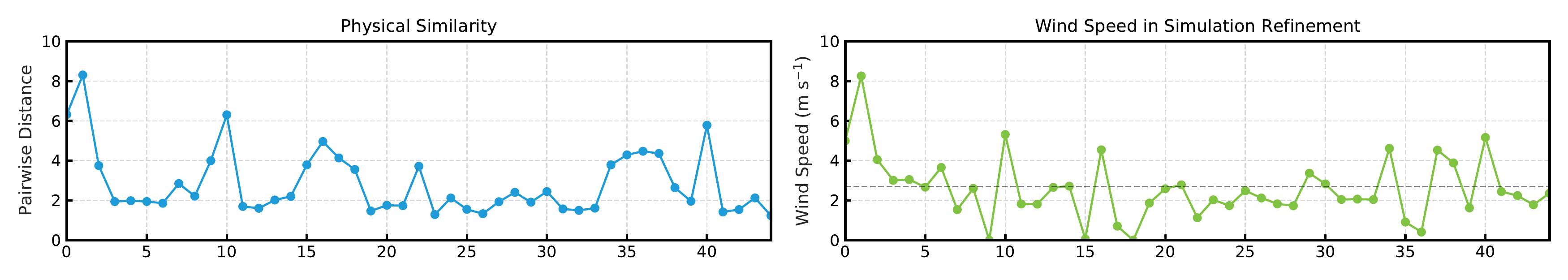}
  \end{subfigure}
  \vspace{3mm}
  \renewcommand\thefigure{3}
  \caption{Additional results for our \textbf{FlagReal refined-measurements} for a random target video capturing the flag in the wind (corresponding to Figure~8 in the main paper). The five images are the center frames of the real-world target video (left) and simulations throughout the refinement process after $t = 0,10,20,40$ optimization steps. Optimization is performed over all $16$ intrinsic cloth parameters $\btheta_i$ and $1$ external wind speed $\btheta_e$. We only visualize the simulated wind speed as it is the only parameter for which we have ground-truth (dashed line). \emph{Top row:} successful optimization example; although the scale between real observation and simulation is different, our method is able to precisely determine the external wind speed. The real-world video has a ground-truth wind speed of $2.46$ \ms{} while the refinement procedure finds a wind speed of $2.34$ \ms{} in less than $10$ optimization steps. \emph{Center row:} Another successful optimization example. The real-world video has a ground-truth wind speed of $2.96$ \ms{} while the refinement procedure finds a wind speed of $2.36$ \ms{} after $45$ refinement steps. \emph{Bottom row:} failure case; even though after $45$ steps the wind speed is approximately correct, the optimization procedure has not converged. \label{fig:supp_simrefine_flags_results}}
  
\end{figure*}

\begin{figure*}[h!]
    \centering
    \captionsetup{width=.82\linewidth}
    \captionof{lstlisting}[ListingsCaption]{The ArcSim base \textbf{configuration for flags} \cite{narain2012adaptive} as JSON file to be read by the simulator. The simulation runs on a flag mesh of $3:2$ aspect ratio in a constant wind field defined by the wind speed $\*\theta_e$ parameter. During simulation we only consider a wind field in a single direction, but during rendering we use multiple relative camera orientations creating the appearance of varying wind directions. The intrinsic cloth material parameters $\*\theta_i$ reside inside the material configuration file (\Cref{code:json-material-file}). \label{code:json-arcsim-config}} 
    \begin{minipage}{0.8\linewidth}
\begin{lstlisting}[language=json]
{
    "frame_time": 0.04,
    "frame_steps": 8,
    "duration": 20,
    "cloths": [{
        "mesh": "meshes/flag.obj",
	    "transform": {
            "translate": [0, 0, 0], 
            "rotate": [120, 1, 1, 1]
        },
        "materials": [{
            "data": "materials/camel-ponte-roma.json",
            "thicken": 2,
            "strain_limits": [0.95, 1.05]
    	}],
        "remeshing": {
            "refine_angle": 0.3,
            "refine_compression": 0.01,
            "refine_velocity": 1,
            "size": [20e-3, 500e-3],
            "aspect_min": 0.2
        }
    }],
    "handles": [{
        "nodes": [0,3]
    }],
    "gravity": [0, 0, -9.81],
    "wind": {
        "velocity": [wind_speed, 0, 0]
    },
    "magic": {
        "repulsion_thickness": 10e-3, 
        "collision_stiffness": 1e6
    }
}
\end{lstlisting}
    \end{minipage}
  \end{figure*}

  \begin{figure*}[h!]
      \centering
      \captionsetup{width=.82\linewidth}
      \captionof{lstlisting}[ListingsCaption]{The ArcSim base \textbf{configuration for hanging cloth} as JSON file to be read by the simulator. The wind speed is defined on the horizontal plane ($x$ and $y$ components). Again, the starting point of the intrinsic cloth material parameters $\*\theta_i$ are given in \Cref{code:json-material-file}. However, in comparison to the flag simulations, we set a much larger variety of fabrics and define the fabric area weight range to correspond to the hanging cloth dataset \cite{bouman2013estimating}. \label{code:json-arcsim-config-cloth}} 
      \begin{minipage}{0.8\linewidth}
\begin{lstlisting}[language=json]
{
    "frame_time": 0.04,
    "frame_steps": 8,
    "duration": 20,
    "cloths": [{
        "mesh": "meshes/square.obj",
	    "transform": {
            "translate": [0, 0, 0], 
            "rotate": [120, 1, 1, 1]},
        "materials": [{
	    "data": "materials/camel-ponte-roma.json",
            "thicken": 1,
            "strain_limits": [0.95, 1.05]
	    }],
        "remeshing": {
            "refine_angle": 0.3,
            "refine_compression": 0.01,
            "refine_velocity": 1,
            "size": [20e-3, 500e-3],
            "aspect_min": 0.2
        }
    }],
    "motions": [],
    "handles": [{"nodes": [2,3]}],
    "gravity": [0, 0, -9.8],
    "wind": {"velocity": [wind_speed_x, wind_speed_x, 0]},
    "magic": {"repulsion_thickness": 10e-3, "collision_stiffness": 1e6}
}
\end{lstlisting}
      \end{minipage}
    \end{figure*}

  \begin{figure*}[h!]
    \centering
    \captionsetup{width=.82\linewidth}
    \captionof{lstlisting}[ListingsCaption]{The \textbf{ArcSim material configuration} \cite{narain2012adaptive} as JSON file to be consumed by the simulator. As base material, we use ``camel ponte roma'' with its properties determined in the mechanical setup by \cite{wang2011data}. This file specifies the cloth's area weight, bending stiffness coefficients and stretching coefficients. As flags are of strong, weather-resistant material, we optimize over the area weight ($1\times$) and bending parameters ($15\times$). Together these $16$ parameters define $\*\theta_i$. For hanging cloth, we also keep the bending parameters fixed to constrain the number of free parameters. \label{code:json-material-file}} 
    \begin{minipage}{0.85\linewidth}
\begin{lstlisting}[language=json]
{
    "density": 0.135,
    "bending": [
        [36.3483e-6, 49.5855e-6, 45.7440e-6, 47.4133e-6, 20.7266e-6],
        [33.0132e-6, 29.7443e-6, 35.1036e-6, 34.0410e-6, 14.4399e-6],
        [37.1575e-6, 34.1074e-6, 33.2294e-6, 34.6855e-6, 10.4399e-6]
    ],
    "stretching": [
        [31.146198,  -12.802702,  44.028667,  31.896357],
        [78.707756,   26.754574, 268.680725,  27.743423],
        [67.368431,   77.767944, 182.273407, -14.661531],
        [113.367035,  54.802021, 175.126572,  44.657330],
        [144.294830, 111.404854, 138.422150, -29.861851],
        [143.933365,  49.654823, 191.777588,  39.491055]
    ]
}
\end{lstlisting}
    \end{minipage}
    \\[10mm]
    \captionof{table}{Exhaustive overview of the \textbf{render parameters} $\zeta$ for rendering the FlagSim and ClothSim datasets. \label{tab:supp-blender-render-parameters} }
    \ra{1.2}
    \small
    \begin{tabular}{llll}
      \toprule
      \textbf{Name} & \textbf{Description} & \textbf{Value/Range (Flags)} & \textbf{Value/Range (Cloth)} \\ 
      \midrule
      \texttt{background\_image} & Background image of scene & \multicolumn{2}{c}{Sampled from SUN397 \cite{xiao2010sun}} \\ 
      \texttt{background\_offset} & Background image translation & \multicolumn{2}{c}{$\sim\text{Uniform}(-20,+20)$} \\ 
      \texttt{background\_scale} & Background image scale  & \multicolumn{2}{c}{$\sim\text{Uniform}(0.6,1.0)$} \\ 
      \midrule
      \texttt{sun\_height} & The sun's height above the ground plane & \multicolumn{2}{c}{$\sim\text{Uniform}(4,10)$} \\ 
      \texttt{sun\_radius} & The sun's distance to mesh & \multicolumn{2}{c}{$\sim\text{Uniform}(0,\hphantom{0}5)$} \\ 
      \texttt{sun\_strength} & The sun's illumination strength & \multicolumn{2}{c}{$\sim\text{Uniform}(2,10)$} \\ 
      \texttt{sun\_shadow\_soft\_size} & The sun's shadow hardness & \multicolumn{2}{c}{$\sim\text{Uniform}(2,10)$} \\ 
      \midrule
      \texttt{cycles\_samples} & Cycles \cite{blender2018} number of render samples & \multicolumn{2}{c}{50} \\ 
      \texttt{cycles\_bounces} & Cycles \cite{blender2018} light bounces, object dependent & \multicolumn{2}{c}{$[0,6]$} \\ 
      \midrule
      \texttt{camera\_height} & The height above the ground plane & $\sim\text{Uniform}(0.2,3)$ & $\sim\text{Uniform}(0.5,2)$ \\ 
      \texttt{camera\_radius} & The distance to the mesh & $\sim\text{Uniform}(4,6)$ & $\sim\text{Uniform}(1,2.5)$ \\ 
      \texttt{camera\_angle} & The orientation w.r.t. wind direction & $\sim\text{Uniform}(-15,+15)$ & $\sim\text{Uniform}(-5,+5)$\\
      \midrule
      \texttt{mesh\_height} & The flag's height above the ground plane & $4.6$ & $2$ \\ 
      \texttt{mesh\_aspect\_ratio} & The flag's aspect ratio & $3:2$ & $1:1$ \\ 
      \texttt{mesh\_texture} & The flag/cloth texture & Sampled from $12$ countries & Sampled from \cite{yang2017learning} \\ 
      \bottomrule
  \end{tabular}
  
  \end{figure*}  

\end{document}